\documentclass[journal]{IEEEtran}

\usepackage{cite}	% *** CITATION PACKAGES ***
\usepackage[pdftex]{graphicx}	% *** GRAPHICS RELATED PACKAGES ***
\usepackage{amsmath}	% *** MATH PACKAGES ***
\usepackage{amsfonts}
\usepackage{amssymb}
\usepackage{cases}
\usepackage{ntheorem}
\usepackage{array}	% *** ALIGNMENT PACKAGES ***
\ifCLASSOPTIONcompsoc	% *** SUBFIGURE PACKAGES ***
	\usepackage[caption=false,font=normalsize,labelfont=sf,textfont=sf]{subfig}
\else
	\usepackage[caption=false,font=footnotesize]{subfig}
\fi
\usepackage{stfloats}	% provides the ability to put double column floats at the bottom of the page as well as the top.
\usepackage[inline]{enumitem}
\usepackage{xcolor}
\usepackage{soul}
\sethlcolor{pink}

\newcommand{\bbf}{\boldsymbol{f}}
\newcommand{\pbf}{\boldsymbol{f}^\prime}
\newcommand{\ppbf}{\boldsymbol{f}^{\prime \prime}}
\newcommand{\pppbf}{\boldsymbol{f}^{\prime \prime \prime}}
\newcommand{\bs}{\boldsymbol{s}}
\newcommand{\dbs}{\dot{\boldsymbol{s}}}
\newcommand{\ddbs}{\ddot{\boldsymbol{s}}}
\newcommand{\bx}{\boldsymbol{x}}
\newcommand{\by}{\boldsymbol{y}}
\newcommand{\bg}{\boldsymbol{g}}
\newcommand{\be}{\boldsymbol{e}}
\newcommand{\bsi}{\boldsymbol{\psi}}

\newcommand{\bq}{\boldsymbol{q}}

\newcommand{\ie}{\textit{i}.\textit{e}. }
\newcommand{\eg}{\textit{e}.\textit{g}. }

\newcommand{\PoincareBendixson}{Poincar\'{e}-Bendixson }

\DeclareMathOperator{\sat}{sat}
\newcommand{\dist}{\text{dist}}
\newcommand{\dis}{\text{d}}
\newcommand{\diag}{\text{diag}}

\theorembodyfont{\normalfont}
\theoremseparator{:}
\theoremheaderfont{\bfseries}

\theoremseparator{.}
\theoremheaderfont{\itshape\bfseries}
\newtheorem{thm}{Theorem}
\newtheorem{rmk}{Remark}

\begin{document}

\title{An Integrated Programmable CPG \\ with Bounded Output}


\author{Venus~Pasandi,~\IEEEmembership{Member,~IEEE,}
        Hamid~Sadeghian,
        Mehdi~Keshmiri,
        and~Daniele~Pucci,~\IEEEmembership{Member,~IEEE}% <-this % stops a space
\thanks{V. Pasandi and D. Pucci are with the Artificial and Mechanical Intelligence Laboratory, Italian Institute of Technology, Genova, Italy.
H. Sadeghian is with the Munich Institute of Robotics and Machine Intelligence, Technical University of Munich, Munich, Germany and the Engineering Department, University of Isfahan, Isfahan, Iran.
V. Pasandi and M. Keshmiri is with the Department of Mechanical Engineering, Isfahan University of Technology, Isfahan, Iran.
E-mails: venus.pasandi@iit.it, hamid.sadeghian@tum.de, h.sadeghian@eng.ui.ac.ir, mehdik@cc.iut.ac.ir, daniele.pucci@iit.it}}
\maketitle

\begin{abstract}
	Cyclic motions are fundamental patterns in robotic applications including industrial manipulation and legged robot locomotion.
	This paper proposes an approach for the online modulation of cyclic motions in robotic applications.
	For this purpose, we present an integrated programmable Central Pattern Generator (CPG) for the online generation of the reference joint trajectory of a robotic system out of  a library of desired periodic motions.
	The reference trajectory is then followed by the lower-level controller of the robot.
	The proposed CPG generates a smooth reference joint trajectory convergence to the desired one while preserving the position and velocity joint limits of the robot.
	The integrated programmable CPG consists of one novel bounded output programmable oscillator.
	We design the programmable oscillator for encoding the desired multidimensional periodic trajectory as a stable limit cycle.
	We also use the state transformation method to ensure that the oscillator's output and its first time derivative preserve the joint position and velocity limits of the robot.
	With the help of Lyapunov based arguments, We prove that the proposed CPG provides the global stability and convergence of the desired trajectory.
	The effectiveness of the proposed integrated CPG for trajectory generation is shown in a passive rehabilitation scenario on the Kuka iiwa robot arm, and also in walking simulation on a seven-link bipedal robot.
\end{abstract}


% \fi
\IEEEpeerreviewmaketitle

\section{Introduction}

\begin{figure}[!t]
	\centering
	\includegraphics[width=0.44\textwidth]{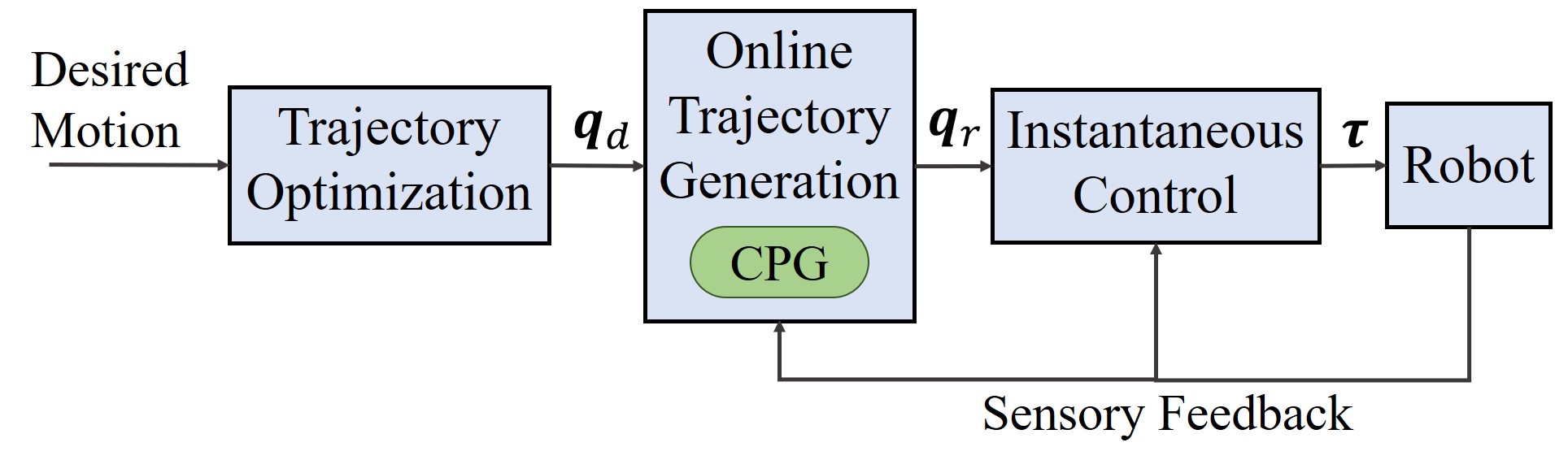}
	\caption{The modern control architecture for generating the cyclic motion in a robotic system.
	The desired joint trajectory, reference joint trajectory and actuator torques are denoted by $\boldsymbol{q}_d$, $\boldsymbol{q}_r$ and $\boldsymbol{\tau}$, respectively.}
	\label{fig:roboticControlArchitecture}
\end{figure}

\IEEEPARstart{C}{yclic} motions are fundamental patterns in robotic applications including walking, swimming, flying, crawling, rehabilitation, and pick and place tasks.
The modern control architectures for generating cyclic motions consist of three main loops; trajectory optimization, online trajectory generation, and instantaneous control (Fig. \ref{fig:roboticControlArchitecture}).
The trajectory optimization designs the desired periodic joint trajectory resulting in the desired cyclic motion.
Then, the online trajectory generation produces the reference joint trajectory converging to the desired one from any initial condition.
Finally, the reference trajectory is followed by the instantaneous joint control of the robot.
In this paper, we propose an integrated programmable Central Pattern Generator (CPG) with bounded output for online generation of a smooth reference joint trajectory from a library of the desired periodic ones.

The trajectory optimization for a desired cyclic motion is often complicated and computationally intensive because of the highly nonlinear coupled dynamics and high degree of freedom (DOF) of robotic systems.
Thus, the trajectory optimization techniques presented in the literature, are often offline~\cite{mohanan2018survey}.
In the modern control architecture, the trajectory optimization generates offline the desired joint trajectory considering the desired cyclic motion and the robot kinematic and dynamic features and limitations.
Then, the online trajectory generation adapts the desired trajectory to the current robot/environment conditions.
More specifically, the online trajectory generation provides the transition between the desired motions, abrupt response to the desired motion changing, and also adaptation to robot/environment with the help of sensory feedback~\cite{zhao2015trajectory}.

The CPG is widely used as an online reference trajectory generation algorithm for cyclic motions.
The CPG comprises a nonlinear dynamical system containing a globally stable periodic trajectory.
Assuming that the stable trajectory of the CPG matches the desired one, the CPG output converges to the desired trajectory from any initial condition, and thus the CPG generates a reference joint trajectory resulting in the desired motion for the robotic system from any initial condition~\cite{yu2014survey}.

The CPG is constructed based on periodic oscillators.
A periodic oscillator is a dynamical system providing a repetitive variation in time.
The periodic oscillators widely used for creating CPGs are limit cycle oscillators.
A limit cycle oscillator is a dynamical system with an Asymptotically Orbitally Stable (AOS) isolated periodic trajectory, called a stable limit cycle.
A multi-dimensional oscillator that the dimension of its output is equivalent to the robot DOF can be used as a CPG provided that the limit cycle of the oscillator matches the desired joint trajectory.
Moreover, a multiple of low-dimensional oscillators can be synchronized together for creating a multi-dimensional network of oscillators whose output dimension equals the robot DOF.
The synchronized network of oscillators can be used as a CPG if the multi-dimensional stable trajectory of the network matches the desired joint trajectory.

The well-known oscillators like Matsuoka, Hopf, Von der Pole, etc. can be used for creating a CPG.
In this case, the stable periodic trajectory of the CPG depends on the CPG parameters.
The CPG parameters include the parameters of the oscillator if a multi-dimensional oscillator is used, and include both the parameters of oscillators and synchronization connections if a synchronized network of oscillators is used.
The CPG parameters are tuned offline/online for adapting the stable periodic trajectory of the CPG to the desired one.
In this case, the CPG does not respond abruptly to the desired motion changing because of the tuning time.

For resolving the problem of parameter tuning, the CPG is constructed by programmable oscillators.
A programmable oscillator encodes the desired periodic trajectory via a stable limit cycle.
In mathematical terms, a programmable oscillator ensures AOS of the desired trajectory irrespective of the oscillator parameters.
In conclusion, the stable periodic trajectory of the CPG constructed by the programmable oscillators matches the desired trajectory without parameter tuning.
Various types of programmable oscillators have been proposed in the literature, like Dynamical Movement Primitives (DMP) \cite{ijspeert_dynamical_2013}, morphed phase oscillator \cite{ajallooeian_general_2013}, polynomial based oscillator \cite{okada_polynomial_2002}, alternation based oscillator \cite{ajallooeian_design_2012}, and modified data driven vector field oscillator \cite{pasandi2020programmable}.
The main limitations of the presented programmable oscillators are:
\begin{enumerate*}[label=(\roman*)]
	\item ensuring Asymptotic Stability (AS) instead of AOS of the desired trajectory.
	Though, providing AOS of the desired trajectory results in optimal motion tracking \cite{pasandi2020programmable}.
	\item ensuring the local convergence of the desired trajectory. In oscillators providing local convergence of the desired trajectory, the trajectory converges to	the desired one when the states are in the associated region of
	attraction.
	Thus, the desired trajectory cannot be changed at any time instant.
	Moreover, for the cases where the current desired trajectory is out of the region of attraction of the new one, additional logic is required for switching to the new desired trajectory,
	\item generating unbounded output. Indeed, respecting output limits ensures the integrity of the robot, and avoids the generation of fast motions which can be harmful and untraceable,
	\item not guaranteeing the permanent output synchronization. Using the asynchronous output of the oscillator as the joints reference trajectory of a robotic system results in an undesired robotic motion. For instance, in bipedal walking, the asynchronous reference trajectory disrupts the postural stability of the robot~\cite{pasandi2020programmable}.
\end{enumerate*}

This paper extends the previous work of the authors for designing a programmable CPG for robotic applications \cite{pasandi2020programmable} and proposes an integrated programmable CPG for online reference trajectory generation from a library of desired multidimensional periodic trajectories.
Comparing to the author's previous work, the proposed integrated CPG consists of a single bounded output multi dimensional programmable oscillator instead of a synchronized network of multiple low-dimensional ones.
Thanks to the integrated CPG, we
\begin{enumerate*}[label=(\roman*)]
	\item provide the possibility of ensuring both the AS and AOS of the desired trajectory,
	\item preserve the desired bound on the time derivative of the output of CPG,
	\item ensure the permanent output synchronization.
\end{enumerate*} 
For this purpose, the previously proposed data-driven vector field oscillator (DVO) for two dimensional trajectories \cite{pasandi2020programmable} is extended to a programmable oscillator for multidimensional ones.
The programmable oscillator ensures global AOS of the multidimensional desired trajectory irrespective of its parameters.
We also present a state transformation technique for generating a bounded output preserving the joint position and velocity limits of the robot.
The Lyapunov stability analysis shows that the proposed bounded output programmable oscillator ensures the global AOS of the desired trajectory and the desired output limitations.
In comparison with the existing CPGs, the contribution of the proposed CPG is four-folds.
First, the output of the proposed CPG is permanently synchronous.
Second, the proposed CPG is capable of ensuring both the AOS and AS of the desired trajectory.
Third, the proposed CPG ensures the global convergence of the desired trajectory.
Fourth, the output of the proposed CPG and its time derivative respects the desired limitations.
We have evaluated the performance of the proposed CPG through several simulations on a seven-link biped within walking scenario and experiments on the Kuka iiwa robot arm within a passive robotic rehabilitation scenario.

The rest of the paper is organized as follows.
Section \ref{sec:relaredWorks} briefly reviews the design of a programmable oscillator.
Section \ref{sec:notations} expresses the notations and definitions used in the paper.
In Section \ref{sec:programmableOscillator}, we present the programmable oscillator.
In Section \ref{sec:boundedOutputProgrammableOscillator}, we propose a state transformation technique for restricting the output of the programmable oscillator and its first time derivative.
Section \ref{sec:integratedCPG} describes the integrated CPG consisting of the bounded output programmable oscillator.
 The simulation and experiment results are illustrated in Section \ref{sec:validation}. 
The paper is finally concluded in Section \ref{sec:conclusion}.

\section{Related Work}
\label{sec:relaredWorks}

The online trajectory generation in the modern control architectures usually follows three main objectives, including
\begin{itemize}
	\item \textbf{smooth motion modulation (SMM)}
	ensures the smooth transition from the current state to the desired trajectory and transition from a desired trajectory to another one.
	The desired trajectory resulting in the desired steady state motion is designed by a trajectory optimization algorithm.
	\item \textbf{environment adaptation (EA)}
	ensures the adaptation of the reference trajectory to the environment in applications like obstacle avoidance and human-robot interaction.
	\item \textbf{robot adaptation (RA)}
	ensures the adaptation of the reference trajectory to the robot mechanic, kinematic and dynamic changes happened by purpose (\eg modular robots) and/or by some faults (\eg an actuator failure).
\end{itemize}

Moreover, an online trajectory generator algorithm can provide extra interesting features, including
\begin{itemize}
	\item \textbf{joints position limits maintenance (JPLM)}
	ensures the integrity of the robot.
	\item \textbf{joints speed limits maintenance (JVLM)}
	avoids fast motions which can be harmful and untraceable. 
	\item \textbf{low computational complexity (CC)}
	results in low computation time, and thus fast control reaction.
	\item \textbf{permanent trajectory synchronicity (PTS)}
	prevents the genration of an asynchronous reference trajectory that can generate undesired Cartesian motions like disrupting the robot balance in walking, swimming, and flying motions.
	\item \textbf{optimum trajectory generation (OTG)}
	ensures that the reference trajectory is optimal in the view of a desired optimality criteria like time, energy, and stability.
	\item \textbf{natural trajectory generation (NTG)}
	generates human acceptable motions.
	While, considering the least-action principle, a natural reference trajectory is the least-action-required trajectory for performing the desired trajectory.
\end{itemize}

\begin{table*}[!b]
	\centering
	\caption{Comparison of various online trajectory generators}
	\label{tab:compreToRelatedWorks}
	\begin{tabular}{|c|c|c|c|c|c|c|c|c|c|}
		\hline
		Method & SMM & EA & RA & JPLM & JVLM & CC & PTS & OTG & NTG \\\hline
		Time scaling methods & & \checkmark & & \checkmark & \checkmark & medium & \checkmark & \checkmark & \\\hline
		Trajectory optimization methods & \checkmark & \checkmark & \checkmark & \checkmark & \checkmark & high & \checkmark & \checkmark & \\\hline
		State space optimization methods & \checkmark & \checkmark & \checkmark & \checkmark & \checkmark & high & & \checkmark & \\\hline
		Central Pattern Generator & \checkmark & \checkmark & & \checkmark & & low & \checkmark & & \checkmark \\\hline
		Proposed Integrated CPG & \checkmark & & & \checkmark & \checkmark & low & \checkmark & & \checkmark \\\hline
	\end{tabular}
\end{table*}

	Various interesting methods have been proposed in the literature for the online trajectory generation for robotic systems.
	These methods can be categorized into four groups, namely
	\begin{enumerate}
		\item \textbf{time scaling methods} deforming the original timing law of the desired trajectory to provide environment adaptation and/or safe motion \cite{faroni2020real}.
		\item \textbf{trajectory optimization methods} design the optimum trajectory that mends the overall performance or reduces the consumption of the resources where the restricted system remains maintained \cite{almasri2021trajectory}.
		\item \textbf{state space optimization methods} search in the state space of the system for finding the optimum trajectory from the current state to the desired one \cite{kaushik2020fast,kim2011online}.
		\item \textbf{central pattern generator methods} \cite{yu2014survey}.
	\end{enumerate}
	\tablename{~\ref{tab:compreToRelatedWorks}} compares the above methods in terms of the objectives and features we defined at the beginning for a proper online trajectory generation algorithm.
	Note that each method of a group may address one or multiple of the features mentioned for that group.
	For example, there is no CPG proving SMM, EA, JPLM, PTS, and NTG together.
	The CPG method has generally two main interesting advantages to the other methods: low computational complexity and natural trajectory generation.
	Since the CPG is essentially a set of differential equations, its computational complexity is less than the other methods.
	Moreover, the CPG provides a natural trajectory since it ensures the AOS, instead of AS, of the desired trajectory.
	The natural trajectory results in natural and human acceptable motion.
	Besides that, the reduction of unnecessary stability constraints increases the energy and time efficiency of the system \cite{hobbelen2007limit}.
	In this paper, we present a general CPG framework that provides SMM, JPLM, JVLM, PTS, and NTG together.
	The proposed framework is also capable of being integrated into the sensory feedback for providing EA and RA.
	Our future work consists in proposing a technique for integrating the feedback information into the proposed CPG.
\section{Notations and Definitions}
\label{sec:notations}

\begin{itemize}[leftmargin=*]
\item $\mathbb{R}$ and $\mathbb{R}^+$ are the set of real and positive real numbers.
%\item $\mathcal{K}$ is the set of functions $\rho : \mathbb{R}^+ \rightarrow \mathbb{R}^+$ that are continuous, positive definite and strictly increasing.
%\item $\mathcal{K}_\infty$ is the set of functions $\rho : \mathbb{R}^+ \rightarrow \mathbb{R}^+$ that $\rho \in \mathcal{K}$ and $\lim\limits_{x\rightarrow \infty} \rho(x) = \infty$.
%\item $\mathcal{KL}$ is the set of functions $\rho : \mathbb{R}^+ \times \mathbb{R}^+ \rightarrow \mathbb{R}^+$ that for every $s \in \mathbb{R}^+$, $\rho(\cdot,s) \in \mathcal{K}$, $\rho(s,\cdot)$ is a continuous increasing function, and $\lim\limits_{y \rightarrow \infty} \rho (s,y) = 0$.
\item $I_n$ is the $n \times n$ identity matrix.
\item $\mathbf{0}_n$ is an $n$ dimensional zero column vector.
\item $\mathbf{1}_n$ is an $n$ dimensional column vector whose components are equal to one.
\item The $i^{th}$ component of a vector $\boldsymbol{q} \in \mathbb{R}^n$ is written as $q_i$.
\item The transpose operator is denoted by $(\cdot)^\top$.
\item The diagonal matrix of a vector $\boldsymbol{q} \in \mathbb{R}^n$ is denoted by $\breve{\boldsymbol{q}}$.
\item The Euclidean norm of $\boldsymbol{q} \in \mathbb{R}^n$ is denoted by $\|\boldsymbol{q}\|$.
\item The absolute value vector of $\boldsymbol{q} \in \mathbb{R}^n$ is denoted by $|\boldsymbol{q}|$.
\item For a vector $\bq \in \mathbb{R}^n$, the function $\tanh(q):\mathbb{R}^n \rightarrow \mathbb{R}^n$ is defined as $\tanh (q) = \left[\tanh(q_1),\tanh(q_2),...,\tanh(q_n)\right]^\top$.
\item Given a function $\boldsymbol{g}(x(t)):\mathbb{R} \rightarrow \mathbb{R}^n$ where $t$ represents the time, its first derivative w.r.t. $x$ is denoted as $\boldsymbol{g}^\prime= \frac{d\boldsymbol{g}}{dx}$ while its first derivative w.r.t. $t$ is denoted as $\dot{\boldsymbol{g}} = \frac{d\boldsymbol{g}}{dt}$.
\item  A \emph{$ \mathcal{C}^k $-continuous function} has $k$ continuous derivatives.
\item  A function ${\bbf(t):\mathbb{R} \rightarrow \mathbb{R}^n}$ is a \emph{$T$-periodic function} if for some constant $p$, we have ${\bbf(t+p) = \bbf(t)}$ $ \forall t$, and $T$ is the smallest positive $p$ with such property.
\end{itemize}
\section{Background}
\label{sec:background}

%%%%%%%%%%%%%%%%%%%%%%%%%%%%%%%%%%%%%%%%%%%%%%%%%%%%%%%%%%%%%%%%%%%%%%%%%%%%%%%%%%%%%%%%%%%%%%%%%%%%%%%%%%%%%%%%%%%%%%%%%%

\subsection{Stability of a Periodic Trajectory}
\label{sec:background_stability}

Given the dynamical system $\dot{\bx}=h(\bx)$, where $\bx \in \mathbb{R}^n$ is the state vector, the periodic trajectory $\boldsymbol{x}^*(t)$ of the system is
\begin{itemize}
	\item \textbf{stable} if $\forall \epsilon,t_0>0, \exists \delta>0$,
	\begin{equation*}
		\|\bx(t_0) - \bx^*(t_0) \| <\delta \Rightarrow \|\bx(t) - \bx^*(t) \| < \epsilon, \forall t>t_0
	\end{equation*}
	\item \textbf{asymptotically stable (AS)} if it is stable and $\exists\eta >0$,
	\begin{equation*}
		\|\bx(t_0) - \bx^*(t_0)\| <\eta \Rightarrow \lim\limits_{t\rightarrow\infty} \bx(t) = \bx^*(t), \forall t>t_0
	\end{equation*}
	\item \textbf{orbitally stable (OS)} if $\forall \epsilon,t_0 >0, \exists \delta >0$,
	\begin{equation*}
		\text{dist} \left( \bx(t_0),\bx^*(t) \right) <\eta \Rightarrow \text{dist} \left( \bx(t),\bx^*(t) \right) <\epsilon, \forall t>t_0
	\end{equation*}
	$\text{dist} \left( \bx(t),\bx^*(t) \right) = \inf_\tau \text{d}\left( \bx(t),\bx^*(\tau) \right)$ is the distance between $\bx(t)$ and the closed set of the trajectory $\bx^*(t)$ in the state space, while $\text{d}\left( \bx(t),\bx^*(\tau) \right)$ is the distance between two points $\bx(t)$ and $\bx^*(\tau)$ in the state space.
	\item \textbf{asymptotically orbitally stable (AOS)}, called also stable limit cycle, if it is orbitally stable and $\exists \eta >0$, \cite{hale1971functional}
	\begin{equation*}
	\begin{aligned}
		\text{dist} &\left( \bx(t_0),\bx^*(t) \right) <\eta \\
		&\Rightarrow \lim\limits_{t\rightarrow\infty} \text{dist} \left( \bx(t),\bx^*(t) \right) = 0, \forall t>t_0.
	\end{aligned}
	\end{equation*}
\end{itemize}

In the OS or AOS, the distance between the system’s states and the closed set of $\bx^*(t)$ is considered.
While in the stability or AS, the distance between the system’s states and a specific point of $\bx^*(t)$ changing over time is examined.
Therefore, stability/AS is a stricter condition than OS/AOS.
In the literature, providing AS and AOS of the desired trajectory is called trajectory tracking and limit cycle tracking, respectively.

%%%%%%%%%%%%%%%%%%%%%%%%%%%%%%%%%%%%%%%%%%%%%%%%%%%%%%%%%%%%%%%%%%%%%%%%%%%%%%%%%%%%%%%%%%%%%%%%%%%%%%%%%%%%%%%%%%%%%%%%%%%%

\subsection{Programmable Oscillators}

The programmable oscillator is a dynamical system ensuring the AOS of the desired periodic trajectory.
In this section, we review design of a programmable oscillator.

A programmable oscillator can be designed by constructing an autonomous dynamical system for the Lyapunov function describing the desired periodic trajectory~\cite{hirai_method_1972,green1984synthesis}.
%Constructing a non-autonomous dynamical system instead of an autonomous one, a programmable oscillator providing the desired transient motion is designed~\cite{ohno_synthesis_2006}. 
However, formulating the desired periodic trajectory via a Lyapunov function is not straightforward.

The programmable oscillator can be constructed by mapping a well-known oscillator, called the primary oscillator.
For this purpose, a phase-dependent function is first designed for mapping the limit cycle of the primary oscillator to the desired periodic trajectory.
The primary oscillator is then mapped through the designed phase-dependent function.
Thus, a general family of nonlinear phase oscillators generating any desired periodic trajectory is constructed~\cite{ajallooeian_general_2013}.
The phase of the mapped oscillator is an independent state, and hence the desired trajectory is AS and not AOS.
In other words, the mapped oscillator provides trajectory tracking and does not limit cycle tracking of the desired trajectory.
For assuring AOS of the desired trajectory, the primary oscillator can be altered by two state-dependent mapping functions. 
The mapping functions are determined by applying the \PoincareBendixson theorem for the predefined neighborhood of the desired trajectory~\cite{ajallooeian_design_2012}.
The altered oscillator guarantees local stability and convergence of the desired trajectory, and it is not straightforward to extend it to achieve global stability since the \PoincareBendixson theorem is no longer applicable.

The data-driven vector field method originally proposed for generating a discrete system with a desired stable limit cycle~\cite{hirai_synthesis_1982} is another method for designing a programmable oscillator.
The discrete vector field generated in the neighborhood of the desired limit cycle can be approximated by a continuous function, like a polynomial one, and a continuous dynamical system with a locally stable desired limit cycle is constructed~\cite{okada_polynomial_2002}.
To the best of our knowledge, the design of a continuous data-driven vector field for ensuring the global stability of the desired limit cycle has not been yet explored.

The programmable oscillator can be designed based on nonlinear control concepts. 
In this regard, the desired trajectory is parameterized by an exogenous variable called phase variable, and a dynamical system is designed for tracking the parametrized trajectory.
The phase variable is defined as a function of the states of the system such that the limit cycle tracking of the desired trajectory is guaranteed.
In this case, the desired trajectory is a two dimensional periodic trajectory depicting a non-self-intersecting curve in the state space~\cite{pasandi2019data}.
For tracking any two-dimensional desired trajectory, the phase variable is considered as an exogenous state and a dynamical system is also designed for the time evolution of the phase state.
The resulting system, called DVO, provides limit cycle tracking of any two dimensional desired periodic joint trajectory of the robotic system.
A synchronized network of DVOs can be used for tracking a multidimensional desired joint trajectory~\cite{pasandi2020programmable}.
However, such a network generates an asynchronous output when the desired trajectory changes.
The asynchronous reference trajectory results in undesired motions like the robot postural unstability in walking.
\section{Programmable Oscillator}
\label{sec:programmableOscillator}

Assume that the desired periodic trajectory of an $n$ degree of freedom robotic system in joint space is given by $\bq_d(t) = {\small\left[\bbf^\top(t), \dot{\bbf}^\top(t)\right]}$ where ${\boldsymbol{f}(t) \, : \, \mathbb{R}^+ \, \rightarrow \, \mathbb{R}^n}$  is an $n$-dimensional periodic function describing the desired time evolution of the joint positions.
In this section, we design an autonomous dynamical system providing limit cycle tracking of $\bq_d(t)$ irrespective of the system parameters.
The system consists of two coupled dynamical systems; shape dynamic and phase dynamic.
The shape dynamic is a $2n$ dimensional system providing global AS of $\bq_d(\varphi) = {\small\left[ \bbf^\top(\varphi), \bbf^{\prime \top}(\varphi) \right]^\top}$ that is the parameterized function of $\bq_d(t)$ w.r.t. $\varphi$, which is called the phase state.
The phase dynamic determines the time evolution of $\varphi$ for ensuring AOS of $\bq_d(t)$.

Consider the $2n+1$ dimensional autonomous system described by the following differential equations
\begin{subnumcases}{\label{eq:MDVO_dynamic}}
\ddbs = \ppbf (\varphi)- B \left( \dbs-\pbf (\varphi) \right)- K \left( \bs-\bbf   (\varphi) \right), \label{eq:MDVO_shapeDynamic}\\
\begin{aligned}
	\dot{ \varphi } = 1 + \gamma & \left( \left( \bs-\bbf(\varphi) \right)^\top K \pbf(\varphi) \right.\\
	& \left. \quad + \left( \dbs - \pbf(\varphi) \right)^\top \ppbf(\varphi) \right),
\end{aligned} \label{eq:MDVO_phaseDynamic}
\end{subnumcases}
where ${\small\left[ \bs^\top, \dbs^\top, \varphi \right]^\top} \in \mathbb{R}^{(n+n+1)}$ is the vector of states, $\dot{(\cdot)}$ and $(\cdot)^\prime$ denote the derivative w.r.t. $t$ and $\varphi$, respectively, and $B,K \in \mathbb{R}^{n \times n}$ and $\gamma \in \mathbb{R}$ are constant coefficients.
We call $\bx = {\small\left[ \bs^\top, \dbs^\top \right]^\top} \in \mathbb{R}^{2n}$ and $\left( \bs, \dbs \right)$ shape states vector and shape space of \eqref{eq:MDVO_shapeDynamic}, respectively.
The equation \eqref{eq:MDVO_shapeDynamic}, called shape dynamic, can guarantee that $\bx$ tracks $\bq_d(\varphi) = {\small \left[\bbf^\top(\varphi), \bbf^{\prime ^\top}(\varphi)\right]^\top}$.
Hence, $\bq_d(\varphi)$ is considered as an instantaneous target point.
The equation \eqref{eq:MDVO_phaseDynamic}, called phase dynamic, determines the time evolution of $\varphi$, and thus the instantaneous target point.
For $\gamma = 0$, the phase dynamics is simplified as $\dot{ \varphi }=1$, and thus $\varphi(t) = \varphi(0) + t$.
In this case, the dynamical system \eqref{eq:MDVO_dynamic} is equal to the classical proportional-derivative controller and provides the trajectory tracking of the desired trajectory.
For $\gamma \neq 0$, the phase state $\varphi$, and consequently the instantaneous target point $\bq_d(\varphi)$ depend on both time and the shape states.
In this case, the phase dynamics enables the system to compute the instantaneous target point as the closest point on the curve of $\bq_d(t)$ in the shape space to the shape states vector by an appropriate value of $\gamma$.
Therefore, such a system can provide the limit cycle tracking of the desired trajectory.
We call the closed-loop system of the shape dynamic and phase dynamic a programmable oscillator.
The following theorem describes properties of the programmable oscillator.

\begin{thm}
	\label{thm:stabilityOfProgrammableOscillator}
	For the programmable oscillator \eqref{eq:MDVO_dynamic}, consider
	\begin{enumerate}[label=A1.\arabic*)]
		\item the function $\bbf(t)$ is a periodic $\mathcal{C}^3$-continuous function,
		\item $B$ and $K$ are positive definite constant matrices, and
		\item the coefficient $\gamma$ is a nonnegative constant,
	\end{enumerate}
 the following results hold;
	\begin{enumerate}
		\item $\bx(t)$ is bounded and converges to $\bq_d (\varphi)$, $\forall \bx(0), \varphi(0)$;
		\item if $\gamma = 0$, then $\left[\bq_d^\top (t + \varphi (0)), t + \varphi (0)\right]^\top$ is globally AS;
		\item if $\gamma \rightarrow \infty$, then $\bq_d (t)$ is a globally stable limit cycle.
	\end{enumerate}
\end{thm}

The proof is given in Appendix \ref{app:MDVOstability}.
Theorem \ref{thm:stabilityOfProgrammableOscillator} explains that the shape states vector $\bx$ converges to the vector of the instantaneous target point $\bq_d(\varphi)$ from any initial condition.
As a result, $\bx(t)$ converges to the curve of $\bq_d(t)$ in the shape space from any initial condition.
Moreover, if $\gamma = 0$, then $\bx(t)$ converges to $\bq_d(t+\varphi(0))$.
If $\gamma \rightarrow \infty$, the programmable oscillator ensures limit cycle tracking of ${\bq_d(t)}$, \ie ${\mathcal{P}_{\bbf} = \{\bx : \exists \varphi, \, \bx = \bq_d(\varphi)\}}$ is an attracting invariant set.

\begin{rmk}
	For the programmable oscillator, the desired joint trajectory (\ie $\bq_d(t)$) can be periodic or constant.
	For a constant desired joint trajectory, the programmable oscillator is simplified to a linear autonomous system as the following
	\begin{equation}
	\begin{cases}
	\ddbs = -B \dbs -K \left( \bs-\bbf \right), \\
	\dot{ \varphi } = 1.
	\end{cases}
	\end{equation}
	where $\left[\bbf^\top, \mathbf{0}_n^\top,t+\varphi(0)\right]^\top$ is the AS equilibrium point. 
\end{rmk}
\section{Bounded Output Programmable Oscillator}
\label{sec:boundedOutputProgrammableOscillator}

Using the programmable oscillator, we can generate a smooth reference trajectory for tracking the desired periodic joint trajectory from any initial condition.
The reference trajectory is further used in the robot controller for generating the desired cyclic motion (\figurename{\ref{fig:roboticControlArchitecture}}).
However, the reference trajectory can be followed by the robot if it respects physical limits including position and velocity joint limits of the robot.
In this section, the dynamic of the programmable oscillator is modified for producing a bounded output preserving the predefined limits for the output and its first time derivative.

Assume the constraints for the output of the oscillator as
\begin{equation}
\label{eq:EDVO_outputConstrained}
\mathcal{Q} := \left\{ \by \in \mathbb{R}^n \, : \, \by_{min} < \by < \by_{max}, \ |\dot{\by}| < \delta_{\dot{\by}} \right\},
\end{equation}
where $\by_{min}, \by_{max} \in \mathbb{R}^n$ are vectors of the minimum and maximum feasible value of the output, respectively, and ${\delta_{\dot{\by}} \in \mathbb{R}^n}$ is the vector of the maximum feasible rate of change of output.

To generate a feasible output, the output of the bounded output programmable oscillator is defined as \cite{charbonneau2016line}
\begin{equation}
\label{eq:EDVO_outputDefinition}
\by = \by_{avg} + \breve{\delta}_{\by} \tanh{(\bs_1)},
\end{equation}
where $\by_{avg} = \frac{\by_{max}+\by_{min}}{2}$, $\breve{\cdot} = \diag(\cdot)$, ${\delta_{\by} = \frac{\by_{max}-\by_{min}}{2}}$, and $\tanh(\cdot): \mathbb{R}^n \rightarrow \mathbb{R}^n$.
$\bx = \left[\bs_1^\top, \bs_2^\top\right]^\top \in \mathbb{R}^{2n}$ is the shape state vector of the bounded output programmable oscillator.
If $\bs_1$ is bounded, the above equation guarantees that ${\by_{min} = \by_{avg} - \delta_{\by} < \by < \by_{avg} + \delta_{\by} = \by_{max}}$.

To preserve the constraints on the rate of change of the output, we define the dynamics of $\bs_1$ as
\begin{equation}
\label{eq:EDVO_extraDynamics}
J_{\bs_1} \dot{\bs}_1 = \breve{\delta}_{\dot{\by}} \tanh{(\bs_2)},
\end{equation}
where $J_{\bs_1} = \breve{\delta}_{\by} \left( I_n-\tanh^2{\left( \breve{\bs}_1 \right)} \right)$.
According to \eqref{eq:EDVO_outputDefinition} and \eqref{eq:EDVO_extraDynamics}, the time derivative of the output is as
\begin{equation}
\label{eq:BMDVO_outputRateDefinition}
\dot{\by} = J_{\bs_1} \dot{\bs}_1 = \breve{\delta}_{\dot{\by}} \tanh{( \bs_2 )}.
\end{equation}
Thus, if $\bs_2$ is bounded, then $\ |\dot{\by}| < \delta_{\dot{\by}}$.
In other words, if the shape state vector of the bounded output programmable oscillator is bounded, we have $\by \in \mathcal{Q}$.

Assuming that ${\bbf(\varphi): \mathbb{R} \rightarrow \mathbb{R}^n}$ is the desired value of $\by$, one can compute the desired value of the shape state vector $\bx = \left[\bs_1^\top,\bs_2^\top\right]^\top$ denoted by $\bx^* (\varphi) = \left[\bg_p^\top (\varphi), \bg_v^\top (\varphi)\right]^\top$ using the inverse of \eqref{eq:EDVO_outputDefinition} and \eqref{eq:BMDVO_outputRateDefinition} by
\begin{equation}
\label{eq:BMDVO_stateTransformation}
\begin{aligned}
\bg_p (\varphi) &= \tanh^{-1}{\left( \breve{\delta}_{\by}^{-1} \left( \bbf (\varphi) - \by_{avg} \right) \right)}, \\[4pt]
\bg_v (\varphi) &= \tanh^{-1}{\left( \breve{\delta}_{\dot{\by}}^{-1} \pbf (\varphi) \right)}.
\end{aligned}
\end{equation}
Thus, if $\bx$ is bounded and converges to the curve of $\bx^*$ in the shape space, then $\by$ converges to the curve of $\bq_d(t) = {\small\left[\bbf^\top(t), \dot{\bbf}^\top(t)\right]^\top}$ in the space $\left(\by,\dot{\by}\right)$ while preserving the output limitations described in \eqref{eq:EDVO_outputConstrained}.
For this purpose, we rewrite the dynamic of the programmable oscillator for the state vector $\left[\bs_1^\top,\bs_2^\top, \varphi\right]^\top$ while considering the constraint \eqref{eq:BMDVO_outputRateDefinition} as
\begin{subnumcases}{\label{eq:BMDVO_Dynamics}}
J_{\bs_1} \dot{\bs}_1 = \breve{\delta}_{\dot{\by}} \tanh{(s_2)}, \\[5pt]
\begin{aligned}
\dot{\bs}_2 = \, &\bsi_\varphi + \bsi_t - B \be_1 - K \be_2 \\
&- B^{-1}D J_{\bs_1}^{-1} \breve{\delta}_{\dot{\by}} \left( \tanh{(\bs_2)} - \tanh{(\bsi)} \right),
\end{aligned}\\[5pt]
\begin{aligned}
\dot{ \varphi } = 1 + \gamma \Big( \left( \be_1^\top D + \be_2^\top B \right) J_{\bg_p}^{-1} \breve{\delta}_{\dot{\by}} \tanh{(\bg_v)}\\
+ \left( \be_1^\top B + \be_2^\top K \right) \bsi_\varphi \Big), 
\end{aligned} \label{eq:BMDVO_phaseDynamic}\\
\by = \by_{avg} + \breve{\delta}_{\by} \tanh{(\bs_1)},
\end{subnumcases}
where,
\begin{align}
&\be_1 = \bs_1 - \bg_p, \\
&\be_2 = \bs_2 - \bsi, \\
&\bsi = \tanh^{-1} \left( J_{\bs_1} J_{\bg_p}^{-1} \tanh{(\bg_v)} \right),\label{eq:BMDVO_psiDefinition} \\
&J_{\bg_p} = \breve{\delta}_{\by} \left( I_n - \tanh^2{\left( \breve{\bg}_p \right)} \right),
\end{align}
with $B, K, D \in \mathbb{R}^{n\times n}$ and $\gamma \in \mathbb{R}$ are constant coefficients. Moreover,
$\bsi_\varphi$ and $\bsi_t$ are the partial derivatives of $\bsi$ w.r.t. $t$ and $\varphi$ respectively, and are computed as
\begin{equation}
{\footnotesize
\begin{aligned}
&\bsi_\varphi = J_{\bsi}^{-1} J_{\bs_1} J_{\bg_p}^{-1} \left( 2 \breve{\delta}_{\dot{\by}}^2 J_{\bg_p}^{-1}\tanh^2{(\breve{\bg}_v)} \tanh{(\bg_p)} + J_{\bg_v} \bg_a\right), \\
&\bsi_t = -2 J_{\bsi}^{-1} \breve{\delta}_{\dot{\by}}^2 \tanh{(\breve{\bs}_1)} \tanh{(\bs_2)},
\end{aligned}}
\end{equation}
where
\begin{equation}
\begin{aligned}
&J_{\bsi} = \breve{\delta}_{\dot{\by}} \left( I_n - \tanh^2{( \breve{\bsi})} \right),\\ &{J_{\bg_v} = \breve{\delta}_{\dot{\by}} \left( I_n - \tanh^2{\left( \breve{\bg}_v \right)} \right)},
\end{aligned}
\end{equation}
and
\begin{equation}
\label{eq:BMDVO_gaDef}
\begin{aligned}
\bg_a = \bg_v^\prime = J_{\bg_v}^{-1} \ppbf (\varphi).
\end{aligned}
\end{equation}
In the following theorem and remarks, we describe the conditions ensuring the existence of $\bsi$, $J_{\bs_1}$, $J_{\bg_p}$, $J_{\bg_v}$ and $J_{\bsi}$.
\begin{thm}
\label{thm:psiDefinable}
	If $\bbf(t)$ satisfies
	\begin{equation}
	\label{eq:BMDVO_constrainedForPsi}
	{\small
	\breve{\delta}_{\by}^2 \left| \pbf (\varphi) \right| \leq \left| \breve{\delta}_{\by}^2 - \left( \breve{\bbf} (\varphi) - \breve{\by}_{avg} \right)^2\right| \delta_{\dot{\by}}, \quad \forall \varphi \in \left[0,T\right],} %\left| J_{\bg_p}^{-1} \tanh{\left( \bg_v \right)}\right| \le \mathbb{1}_n,
	\end{equation}
	where $T \in \mathbb{R}^+$ is the period of $\bbf(t)$, then the variable $\bsi$ described in \eqref{eq:BMDVO_psiDefinition}, is well-defined.
\end{thm}

\begin{figure*}[ht]
	\centering
	\includegraphics[width = 0.97\textwidth]{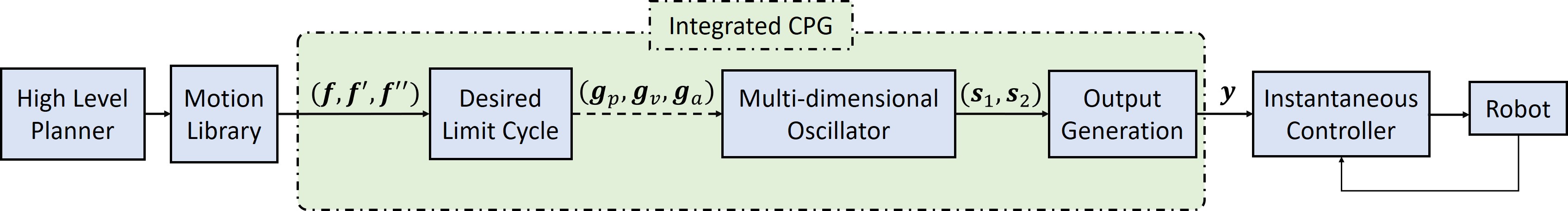}
	\caption{The proposed integrated CPG that consists of a bounded output programmable oscillator for generating the reference trajectory of a robotic system.}
	\label{fig:integratedCPG}
\end{figure*}

The proof is given in Appendix \ref{app:psiDefinable}.
According to \eqref{eq:EDVO_outputConstrained}, if the function $\bbf(t)$ is a feasible output for the robotic system, then $\bbf(\varphi)$ satisfies $|\bbf(\varphi)-\by_{avg}|\leq \delta_{\by}$ and $|\pbf(\varphi)|\leq \delta_{\dot{y}}$ and not necessarily \eqref{eq:BMDVO_constrainedForPsi}.
Therefore, condition \eqref{eq:BMDVO_constrainedForPsi} restricts the set of feasible desired trajectories that can be generated by the programmable oscillator.
In Section \ref{sec:integratedCPG}, this condition is relaxed to provide the possibility of the generation of any feasible desired trajectory through the programmable oscillator.

\begin{rmk}
	$J_{\bg_p}$ and $J_{\bg_v}$ are diagonal positive semi definite (PSD) matrices.
	If $\bbf(t) \in \mathcal{Q}$, then $J_{\bg_p}$ and $J_{\bg_v}$ are positive definite (PD), and thus invertible.
\end{rmk}

\begin{rmk}
	$J_{\bs_1}$ is a diagonal PSD matrix.
	If $\bs_1$ is bounded, then $J_{\bs_1}$ is PD, and thus invertible.
\end{rmk}

\begin{rmk}
	$J_{\bsi}$ is a diagonal PSD matrix.
	If $\bbf(t)$ satisfies \eqref{eq:BMDVO_constrainedForPsi}, then $J_{\bsi}$ is PD, and thus invertible.
\end{rmk}

We call the dynamical system presented in \eqref{eq:BMDVO_Dynamics}, bounded output programmable oscillator.
The following theorem concludes the properties of such an oscillator.

\begin{thm}
\label{thm:BMDVO}
	For the bounded output programmable oscillator presented in \eqref{eq:BMDVO_Dynamics}, if
	\begin{enumerate}[label=A3.\arabic*)]
		\item $\bbf(t)$ is a periodic $\mathcal{C}^3$-continuous function satisfying \eqref{eq:BMDVO_constrainedForPsi},
		\item $B$, $K$, and $D$ are PD diagonal constant matrices, where $KB^{-1}D-B$ is also PD, and
		\item the coefficient $\gamma$ is a nonnegative constant,
	\end{enumerate}
then the following results hold;
\begin{enumerate}
	\item $\bx(t)$ is bounded and converges to $\bx^*(\varphi)$, $\forall \bx(0), \varphi(0)$;
	\item if $\gamma = 0$, $\left[\bx^{* \top}(t+\varphi(0)),t+\varphi(0)\right]^\top$ is globally AS;
	\item if $\gamma \rightarrow \infty$, $\bx^*(t)$ is the globally stable limit cycle, and
	\item the output $\by(t)$ converges to the curve of $\bq_d(t)$ in the space $\left(\by,\dot{ \by }\right)$ while $\by(t) \in \mathcal{Q}$.
\end{enumerate}
\end{thm}

The proof is given in Appendix \ref{app:BMDVO}.
According to the above theorem, the shape state vector is bounded and converges to $\bx^*$ from any initial condition.
Thus, \eqref{eq:BMDVO_phaseDynamic} converges to $\dot{ \varphi } = 1$.
Therefore, the output $\by(t)$ converges to the curve of $\bq_d(t)$ in $\left(\by, \dot{ \by }\right)$ while $\by(t) \in \mathcal{Q}$.
Moreover, if $\gamma = 0$, the output $\by(t)$ converges to $\bbf(t+\varphi(0))$.
If $\gamma \rightarrow \infty$, then the bounded output programmable oscillator ensures the global limit cycle tracking of $\bq_d(t)$.
We call the cases where $\gamma = 0$ and $\gamma \gg 1$, as the AS and AOS modes, respectively.

\section{Integrated Central Pattern Generator}
\label{sec:integratedCPG}

In this section, we introduce the integrated CPG consisting of one bounded output programmable oscillator.
The integrated CPG generates an $n$ dimensional signal $\by$ tracking the $n$ dimensional desired signal $\bbf(t)$ while preserving the predefined limits for $\by$ and $\dot{ \by }$.

The control architecture for a multi degree of freedom robotic system where the reference joint trajectory is generated online using the integrated CPG is depicted in \figurename{\ref{fig:integratedCPG}}.
The high level planner determines the desired motion of the robot, and the motion library provides the desired joint trajectory (\ie $\bq_d(t)$) resulting in the desired motion.
The motion library is a set of the desired periodic joint trajectories designed offline for generating the desired cyclic motions.
The desired trajectory is used for creating the structure of the integrated CPG.
Using~\eqref{eq:BMDVO_stateTransformation}, the desired trajectory of the shape state vector of the integrated CPG (\ie $(\bg_p(t),\bg_v(t))$) is computed according to the desired trajectory.
After that, the reference joint trajectory is generated by integrating the differential equation of the bounded output programmable given in~\eqref{eq:BMDVO_Dynamics}.
The initial condition of the shape states vector of the bounded output programmable oscillator is determined based on the initial condition of the robot by using \eqref{eq:EDVO_outputDefinition} and \eqref{eq:BMDVO_outputRateDefinition}.
The initial condition of the phase state is arbitrary.

\begin{figure}[!t]\vspace*{-0.3cm}
	\centering
	\includegraphics[width=5.0cm]{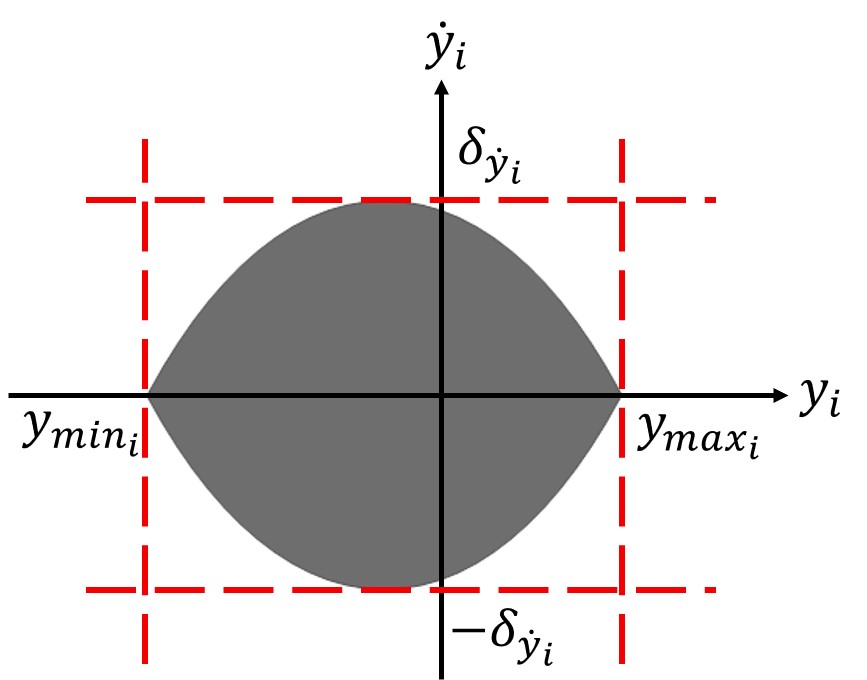}
	\caption{Schematic of the acceptable region for $\bbf_i$ where \eqref{eq:BMDVO_constrainedForPsi} is satisfied. The red dashed lines denote the limitations for $\by_i$ and $\dot{\by}_i$.
		The gray region depicts the acceptable region.}
	\label{fig:BMDVO_acceptableDesiredFunction}
\end{figure}

\begin{figure}[!h]
	\centering
	\includegraphics[width=6.3cm]{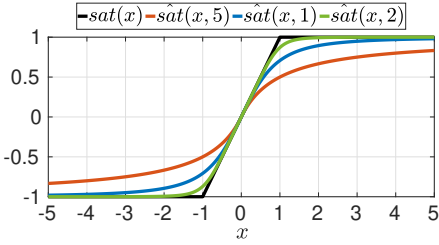}
	\caption{The behavior of $\hat{\sat}(x,p)$ for different values of $p$ w.r.t. $\sat(x)$.}
	\label{fig:BMDVO_saturationEstimation}
\end{figure}

According to Theorem \ref{thm:BMDVO}, the integrated CPG generates the desired joint trajectories satisfying \eqref{eq:BMDVO_constrainedForPsi}.
Considering ${\bq_d : \mathbb{R} \rightarrow \mathbb{R}^{2n}}$ as the desired joint trajectory, the schematic of the acceptable region for ${\small\left[\bbf_i(t),\dot{\bbf}_i(t)\right]}$ where \eqref{eq:BMDVO_constrainedForPsi} is satisfied is depicted in \figurename{\ref{fig:BMDVO_acceptableDesiredFunction}}.
As shown, the acceptable region for ${\small\left[\bbf_i(t),\dot{\bbf}_i(t)\right]}$ (the gray area) is less than the feasible region for ${\small\left[\by_i,\dot{\by}_i\right]}$ presented in \eqref{eq:EDVO_outputConstrained}.
Thus, the set of acceptable desired joint trajectories is less than the set of feasible ones ensuring the output limitations.
For generating any feasible desired trajectory, we modify the definition of $\bsi$ as the following
\begin{equation}
\bsi = \tanh^{-1} \left( \hat{\sat} \left( J_{\bs_1} J_{\bg_p}^{-1} \tanh{(\bg_v)} , p \right) \right),
\label{eq:BMDVO_psiDefinitionModify2}
\end{equation}
where $\hat{\sat}(x,p)$ is the continuous estimation of the saturation function $\sat(x)$ and is defined as
\begin{equation}
\hat{\sat}(x,p) = \dfrac{x}{\sqrt[p]{1+|x|^p}}.
\end{equation}
$p \in \mathbb{R}^+$ is a constant.
The behavior of $\hat{\sat}(x,p)$ for different values of $p$ in comparison with $\sat(x)$ is depicted in Fig. \ref{fig:BMDVO_saturationEstimation}.
As can be seen, we have $\lim\limits_{p \rightarrow \infty} \hat{\sat}(x,p) = \sat(x)$.

\section{Validation}
\label{sec:validation}

In this section, the performance of the proposed integrated CPG for reference trajectory generation is evaluated on two main scenarios; manipulation and bipedal walking. 
In the first case, the integrated CPG is employed empirically on Kuka iiwa robot arm for passive robotic rehabilitation scenario.
Later, the integrated CPG is used to generate a walking pattern for a seven-link bipedal robot in a simulation study.
%In both experiments and simulations, the integrated CPG generates the reference joint trajectory followed by an instantaneous controller where the instantaneous controller is a position control.
%Precisely, the reference joint trajectory is given to a PID controller ensuring the reference trajectory tracking by the robot.

%%%%%%%%%%%%%%%%%%%%%%%%%%%%%%%%%%%%%%%%%%%%%%%%%%%%%%%%%%%%%%%%%%%%%%%%%%%%%%%%%%%%%%%%%%%%%%%%%%%%%%%%%%%%%%%%%%%%%%%%%%%%%%%%%%%%%%%%%%%%%%%%%%

\subsection{Passive Robotic Rehabilitation Scenario}

\begin{figure}[ht]
	\centering
	\includegraphics[width=0.35\textwidth]{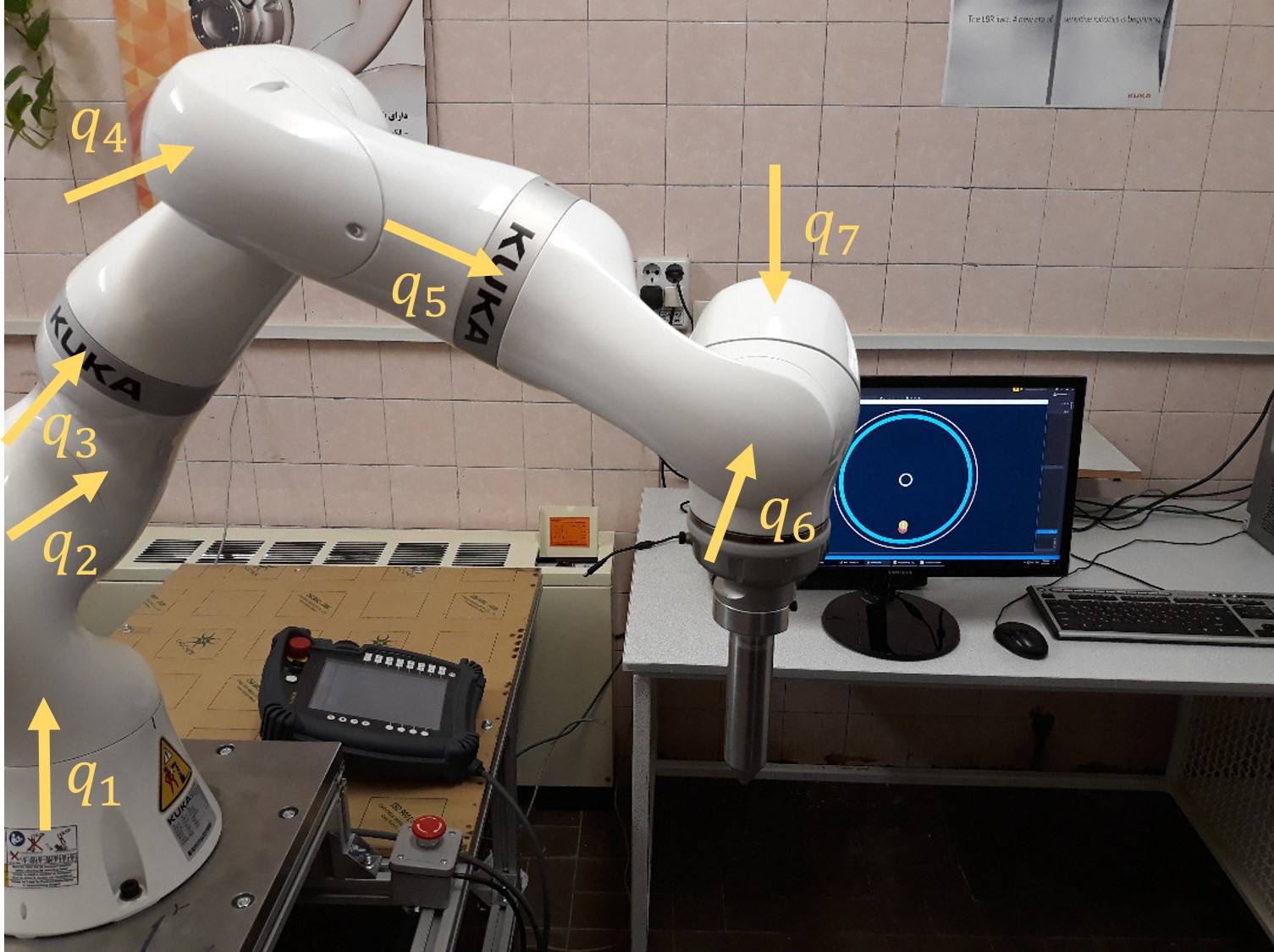}
	\caption{
		The Kuka arm's joints used in the passive rehabilitation experiment.
	}
	\label{fig:kukaIiwaArm}
\end{figure}

\begin{figure*}[!t]
	\centering
	\includegraphics[width=0.9\textwidth]{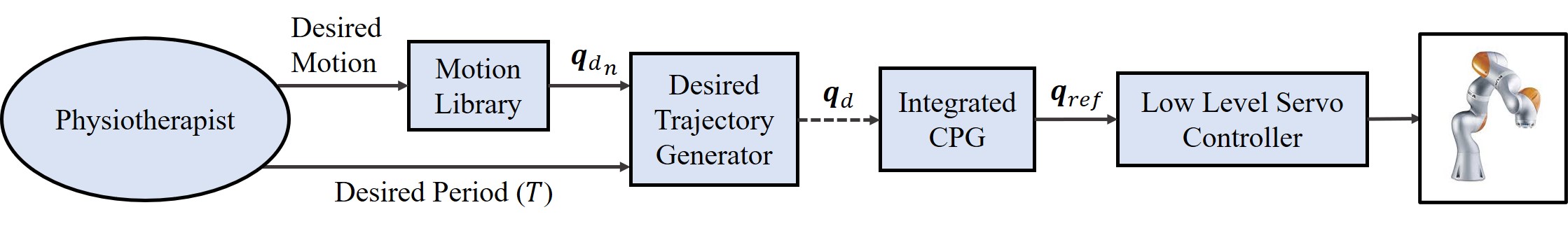}
	\caption{
		The control architecture employed for upper-limb passive robotic rehabilitation scenario.
	}
	\label{fig:kuka_controlArchitecture}
\end{figure*}

In this study, the integrated CPG is employed on Kuka iiwa robot arm shown in \figurename~\ref{fig:kukaIiwaArm} for passive robotic rehabilitation.
For upper-limb passive robotic rehabilitation, the control architecture  consists of three main elements; a task-level planner, a joint space planner, and a low-level controller.
%The task-level controller designs the reference joint trajectory of the robot generating the desired motion.
The desired movements of the end-effector of the robot are first designed by the physiotherapist in the task space.
The desired movements may include repetitive motions on circles, lines, or general paths.
The joint space planner comprises a motion library and the integrated CPG. 
The desired movements are reflected in the joint space and are gathered in a motion library.  
%The high-level controller generates the reference motor force/torque ensuring the reference joint trajectory tracking.
%Finally, the low-level part controls the motors of the robot for generating the reference force/torque.
%The integrated CPG with a motion library as the task-level controller in the control architecture of the Kuka iiwa arm for passive arm rehabilitation task (\figurename{\ref{fig:kukaIiwaArm}}).
%We used proportional-derivative and proportional-integrator-derivative controllers as the high-level and low-level controllers for controlling the seven joints of Kuka arm during the passive rehabilitation.
The proposed integrated CPG is then exploited to produce the reference joint trajectory of the robot based on the desired trajectories in the motion library.
The CPG is responsible for switching between the desired motions as well as changing the tempo rate of the cyclic motions. 
The block diagram of the control architecture is depicted in \figurename{\ref{fig:kuka_controlArchitecture}}.

%For the integrated CPG, the desired trajectory is read from the library of the desired trajectories, called motion library, according to the desired motion chosen by the physiotherapist.
%In this experiment, the library contains two periodic and one constant 14 dimensional joint trajectories.
%The periodic desired trajectories generate a cyclic motion where the end-effector of the robot moves on the horizontal and vertical circles with the radius of $10 (cm)$.
Two vertical and horizontal circles with a radius of $10 (cm)$ in the task space of the robot are defined as the desired motions. The origin of the vertical circle is also used as starting and ending configuration. 
%The constant trajectory postures the robot in a fix configuration where the end-effector locates at the origin of the vertical circle.
The default period of the cyclic motions is $10 (sec)$.
For modulation of the period of the cyclic motions, the desired joint trajectory of the robot is defined as
\begin{equation}
\begin{aligned}
&\bbf(\varphi) = \bbf_n \left( T\varphi/10 \right),\\
&\pbf(\varphi) = \left(T/10\right)\pbf_n \left( T\varphi / 10 \right),\\
&\ppbf(\varphi) = \left(T / 10\right)^2\ppbf_n \left( T\varphi / 10 \right),
\end{aligned}
\end{equation}
where the nominal desired trajectory $\bq_{d_n}(t) = {\small\left[\bbf_n(t),\dot{\bbf_n}(t)\right]^\top}$, is the desired joint trajectory which is read from the motion library, and ${T \in \mathbb{R}^+}$ is the desired period of the desired motion.
In this way, the physiotherapist can choose the desired motion from the set of desired motions provided by the motion library (\ie horizontal/vertical circles and constant configurations) with any arbitrary time period.
However, the desired joint trajectory $\bq_d(t)$ can be applied to the integrated CPG if it preserves the position and velocity limits defined for the CPG output.
Since, $\bbf_n(t)$ considers the position limits, $\bbf(t)$ preserves the position limits as well.
But, $\dot{\bbf}(t)$ preserves the velocity limits if the period of the desired motion satisfies $T < \frac{10 \min_i \delta_{\dot{\by}_i}}{\max_i \max_{\varphi}\pbf_{n_i}(\varphi)}$.

\begin{figure}[!b]
	\centering
	\subfloat{
		\includegraphics[width=0.24\textwidth]{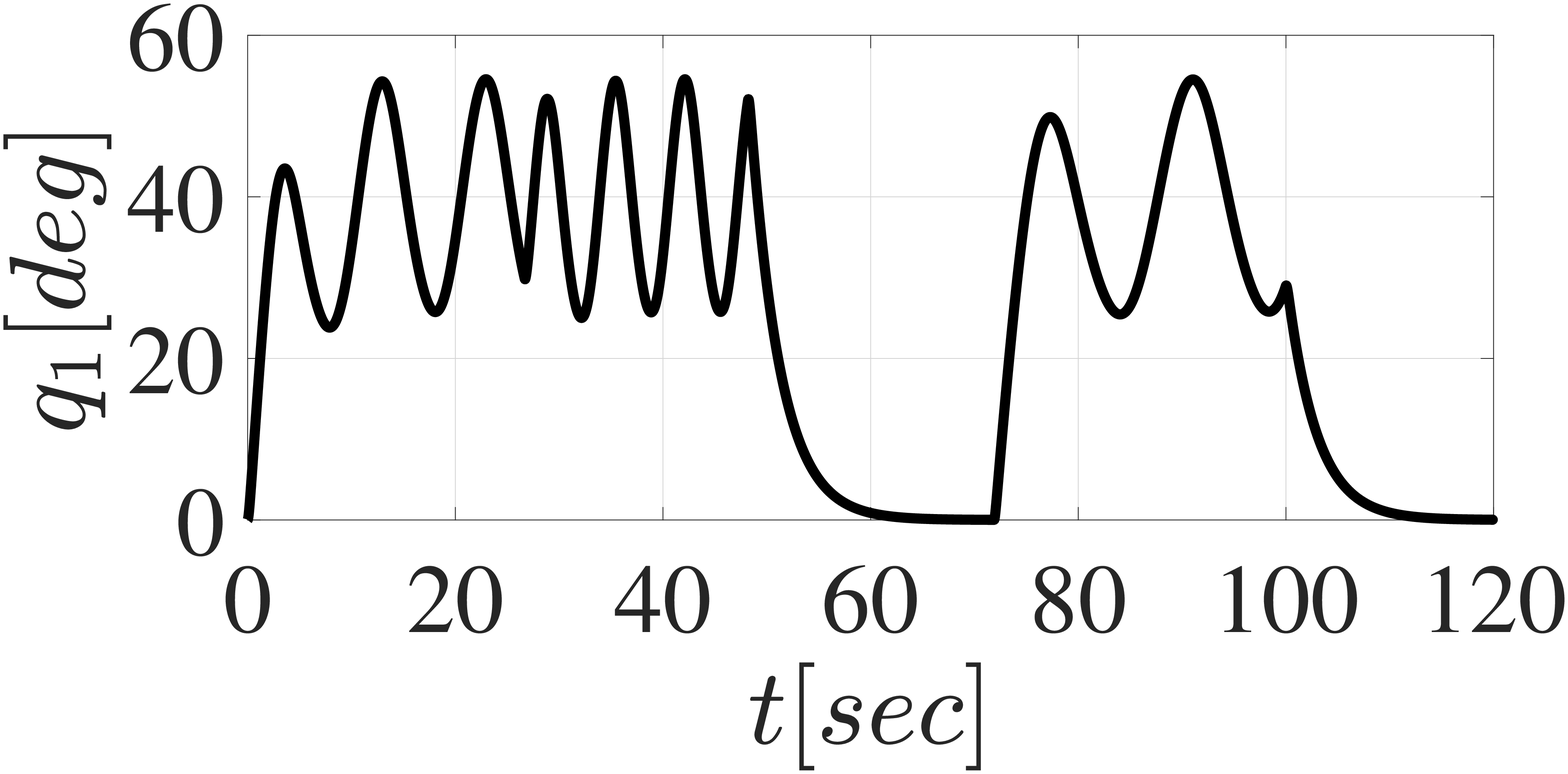}
		\hfill
		\includegraphics[width=0.24\textwidth]{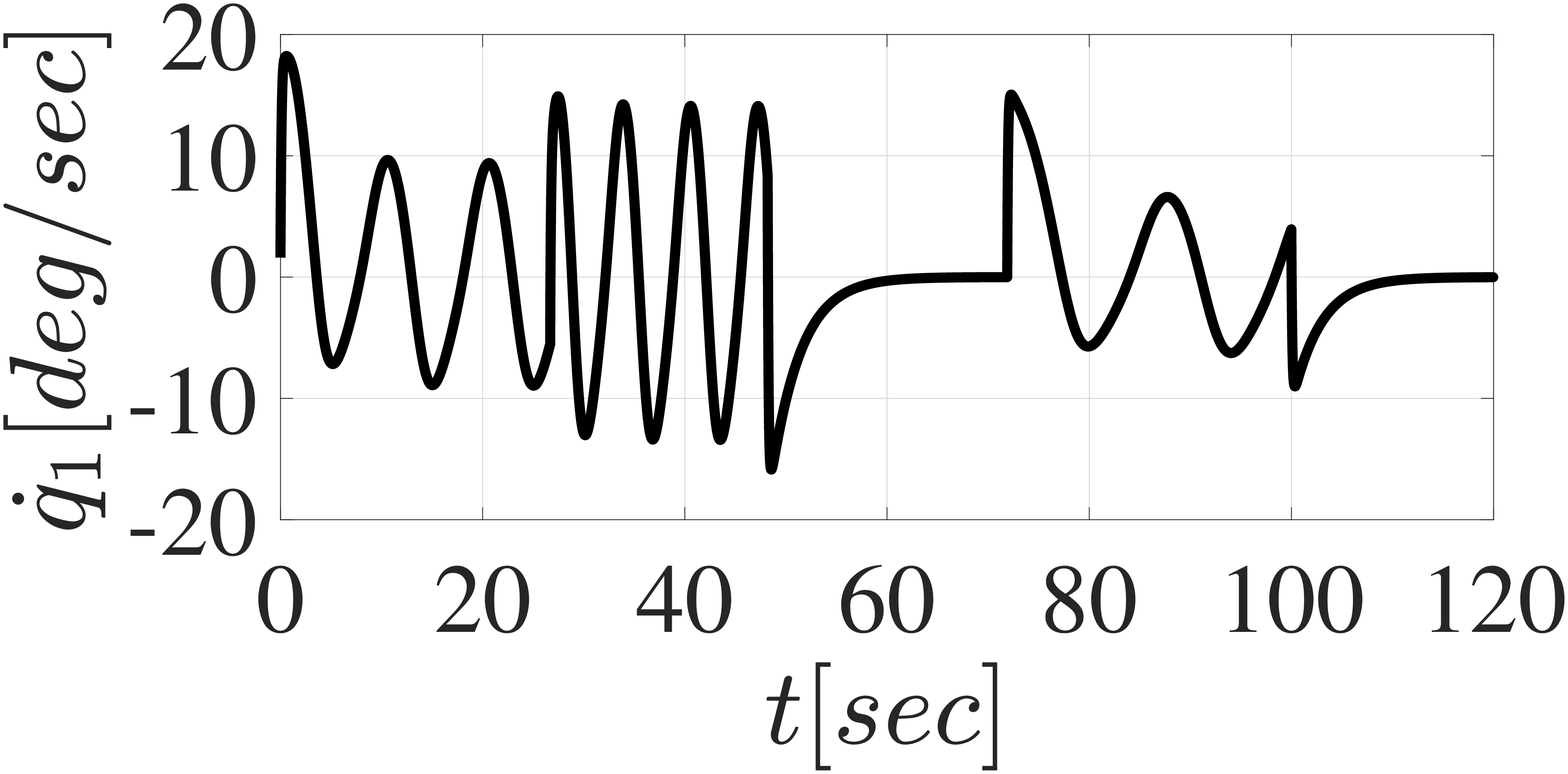}
	}\\
	\subfloat{
		\includegraphics[width=0.24\textwidth]{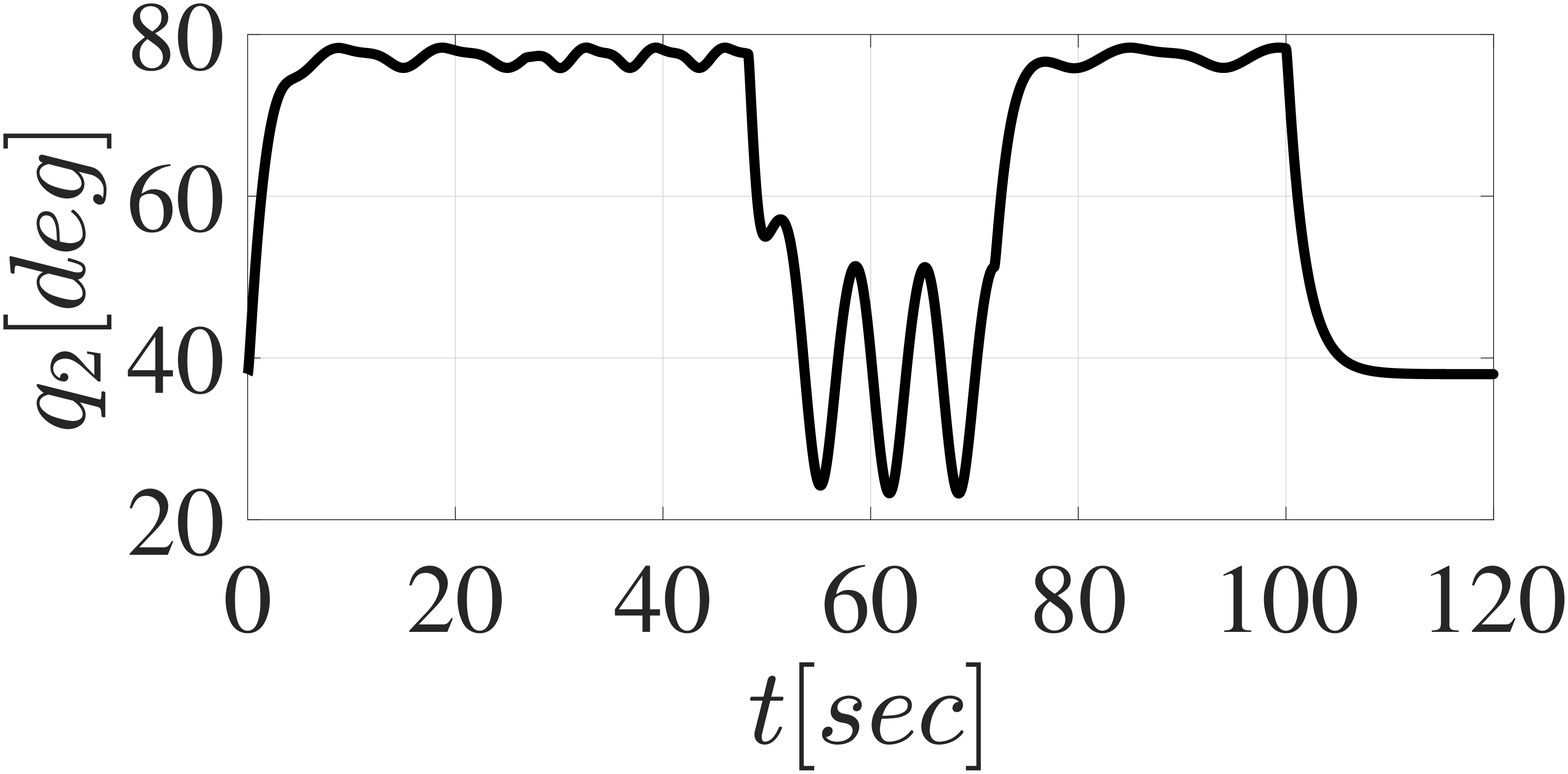}
		\hfill
		\includegraphics[width=0.24\textwidth]{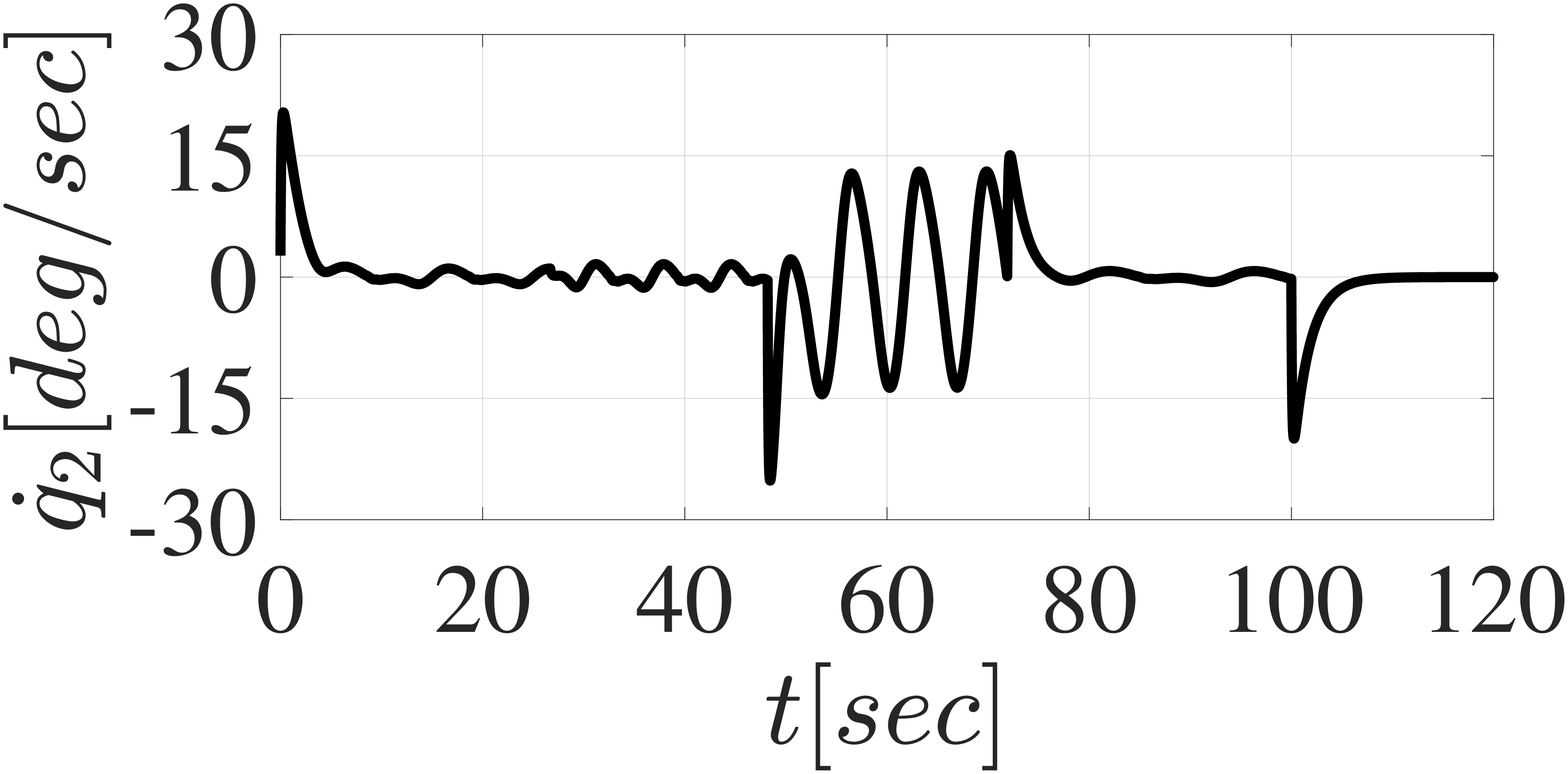}
	}\\
	\subfloat{
		\includegraphics[width=0.24\textwidth]{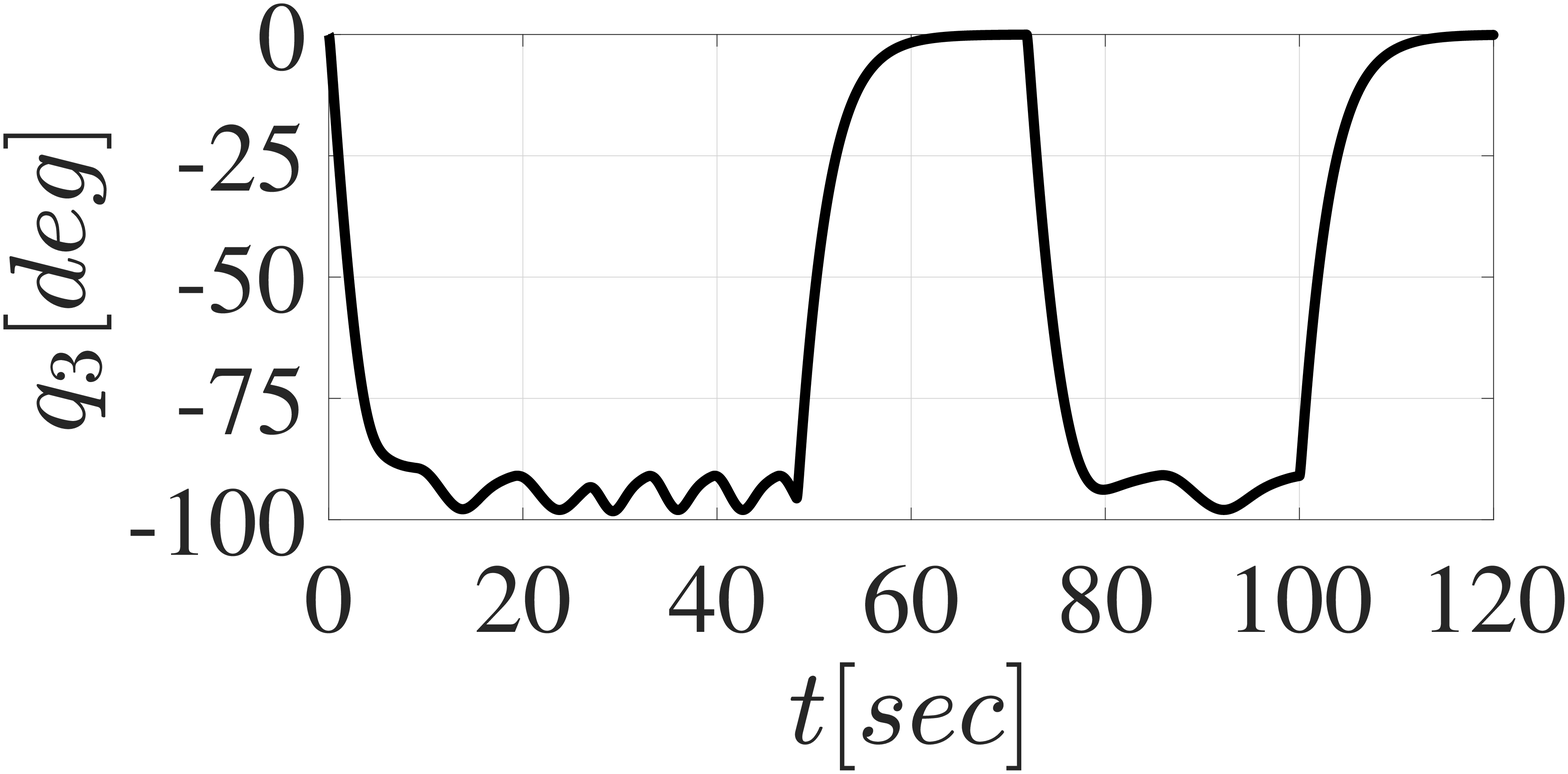}
		\hfill
		\includegraphics[width=0.24\textwidth]{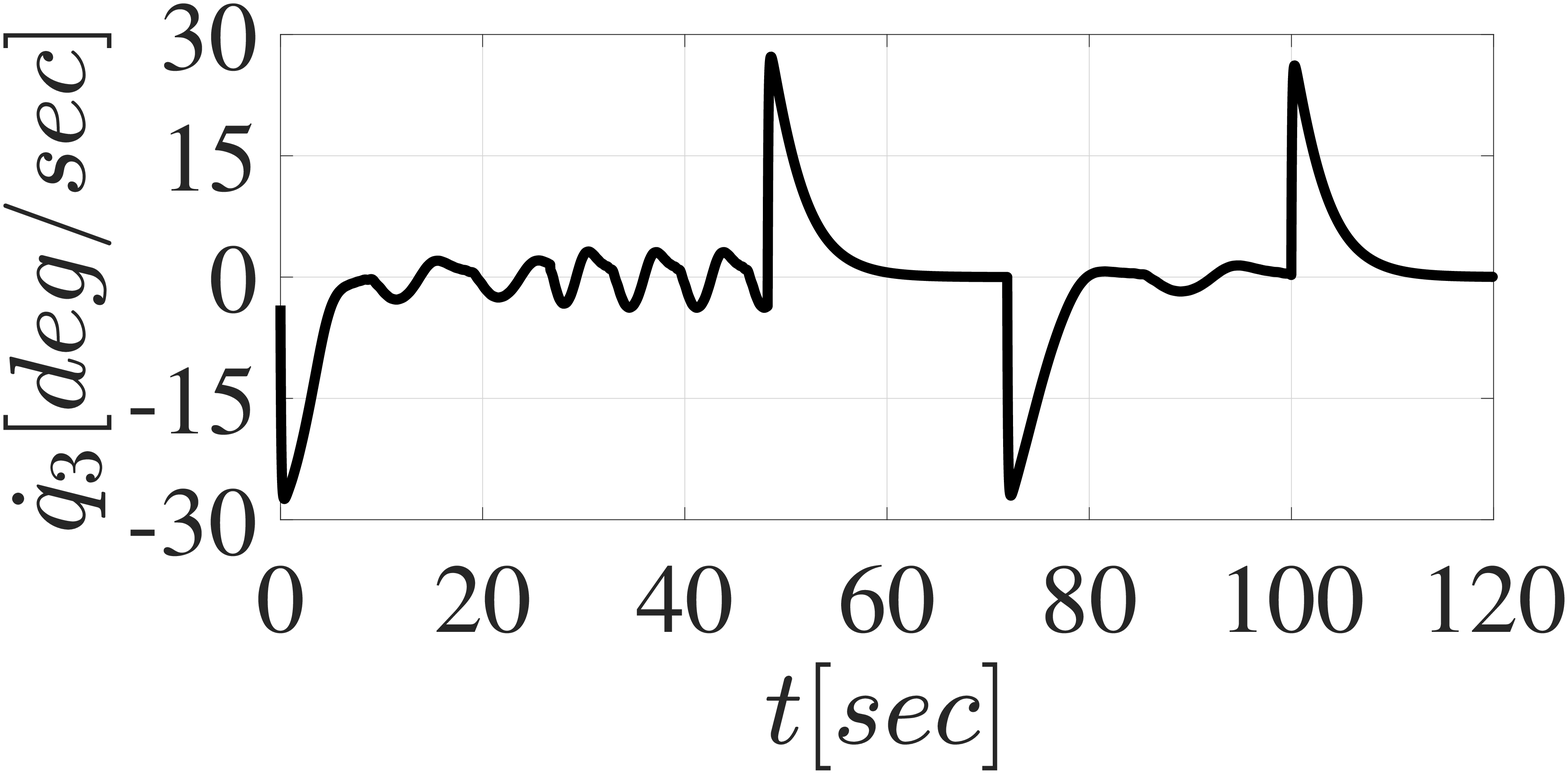}
	}\\
	\subfloat{
		\includegraphics[width=0.24\textwidth]{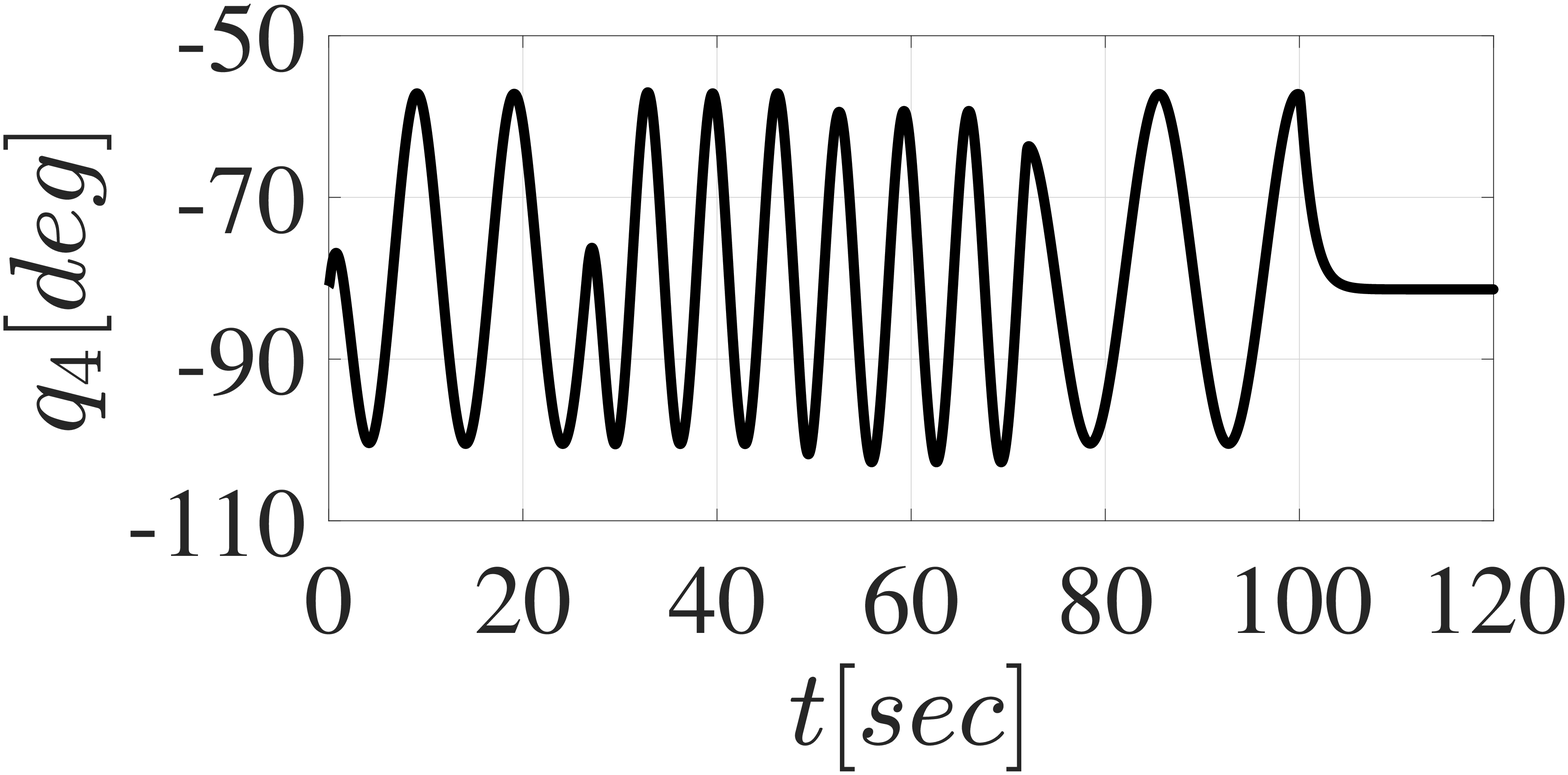}
		\hfill
		\includegraphics[width=0.24\textwidth]{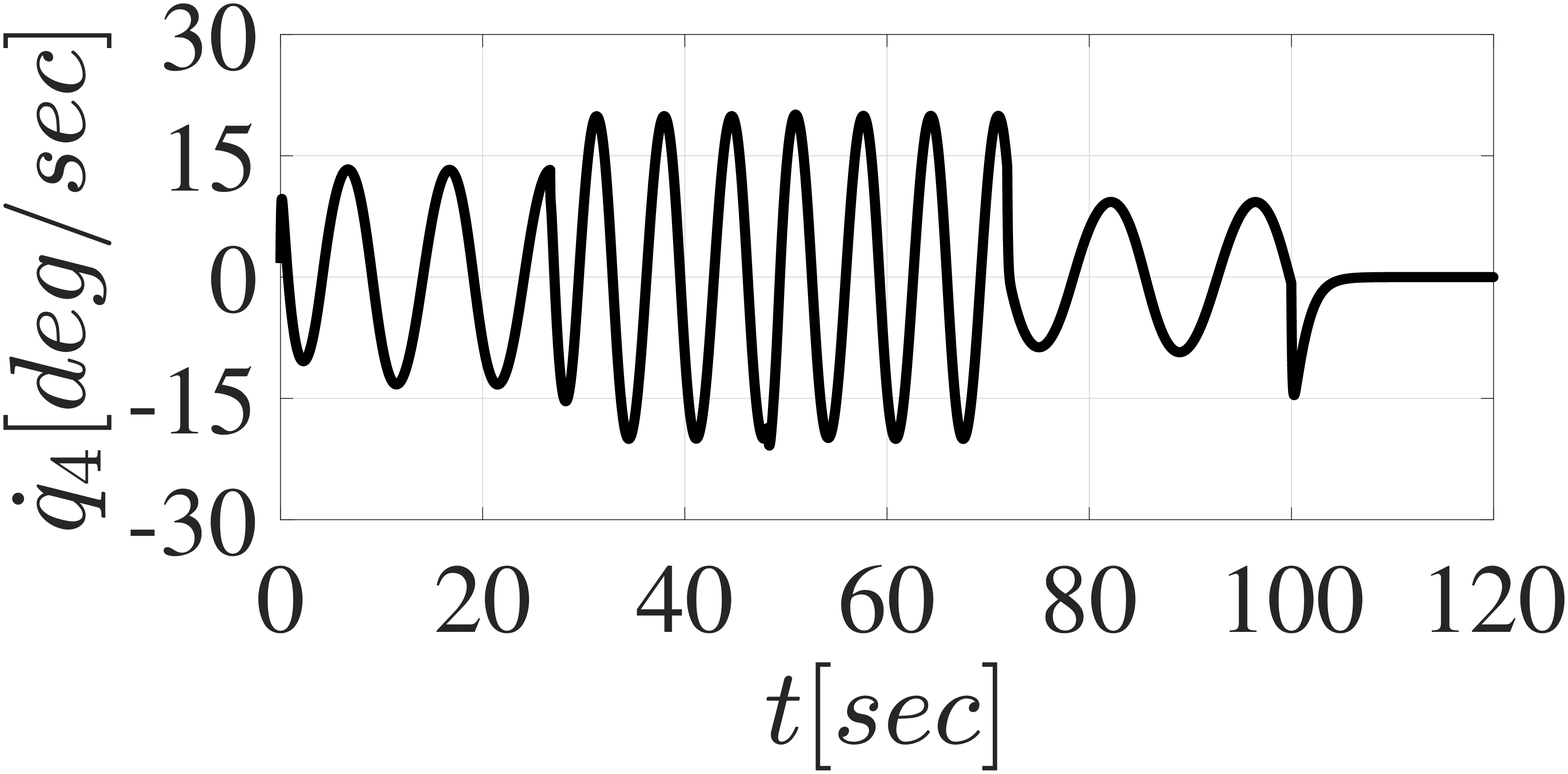}
	}\\
	\subfloat{
		\includegraphics[width=0.24\textwidth]{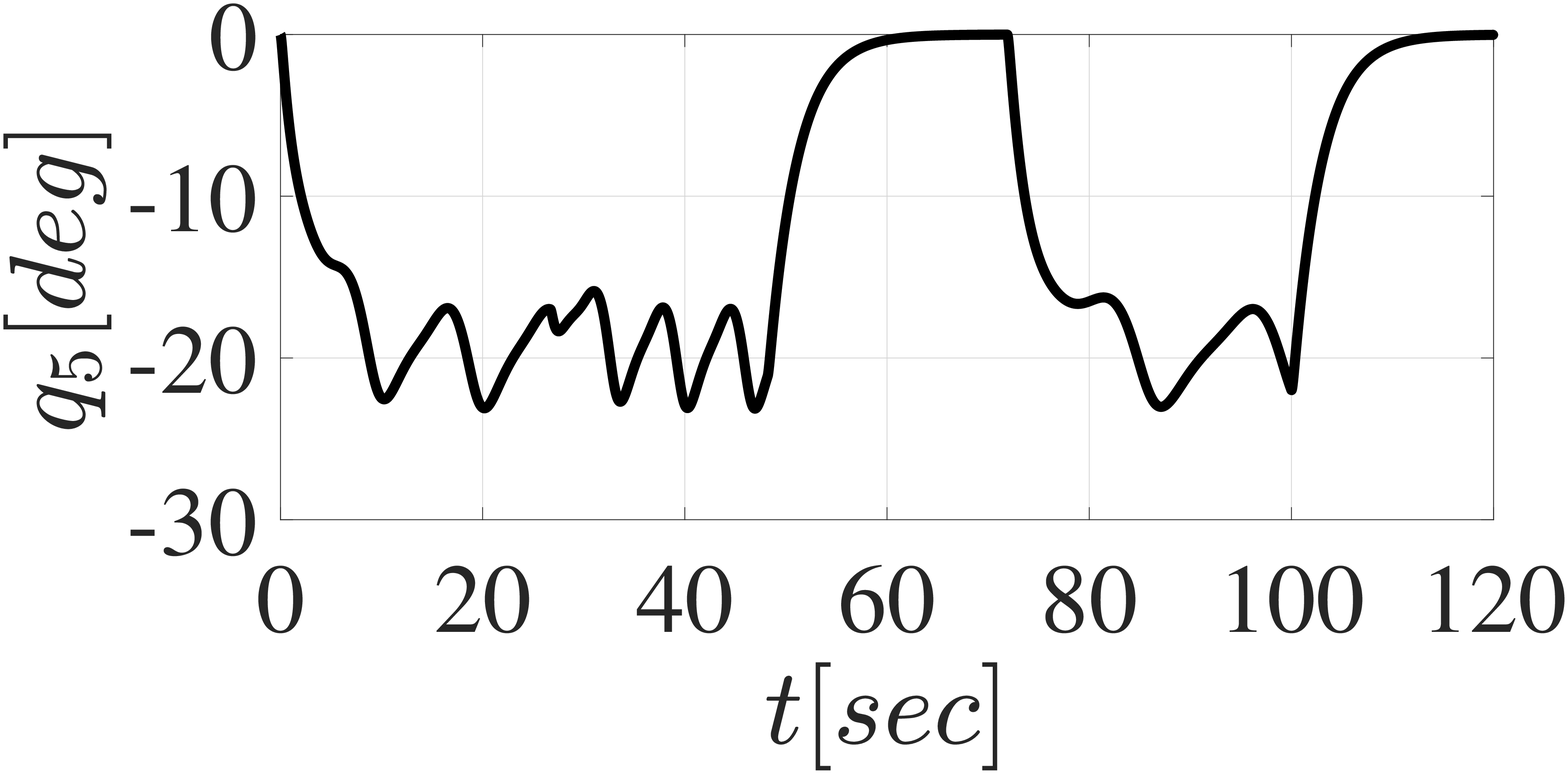}
		\hfill
		\includegraphics[width=0.24\textwidth]{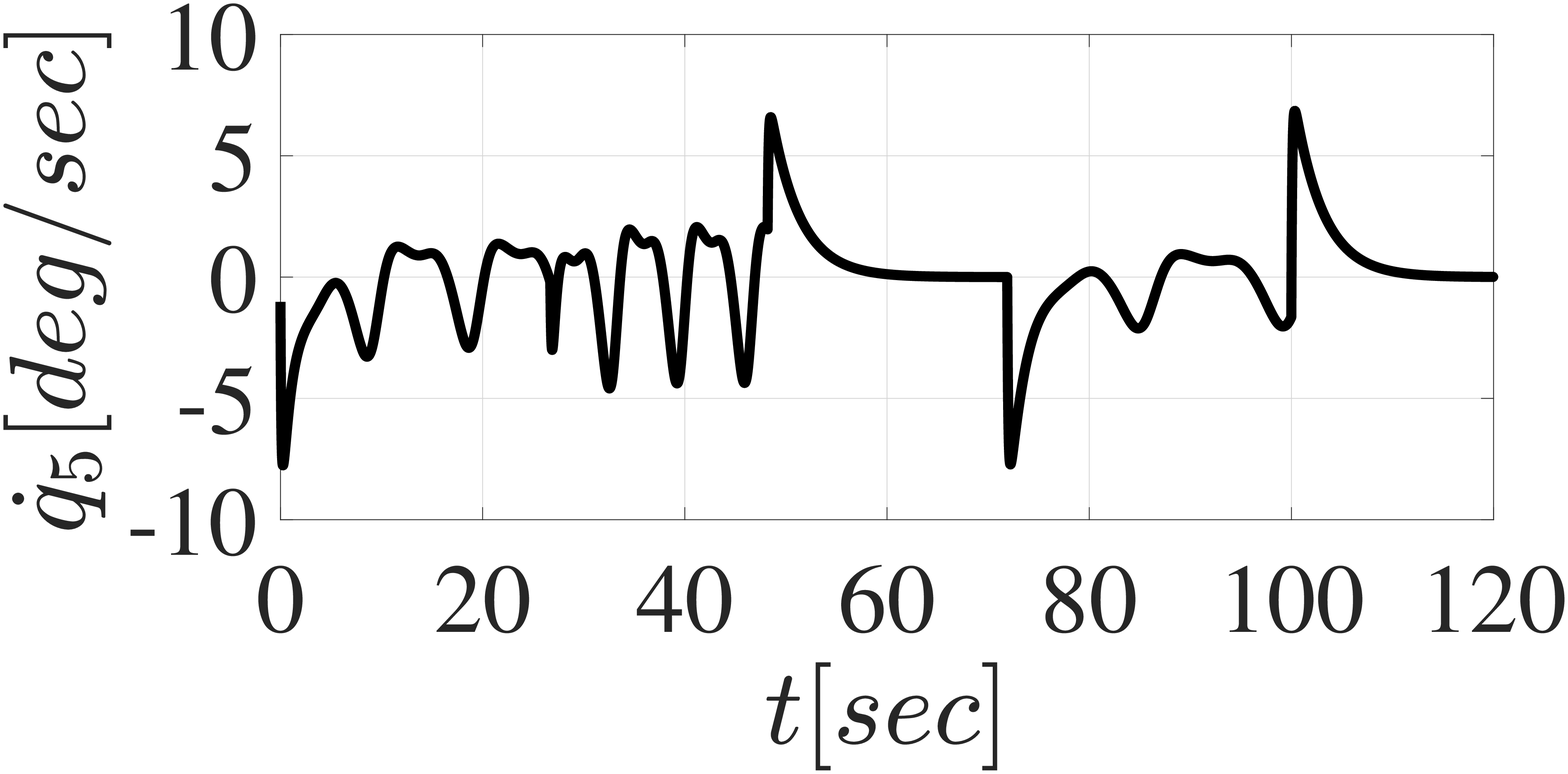}
	}\\
	\subfloat{
		\includegraphics[width=0.24\textwidth]{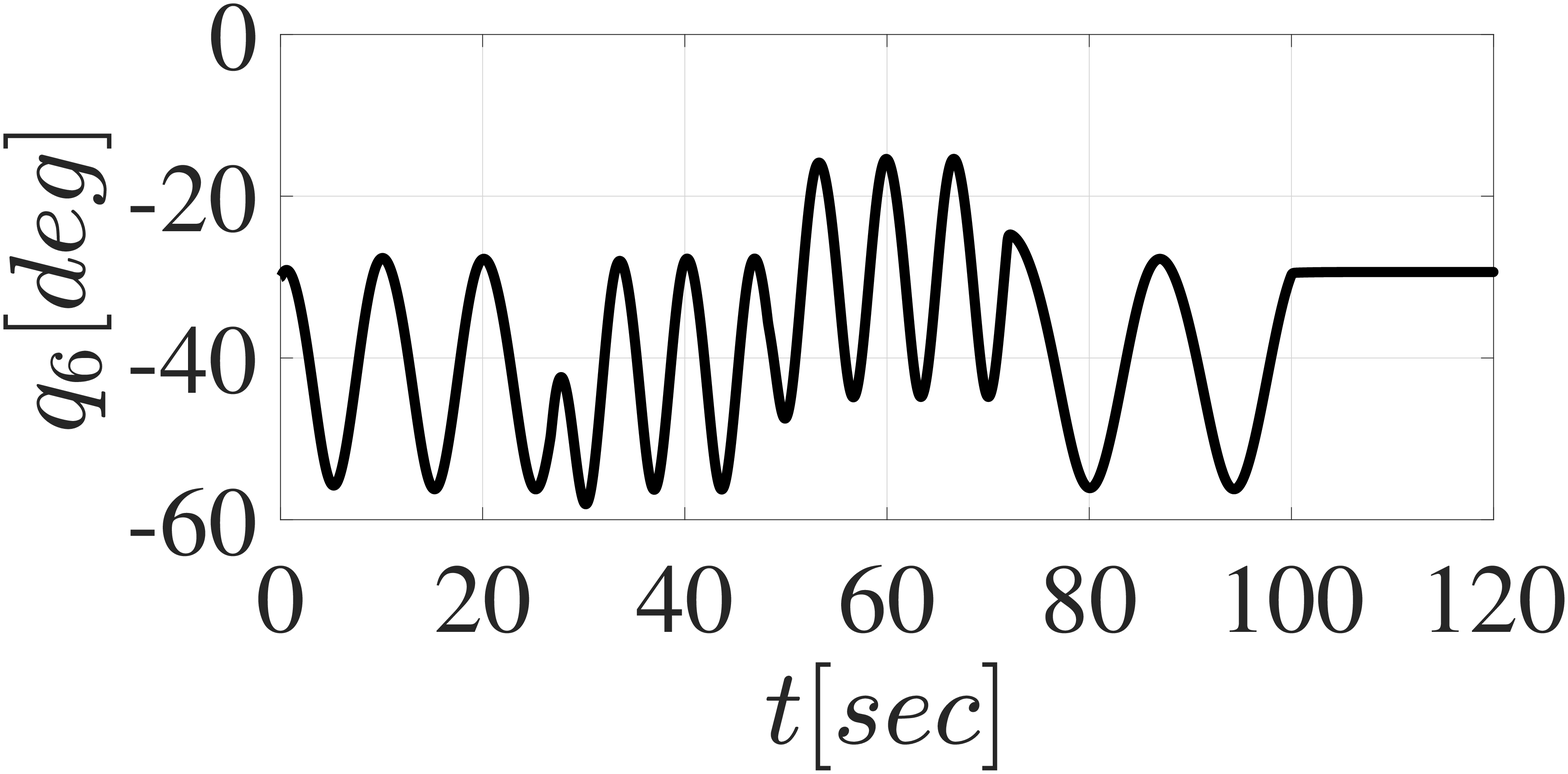}
		\hfill
		\includegraphics[width=0.24\textwidth]{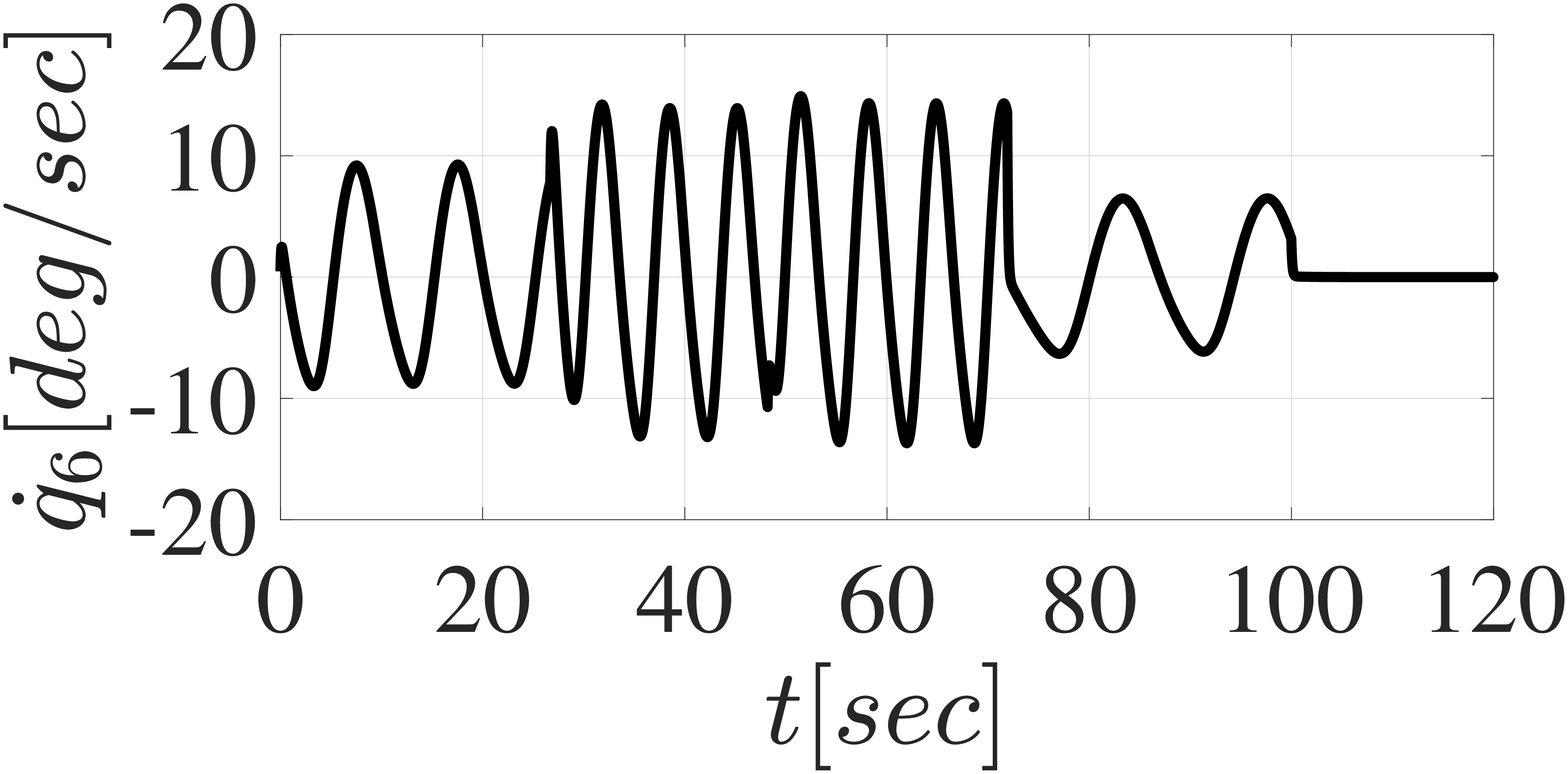}
	}\\
	\subfloat{
		\includegraphics[width=0.24\textwidth]{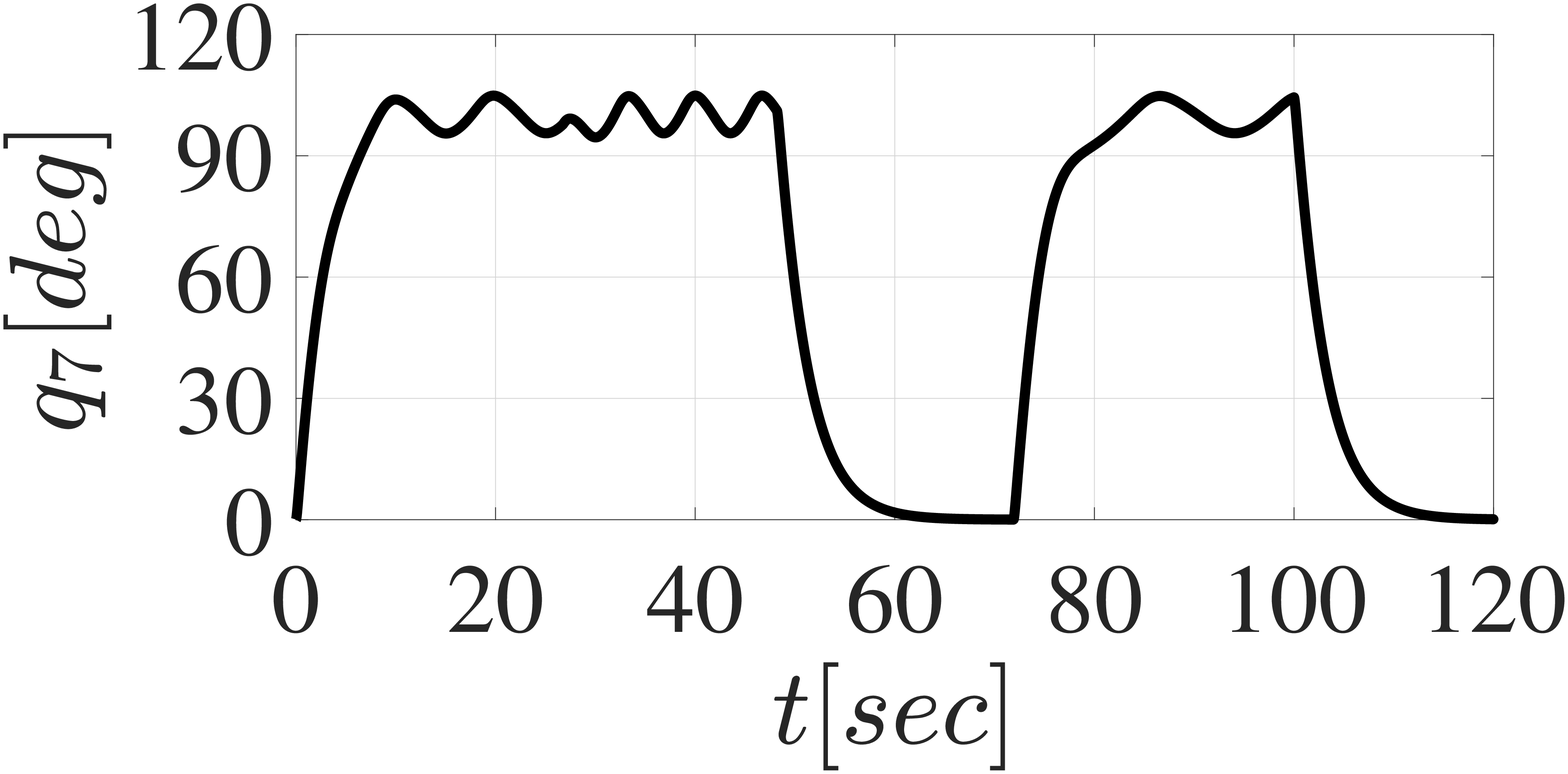}
		\hfill
		\includegraphics[width=0.24\textwidth]{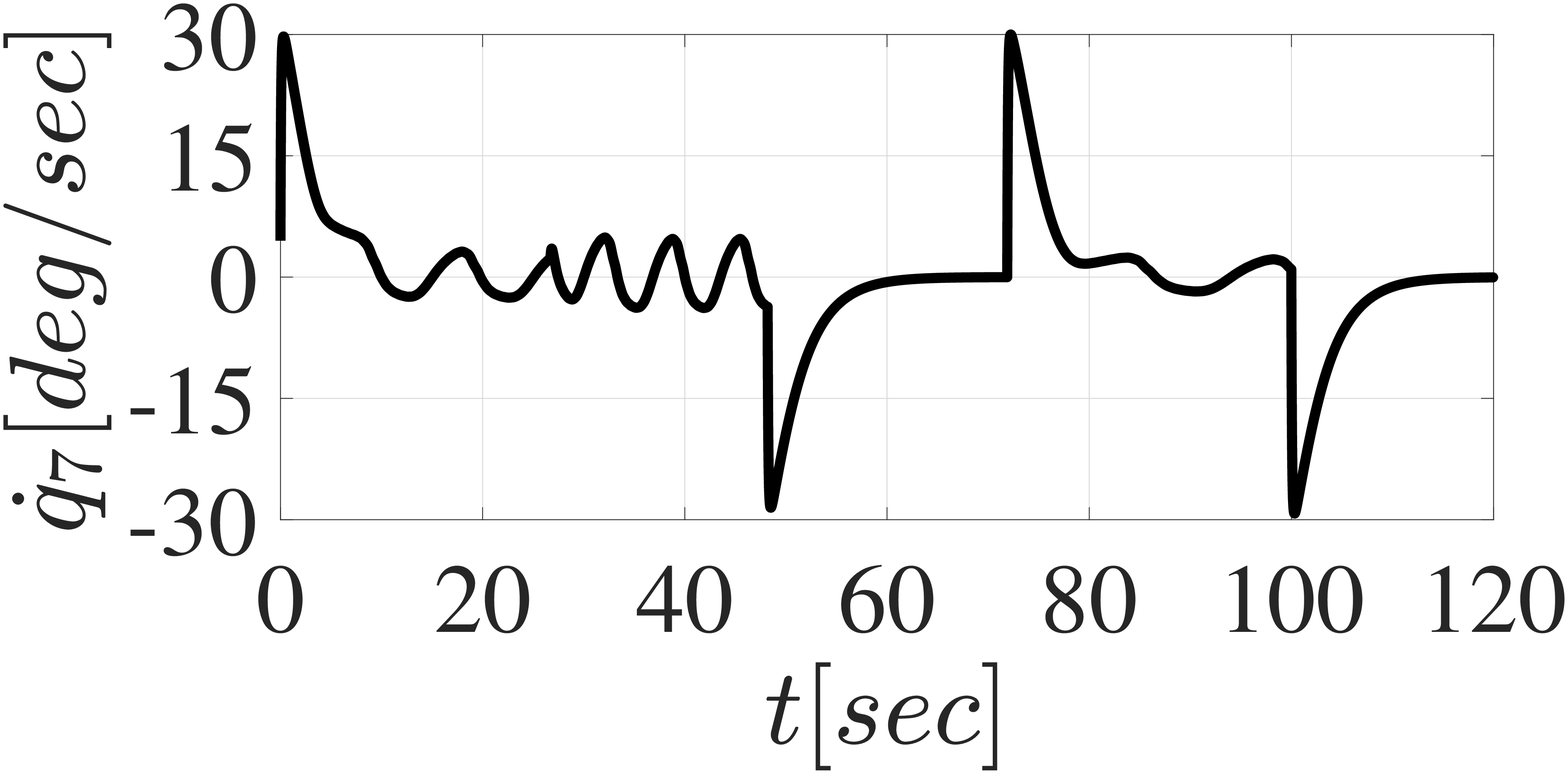}
	}
	\caption{
		The measured joint velocities and positions of the robot arm using the proposed integrated CPG.
		The CPG parameters are tuned to $B = 15 I_7$, $K = 10 I_7$, $D = 25 I_7$, $\gamma = 10$.
	}
	\label{fig:iCPG_kuka_timePlot}
\end{figure}

The desired motion is, at first, the cyclic motion where the end-effector moves on the horizontal circle with $T = 10 (sec)$.
Then, the period of the desired motion changes to $7 (sec)$ at $t = 26 (sec)$.
After that, the desired motion changes to cyclic motion where the end-effector moves on the vertical circle at $t = 48 (sec)$.
The desired motion changes to the horizontal circle with $T = 15 (sec)$ at $t = 72 (sec)$.
Finally, the desired motion changes to the constant posturing at $t = 100 (sec)$.
The above sequences are illustrated in \tablename{~\ref{tab:iCPG_kuka_scenario}}.

\begin{table}[!b]
	\centering
	\caption{The motion sequences which are updated online during  passive robotic rehabilitation scenario}
	\label{tab:iCPG_kuka_scenario}
	\begin{tabular}{|c|c|c|}
		\hline
		$t (sec)$ & desired motion & $T (sec)$\\\hline
		$0$ & horizontal circle	& $10$\\\hline
		$26$ & horizontal circle & $7$\\\hline
		$48$ & vertical circle & $7$\\\hline
		$72$ & horizontal circle & $15$\\\hline
		$100$ & center of the vertical circle & $\dots$\\\hline
	\end{tabular}
\end{table}

The time evolution of the joints positions/velocities of the robot arm is illustrated in \figurename{\ref{fig:iCPG_kuka_timePlot}}.
As can be seen, the joint trajectories change smoothly from one periodic/constant trajectory to the other periodic/constant one.
The joint position limit for $q_1,q_3,q_5$, and $q_7$ is $\left( -170, 170 \right)[deg]$, and for $q_2,q_4$ and $q_6$ is $\left( -120 , 120 \right)[deg]$.
The maximum feasible joint speed is considered as $40 [deg/sec]$ for all the joints.
The trajectories preserve the joint position and velocity limits (\figurename{\ref{fig:iCPG_kuka_timePlot}}).

\begin{figure}[!b]
	\centering
	\subfloat[$t = (0,20) (sec)$]{
		\includegraphics[width=0.24\textwidth]{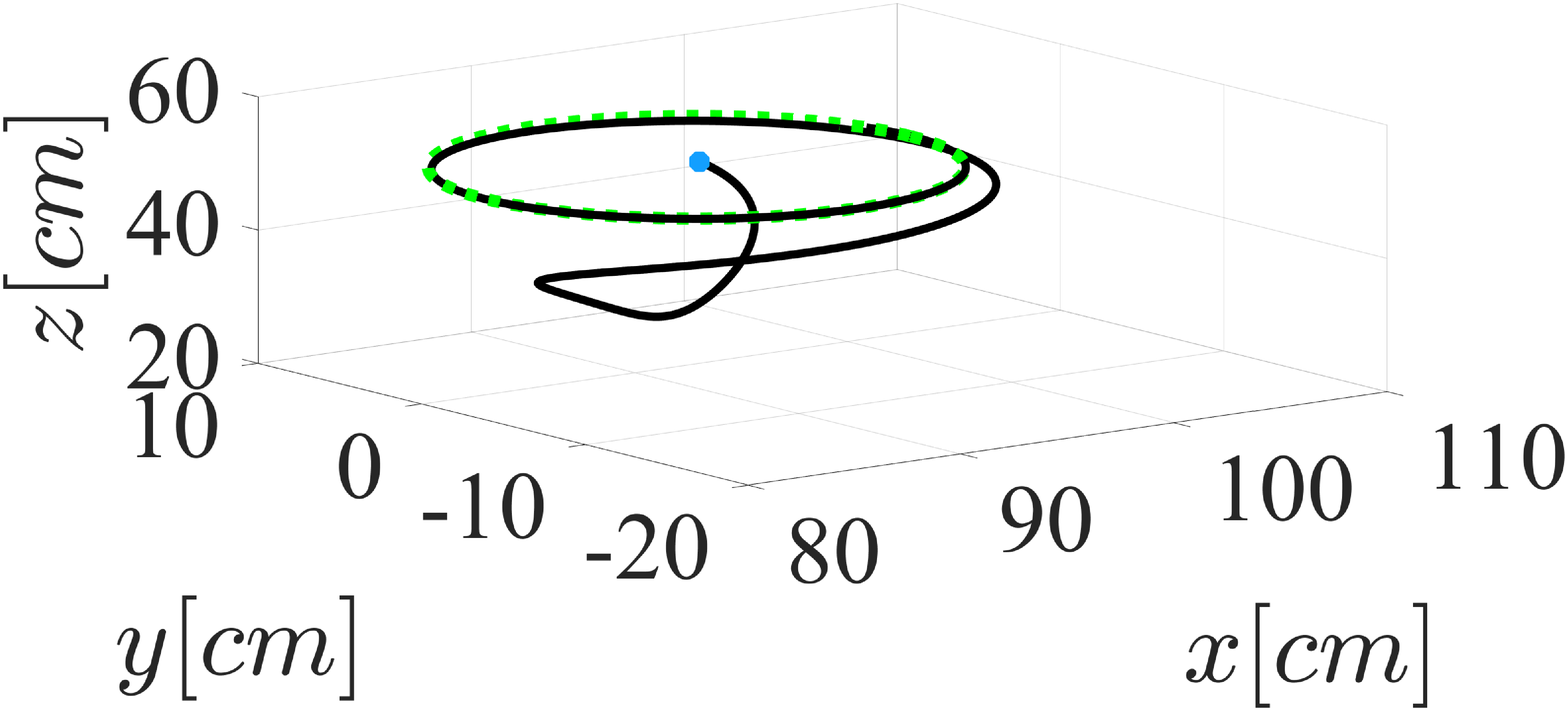}}
	\subfloat[$t = (20,40) (sec)$]{
		\includegraphics[width=0.24\textwidth]{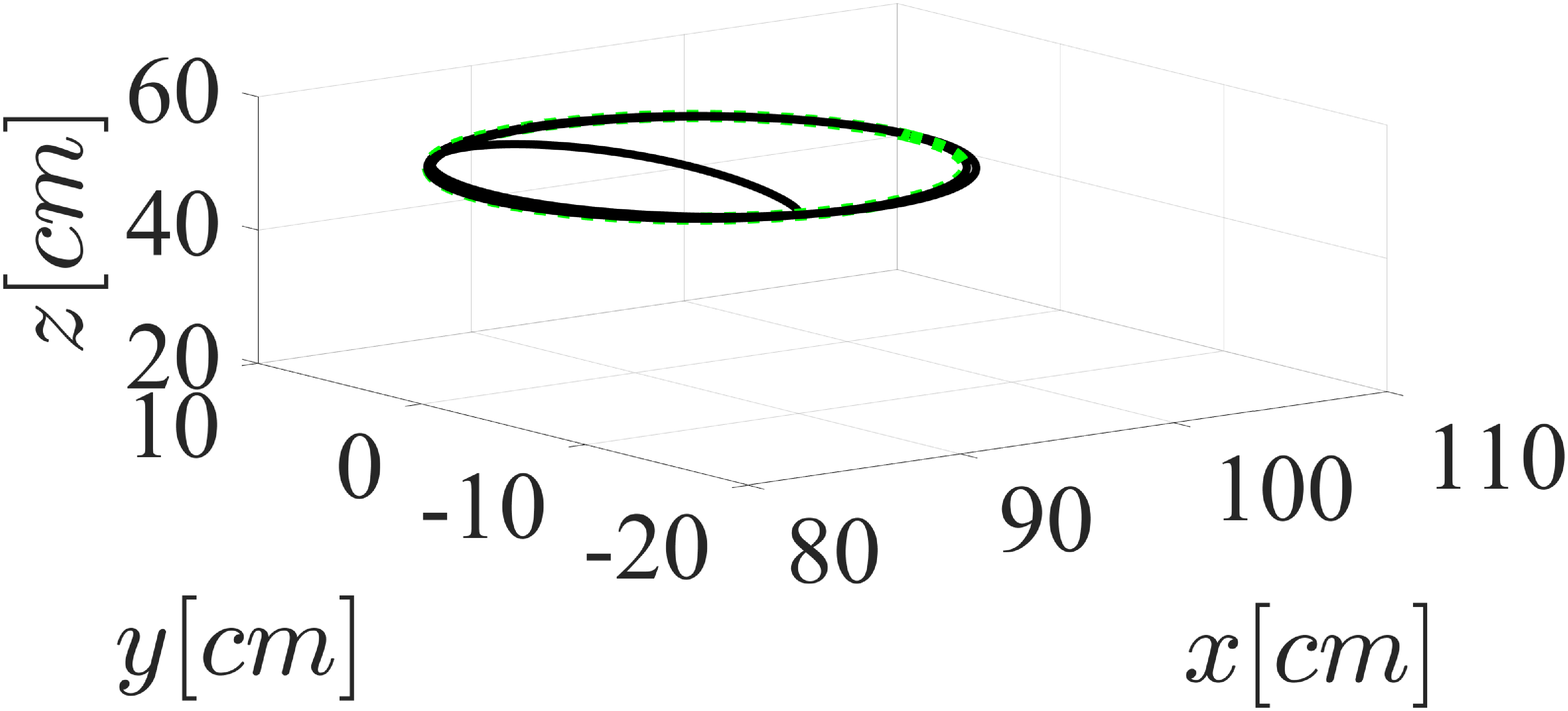}}\\
	\subfloat[$t = (40,65) (sec)$]{
		\includegraphics[width=0.24\textwidth]{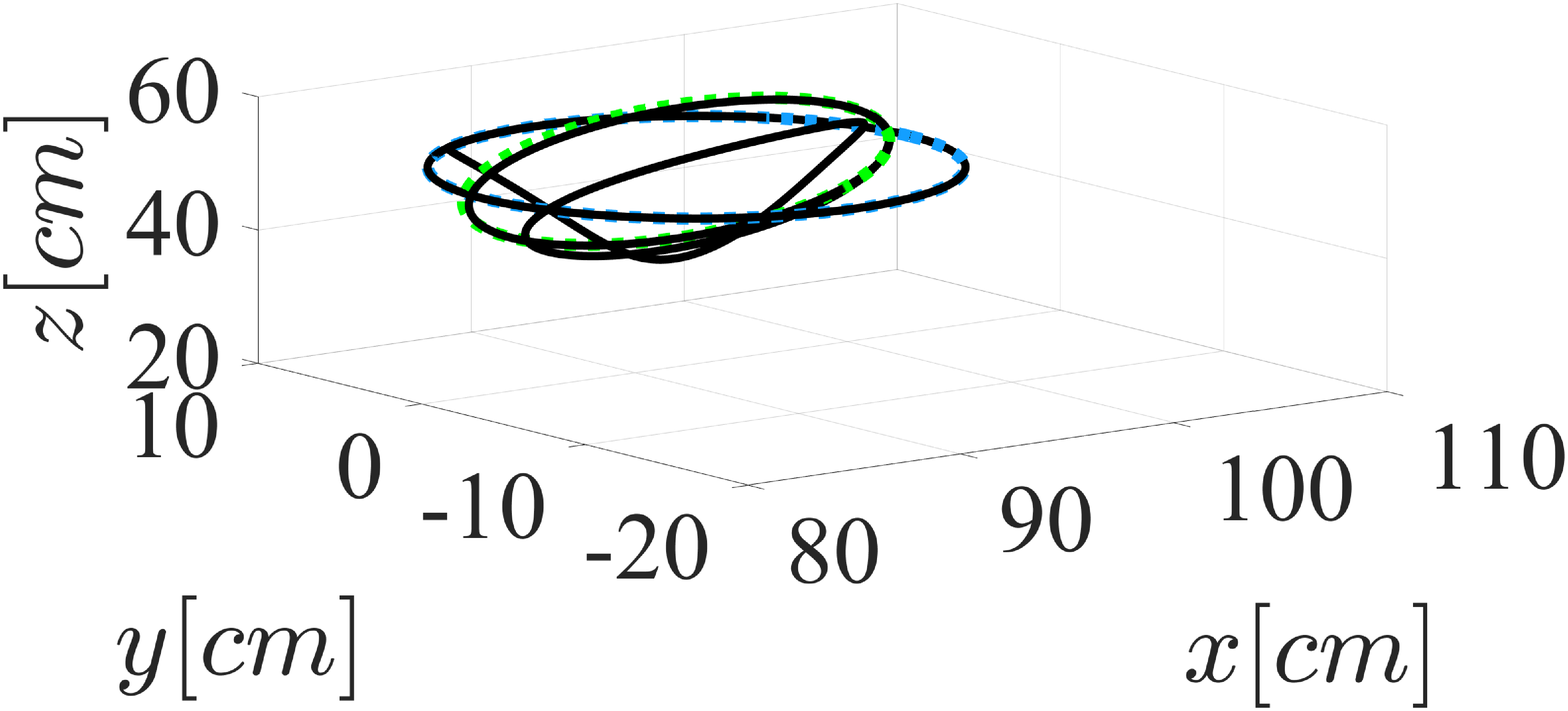}}
	\subfloat[$t = (65,85) (sec)$]{
		\includegraphics[width=0.24\textwidth]{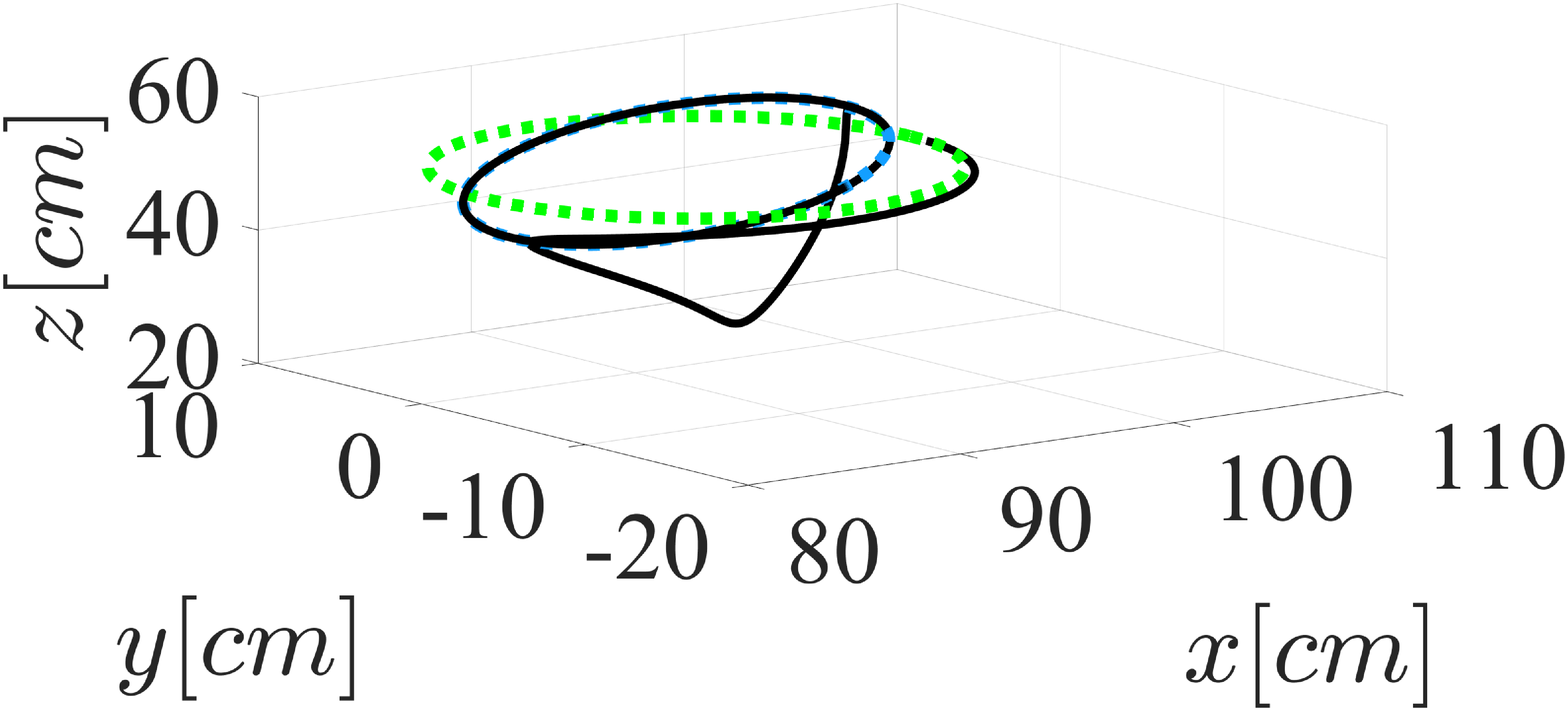}}\\
	\subfloat[$t = (85,120) (sec)$]{
		\includegraphics[width=0.24\textwidth]{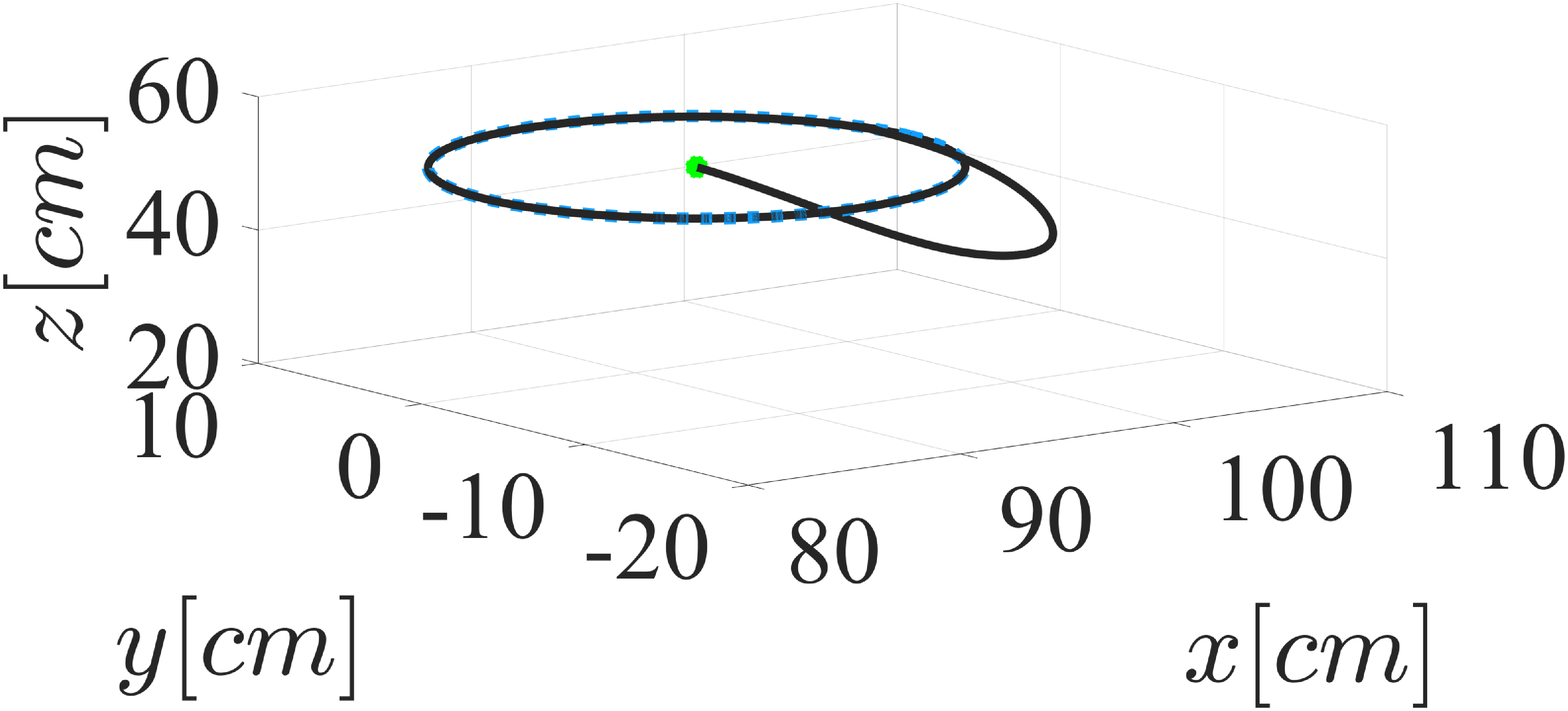}}
	\caption{
	The trajectory of the Kuka end-effector in the Cartesian space while controlling its joints through the proposed integrated CPG.
	The black curve is the end-effector trajectory and the dashed blue and green curves denote previous and new desired trajectories, respectively.
	The blue and green points are the desired end-effector position defined by the third desired motion in the desired motion library.
	The coefficients of the CPG are equal to $B = 15 I_7$, $K = 10 I_7$, $D = 25 I_7$, $\gamma = 10$.
	}
	\label{fig:iCPG_kuka_endeffector}
\end{figure}

The trajectory of the robot end-effector in the Cartesian space is depicted in \figurename{\ref{fig:iCPG_kuka_endeffector}}.
The task space variables converge from one desired trajectory to another one and finally stop at the predefined position.
Therefore, the proposed CPG enables us to switch online between different cyclic motions, smoothly. 

%%%%%%%%%%%%%%%%%%%%%%%%%%%%%%%%%%%%%%%%%%%%%%%%%%%%%%%%%%%%%%%%%%%%%%%%%%%%%%%%%%%%%%%%%%%%%%%%%%%%%%%%%%%%%%%%%%%%%%%%%%%%%%%%%%%%%%%%%%%%%%%%%%

\subsection{Bipedal Walking Pattern Generation}

\begin{figure}[!t]
	\centering
	\includegraphics[width=0.26\textwidth]{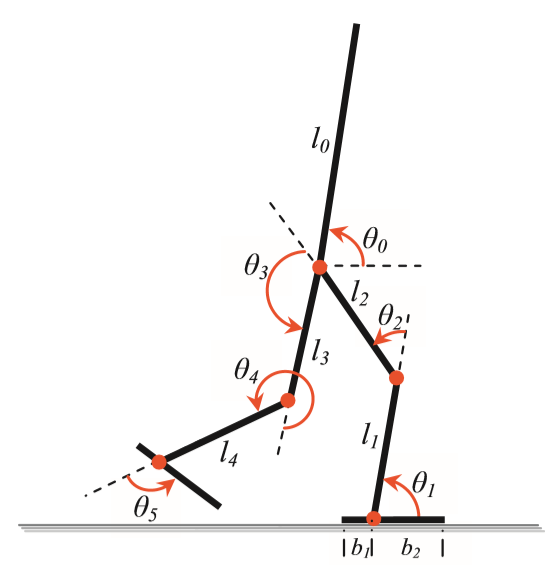}
	\caption{
		The seven link bipedal robot used for the walking simulation \cite{farzaneh_online_2014}.
	}
	\label{fig:sevenLinkBiped}
\end{figure}

A widespread application of CPG concepts in robotics is the control of legged robots locomotion.
Considering a bipedal robot walking at a constant period and step length, the walking pattern is a repetition of a periodic joint trajectory.
In this section, the proposed integrated CPG is used for online joint trajectory generation of a seven link bipedal robot within a simulations study.
The seven-link bipedal robot shown in \figurename{\ref{fig:sevenLinkBiped}} is analyzed in the Simscape MATLAB toolbox.
The physical parameters of the bipedal robot are given in \tablename{~\ref{tab:sevenLinkBiped}}.

\begin{table}[!b]
	\centering
	\caption{
		The physical parameters of the seven link bipedal robot used in the simulation study.
	}
	\label{tab:sevenLinkBiped}
	\begin{tabular}{|c|c|c|}
		\hline
		Link No. & link length (cm) & link mass (kg)\\ \hline
		$l_1$     & $44.28$                & $3.72$               \\ \hline
		$l_2$     & $44.1$                 & $8.0$                \\ \hline
		$l_3$     & $44.28$                & $3.72$               \\ \hline
		$l_4$     & $44.1$                 & $8.0$                \\ \hline
		$l_0$     & $84.6$                 & $46.24$              \\ \hline
		$b_1+b_2$ & $30.5$                 & $1.16$               \\ \hline
	\end{tabular}
\end{table}

\begin{figure}[!b]
	\centering
	\includegraphics[width=0.85\linewidth]{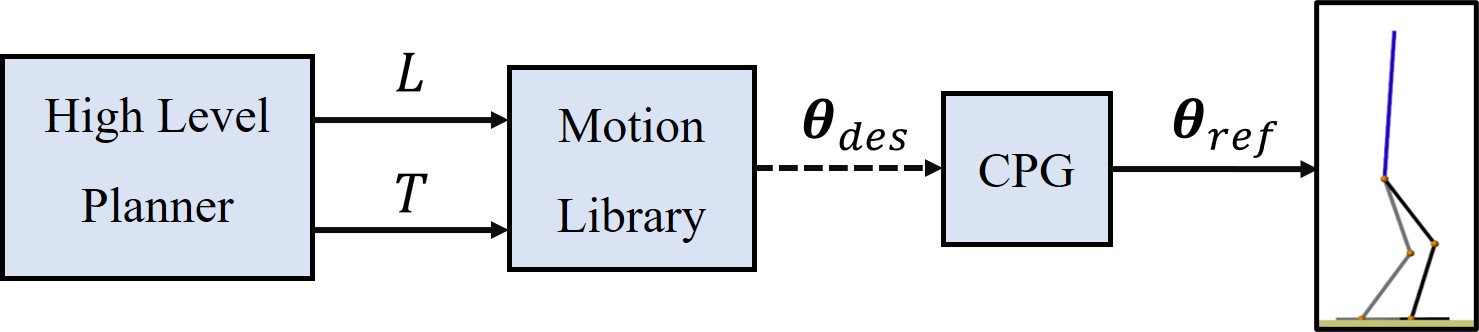}
	\caption{The control architecture used for seven link biped robot walking.
	The biped is simulated in the Simscape MATLAB toolbox.
	The dashed arrow shows that the corresponding signal is just used for updating the following block when the desired motion (\ie $L$ and/or $T$) changes.}
	\label{fig:sevenLinkBiped_controlArchitecture}
\end{figure}

The control architecture for seven-link biped walking is depicted in \figurename{\ref{fig:sevenLinkBiped_controlArchitecture}}.
A high-level planner provides the desired step length ($L$) and step time ($T$). In this study, the values of $L$ and $T$ are as in  \tablename{~\ref{tab:sevenLinkBiped_highLevelController}}.
For the desired step length and period, the motion library determines the desired joint trajectory of the joints.
The motion library is a set of 12-dimensional desired periodic joint trajectories classified based on the step length and period.
The desired trajectory is used for constructing the integrated CPG to generate the reference joint trajectory.
The integrated CPG including a bounded output programmable oscillator is shown in \figurename{\ref{fig:integratedCPG}}.
For the integrated CPG, the lower and upper bounds of the feasible position for $\left[\theta_1, \theta_2, \dots, \theta_5, \theta_0 \right]$ are equal to $\left[0,0,90,180,0,60\right] (deg)$ and $\left[ 180,180,270,360,180,120 \right] (deg)$, respectively.
The maximum speed of all the joints is considered as $180 (deg/sec)$.

Considering the values given in \tablename{~\ref{tab:sevenLinkBiped_highLevelController}}, the motion library consists of five periodic and one constant trajectories.
The periodic trajectories result in walking with different step lengths and periods, and the constant trajectory denotes the standing posture for the robot.
The periodic trajectories are designed based on ZMP stability criteria \cite{vukobratovic2004zero}.
For this purpose, at first, the trajectory of the hip and the swing foot for one step are designed in the Cartesian space according to the desired step length and period.
The position of the lower body joints (\ie $\theta_1, \theta_2,\dots,\theta_5$) are then determined for some specific time instances within the step using inverse kinematic.
After that, the trajectory of ZMP in the walking direction ($zmp_x$) is designed according to the support polygon of the robot.
The trajectory of the center of mass (COM) of the robot in the walking direction ($\overline{x}$) in the Cartesian space is computed for providing the designated $zmp_x$.
For this purpose, we used the simplified $zmp_x$ for the cases where the height of COM is almost constant during the walking, as
\begin{equation}
	g zmp_x \approx g \overline{x} - h \ddot{\overline{x}},
\end{equation}
where $h$ is the average height of the COM, and $g$ is the gravity acceleration.
Considering the position of the lower body joints at the specified time instances and the trajectory of $\overline{x}$, we calculate the position of the upper body joint $\theta_0$ at the specified time instances.
Finally, we approximated the trajectory of all the joints by Fourier series and calculate $\bbf(t) = \left[\theta_1(t), \theta_2(t), \dots, \theta_5(t), \theta_0(t)\right]^\top$, $\dot{\bbf}(t) =  \left[\dot{\theta}_1(t), \dot{\theta}_2(t), \dots, \dot{\theta}_5(t), \dot{\theta}_0(t)\right]^\top$ and $\ddot{\bbf}(t) = \left[\ddot{\theta}_1(t), \ddot{\theta}_2(t), \dots, \ddot{\theta}_5(t), \ddot{\theta}_0(t)\right]^\top$, where $\theta_i(t)$ is the Fourier series estimation of the trajectory of the joint $i$.

\begin{table}[!b]
	\centering
	\caption{
		The desired step length and time for the simulation of the seven link biped walking.
	}
	\label{tab:sevenLinkBiped_highLevelController}
	\begin{tabular}{|c|c|c|}
		\hline
		time Interval (sec) & step length (cm) & step time (sec) \\\hline
		$0-12$              & $50$               & $2$              \\ \hline
		$12-23$             & $50$              & $1$              \\ \hline
		$23-37$            & $80$               & $1.7$            \\ \hline
		$37-48$            & $50$               & $2$              \\ \hline
		$48-58$            & $20$               & $4$            \\ \hline
		$58-68$            & $10$               & $4$            \\ \hline
		$68-70$            & $0$               & $0$            \\ \hline
	\end{tabular}
\end{table}

\begin{figure}[!t]
	\centering
	\subfloat{
		\includegraphics[width=0.5\linewidth]{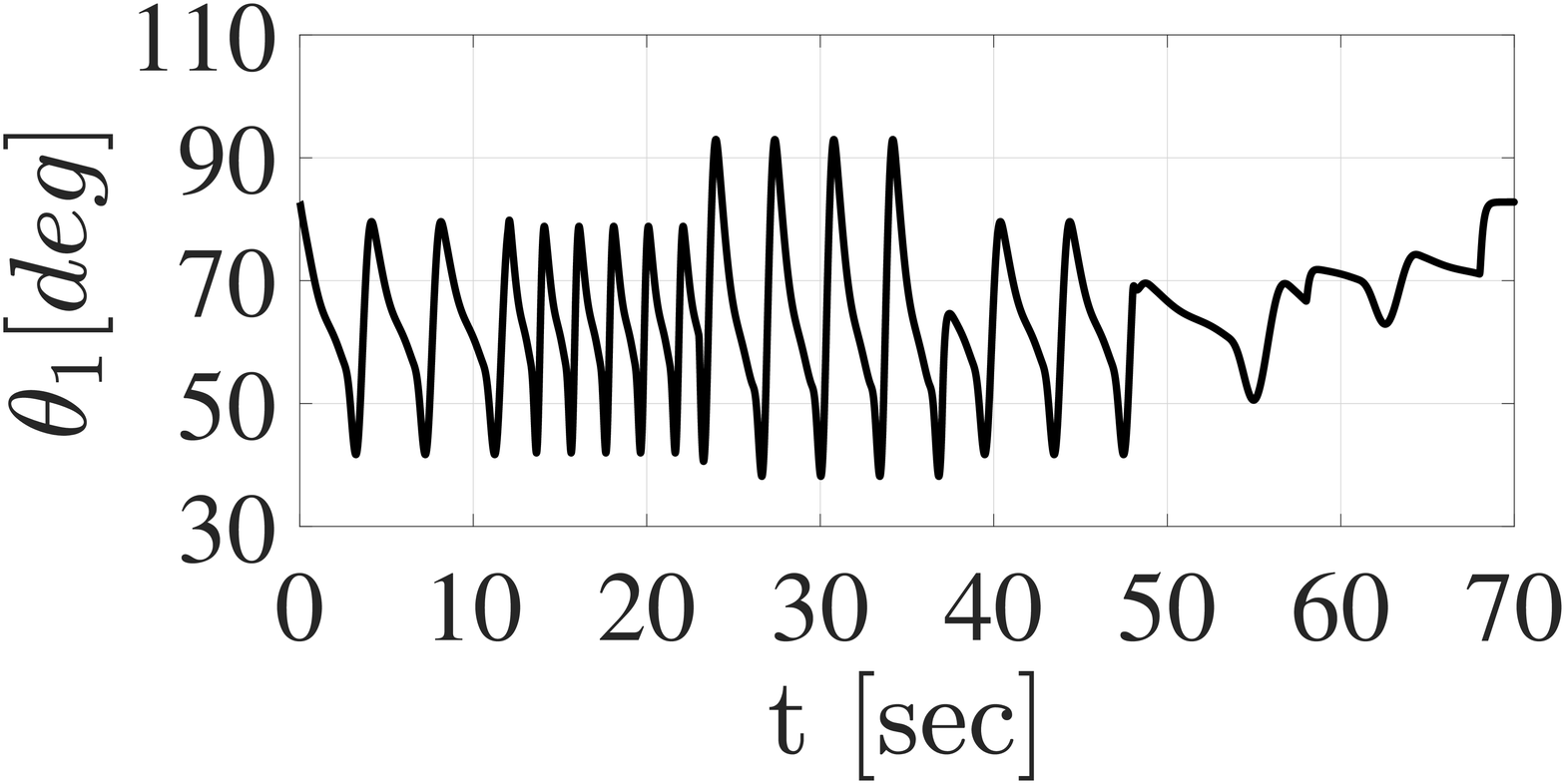}
		\includegraphics[width=0.5\linewidth]{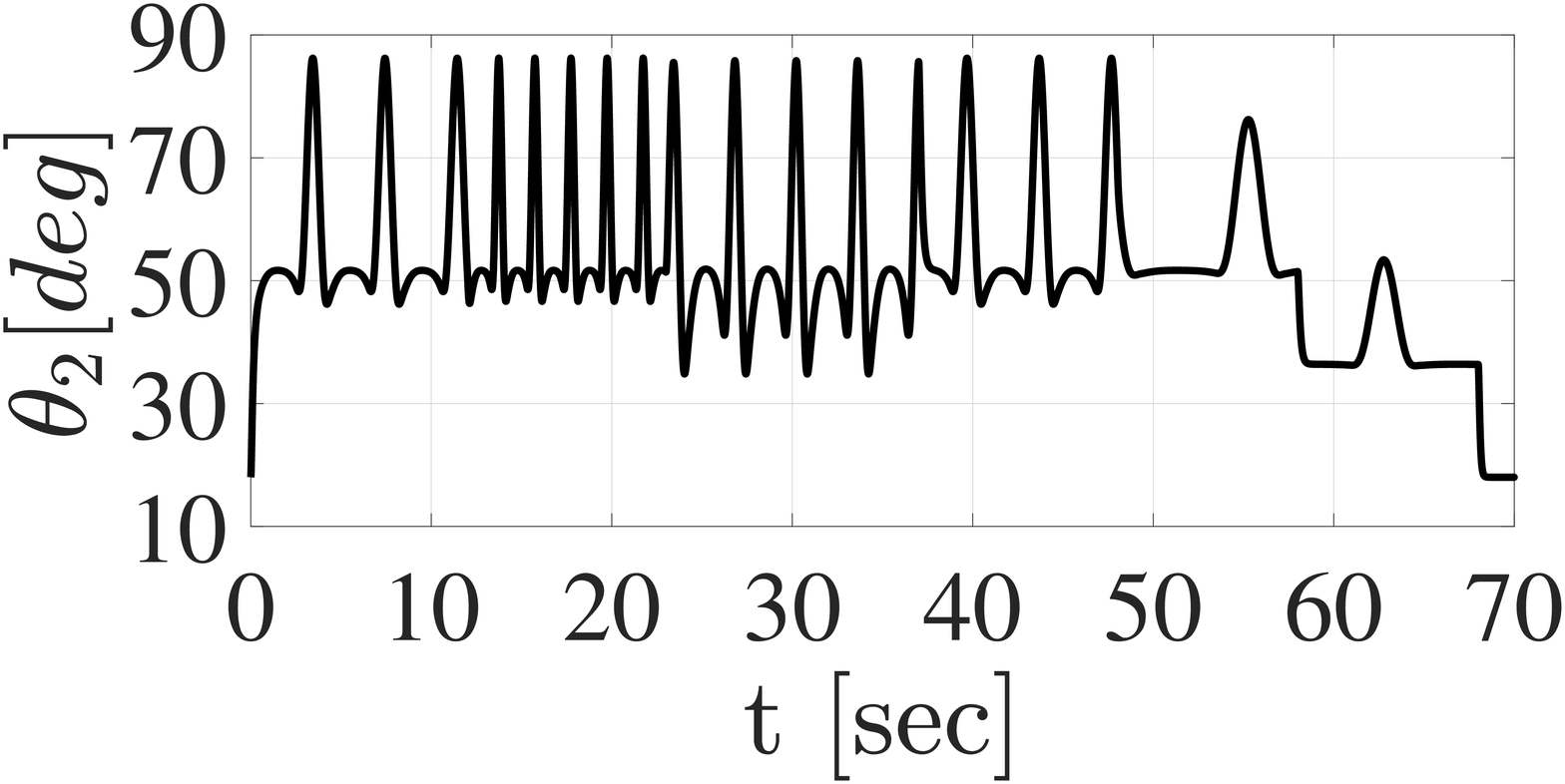}
	}\\
	\subfloat{
		\includegraphics[width=0.5\linewidth]{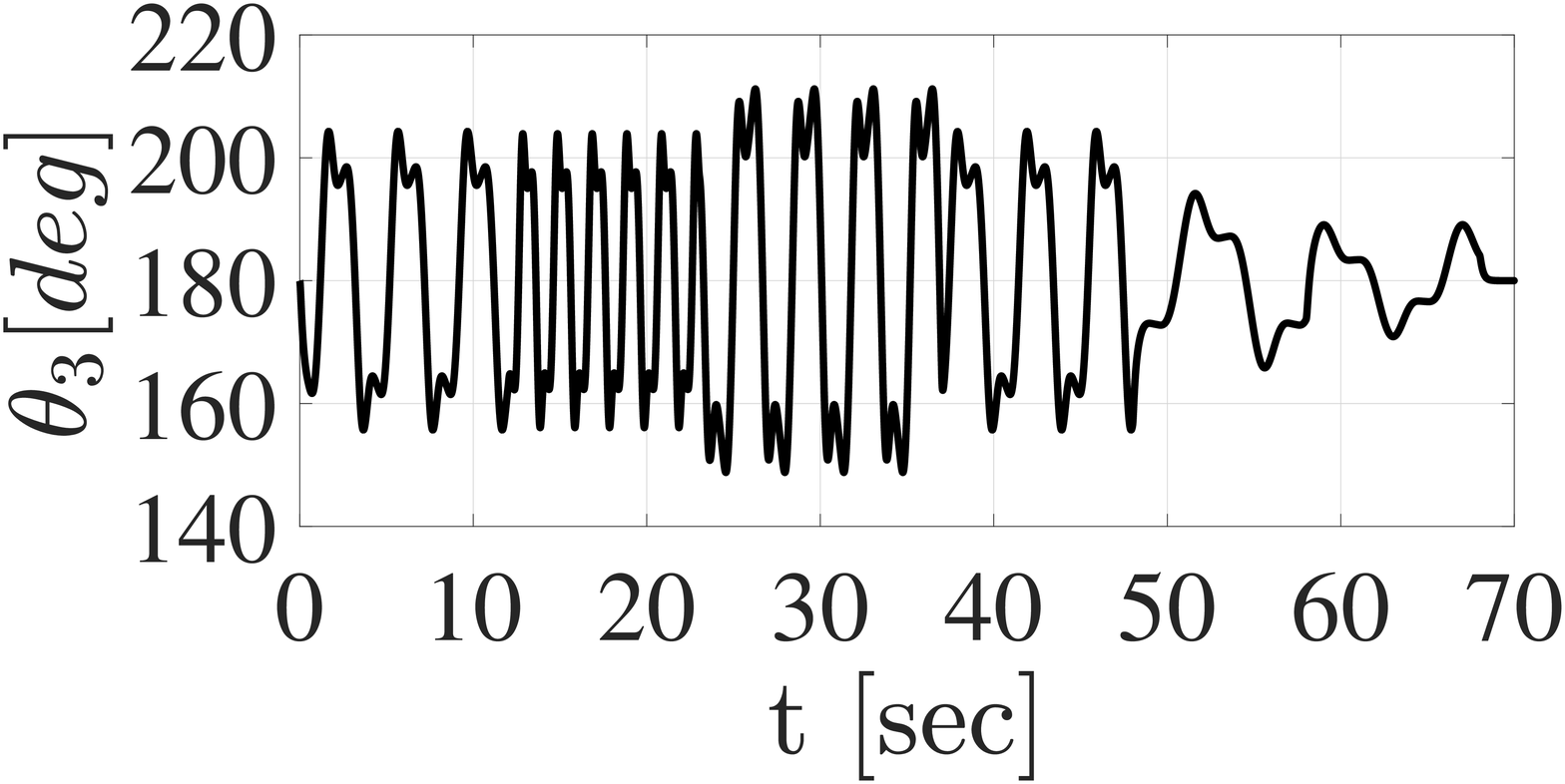}
		\includegraphics[width=0.5\linewidth]{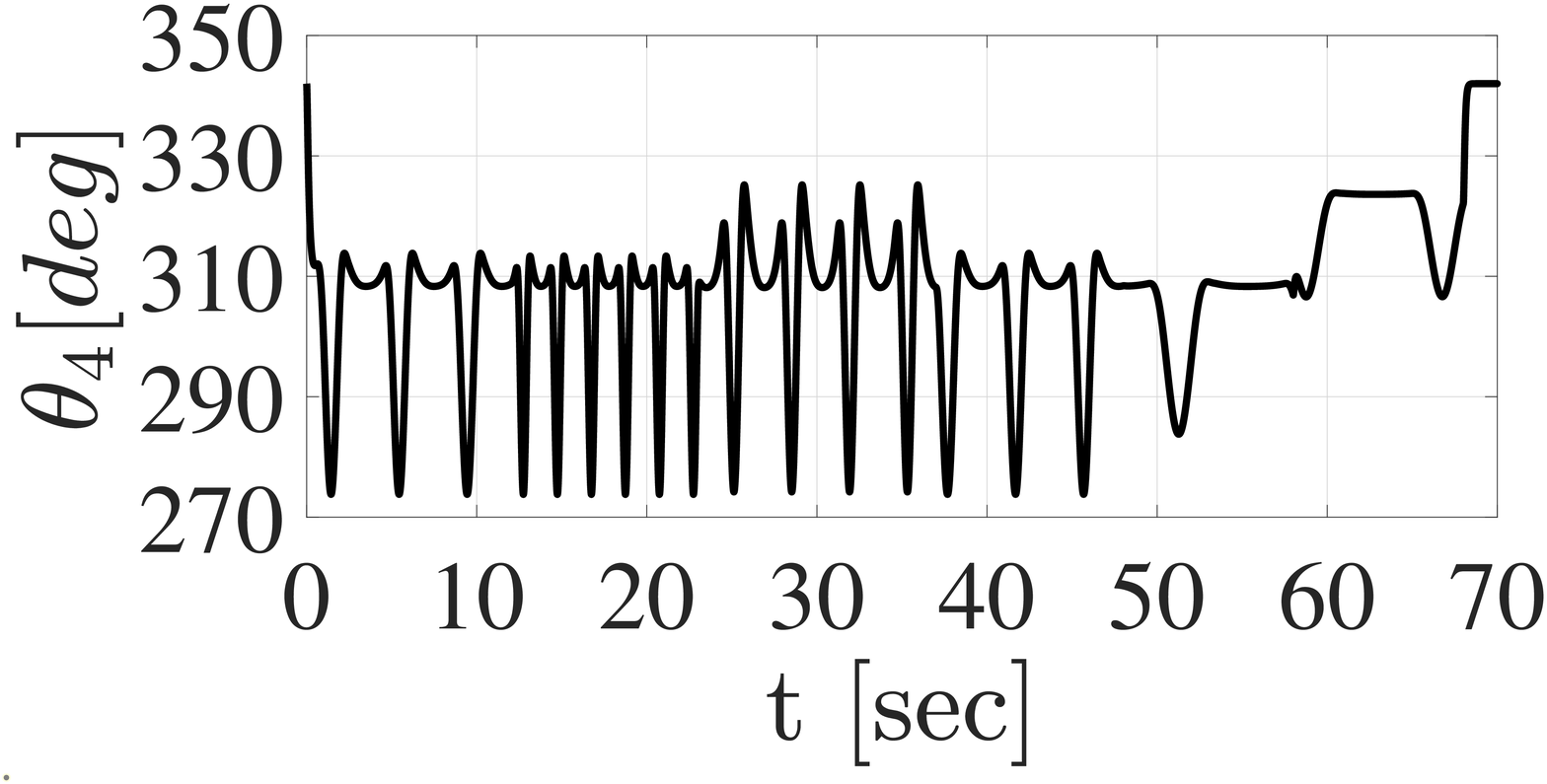}
	}\\
	\subfloat{
		\includegraphics[width=0.5\linewidth]{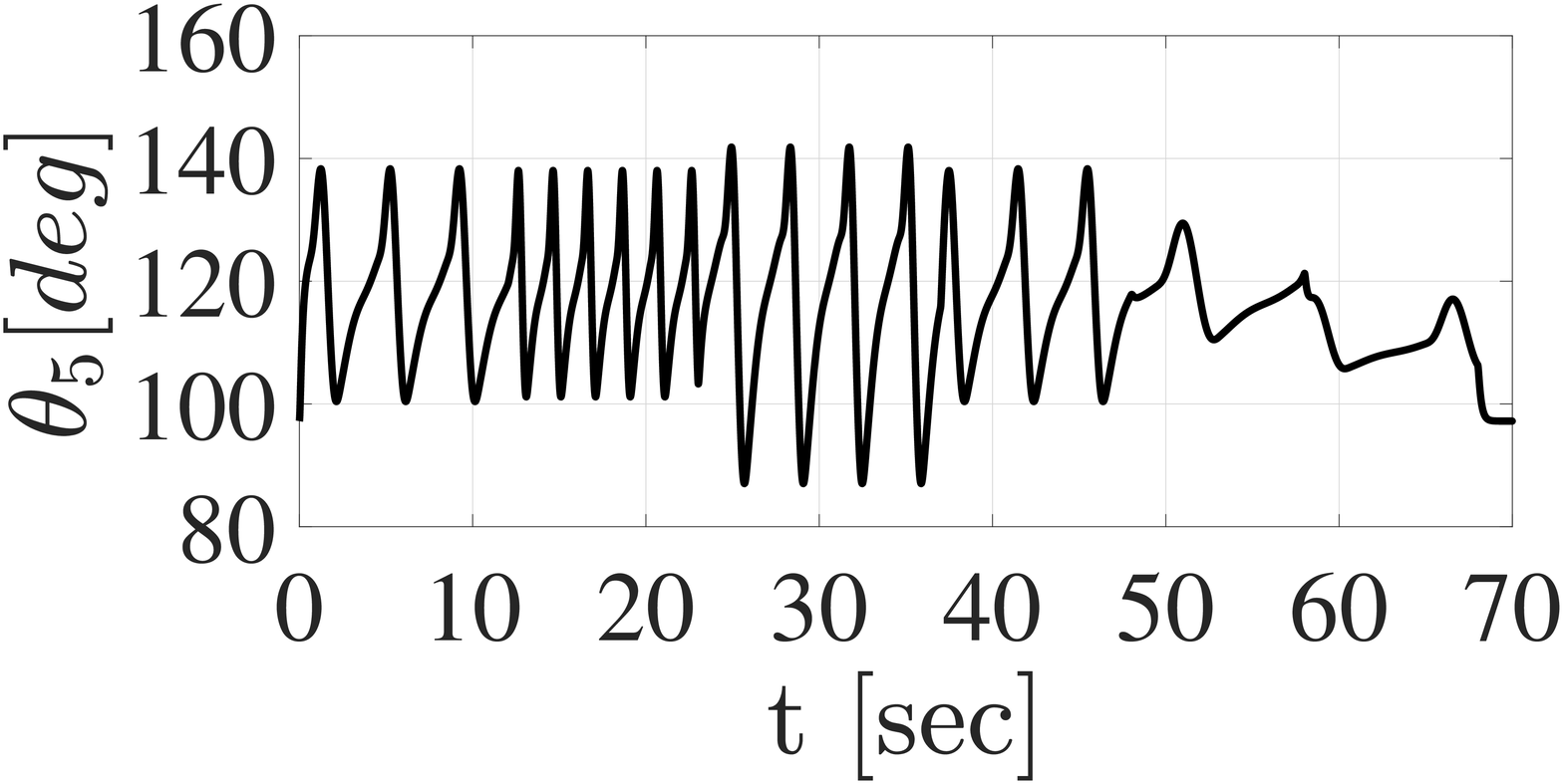}
		\includegraphics[width=0.5\linewidth]{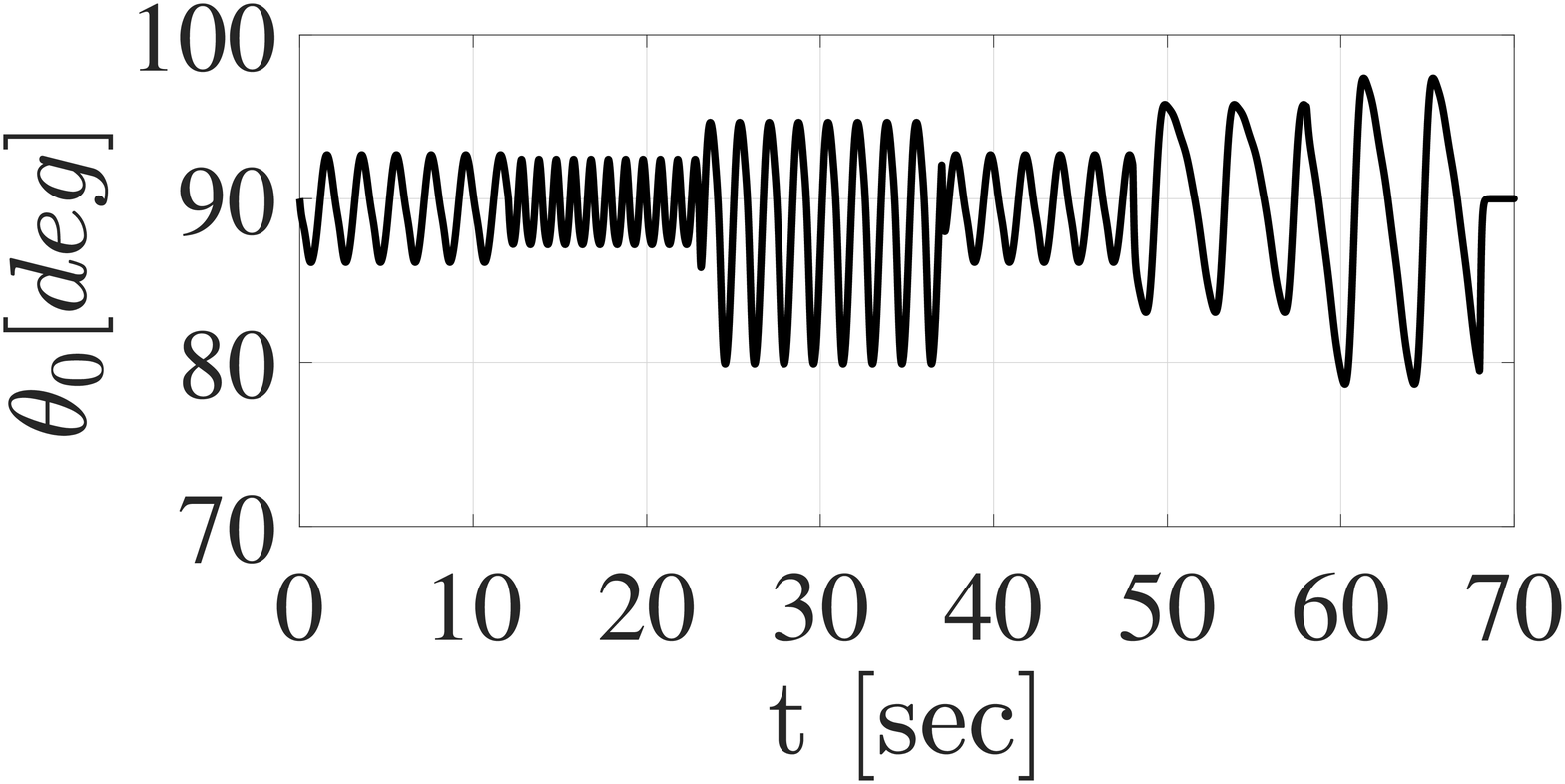}
	}
	\caption{
		The time evolution of the joint position of the seven link biped measured during the walking scenario described in \tablename{~\ref{tab:sevenLinkBiped_highLevelController}}.
		The coefficients of the integrated CPG are
		$B = 40 I_7$,
		$K = 30 I_7$,
		$D = 60 I_7$ and
		$\gamma = 10$.
	}
	\label{fig:sevenLinkWalkingAOS_timePlot_q}
\end{figure}

The time evolution of the joints position of the robot is depicted in \figurename{\ref{fig:sevenLinkWalkingAOS_timePlot_q}}.
The robot is initialized in the standing posture described by a constant desired trajectory provided by the motion library.
As can be seen, the trajectories smoothly converge from one periodic trajectory to another one when the desired motion changes.
Moreover, the position and velocity constraints are respected for all the joints.
For brevity, the time evolution of the joints velocities is not reported here. 

\begin{figure}[!t]
	\centering
	\subfloat[$t \in \left( 0,10 \right) (sec)$]{
		\includegraphics[width=0.5\linewidth]{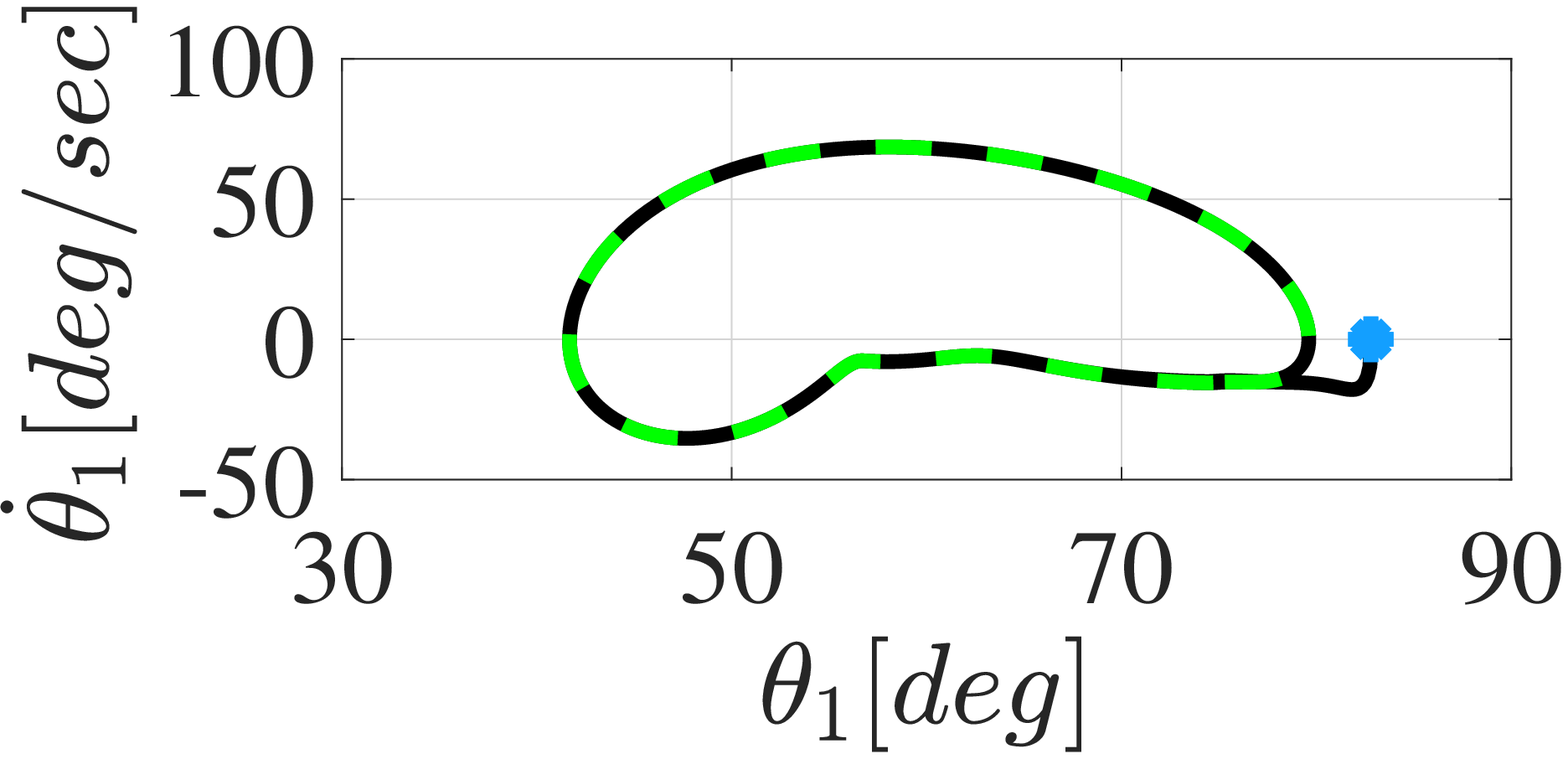}
		\includegraphics[width=0.5\linewidth]{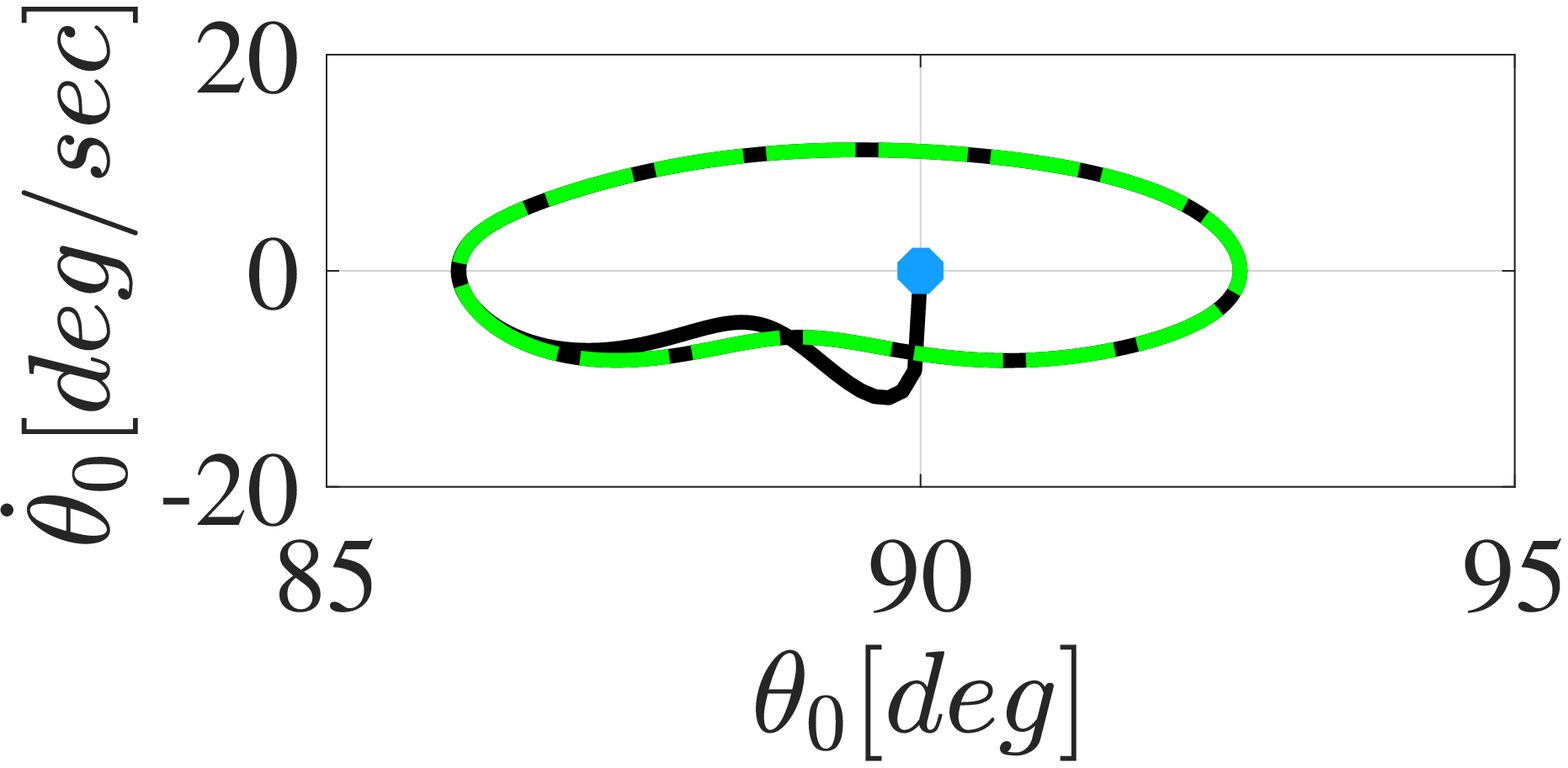}
	}\\
	\subfloat[$t \in \left( 10,20 \right) (sec)$]{
		\includegraphics[width=0.5\linewidth]{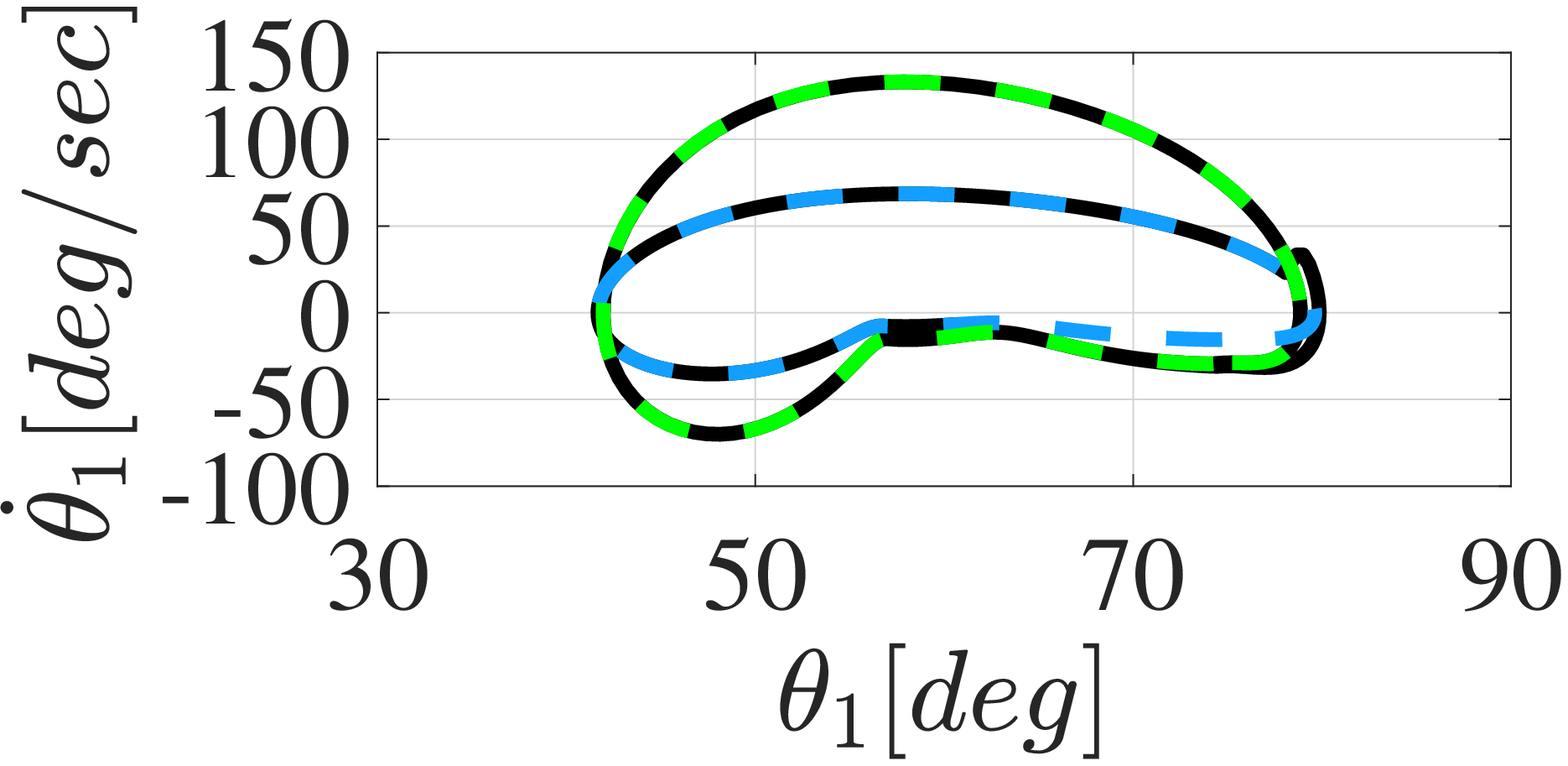}
		\includegraphics[width=0.5\linewidth]{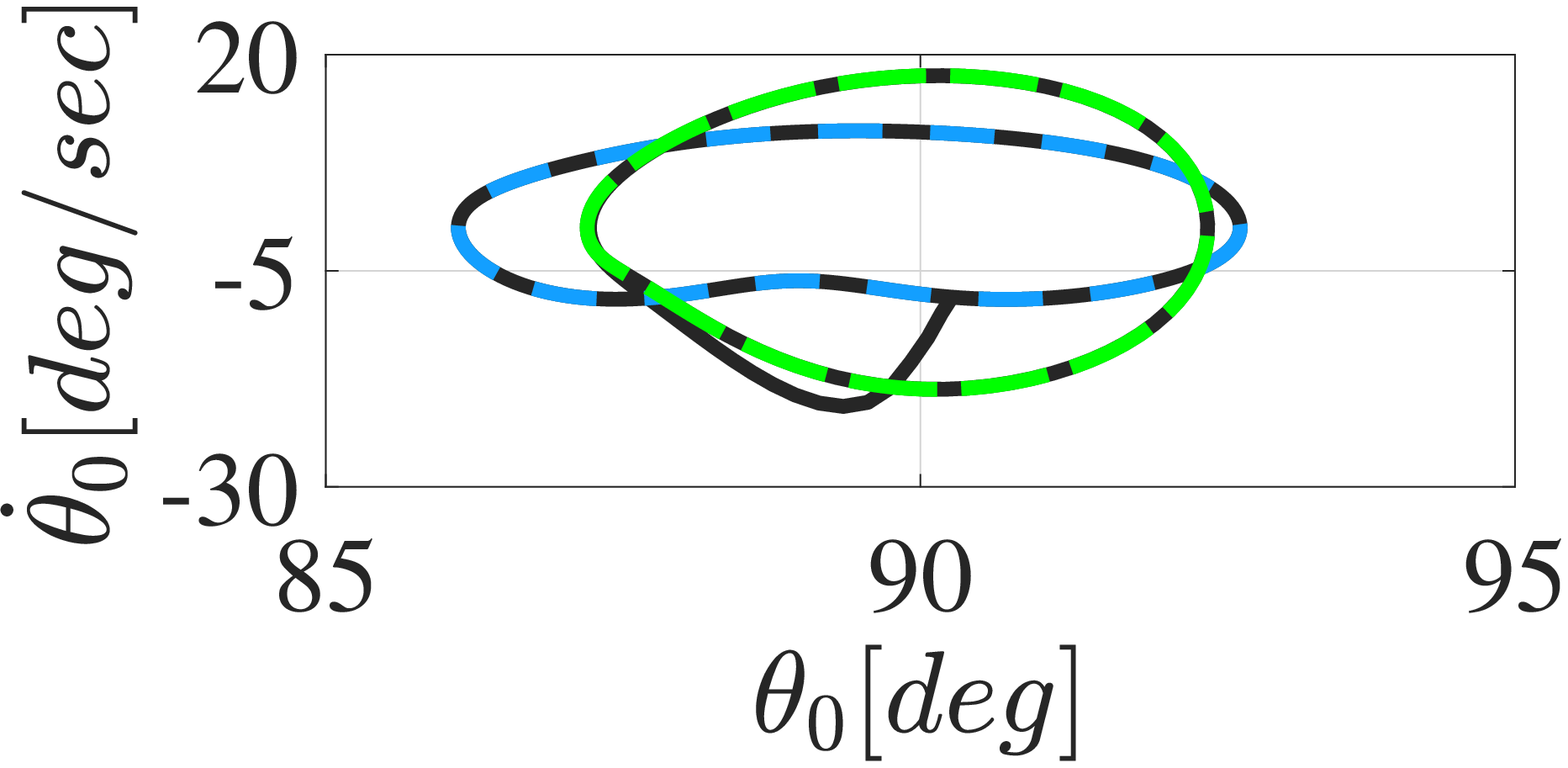}
	}\\
	\subfloat[$t \in \left(20,35\right) (sec)$]{
		\includegraphics[width=0.5\linewidth]{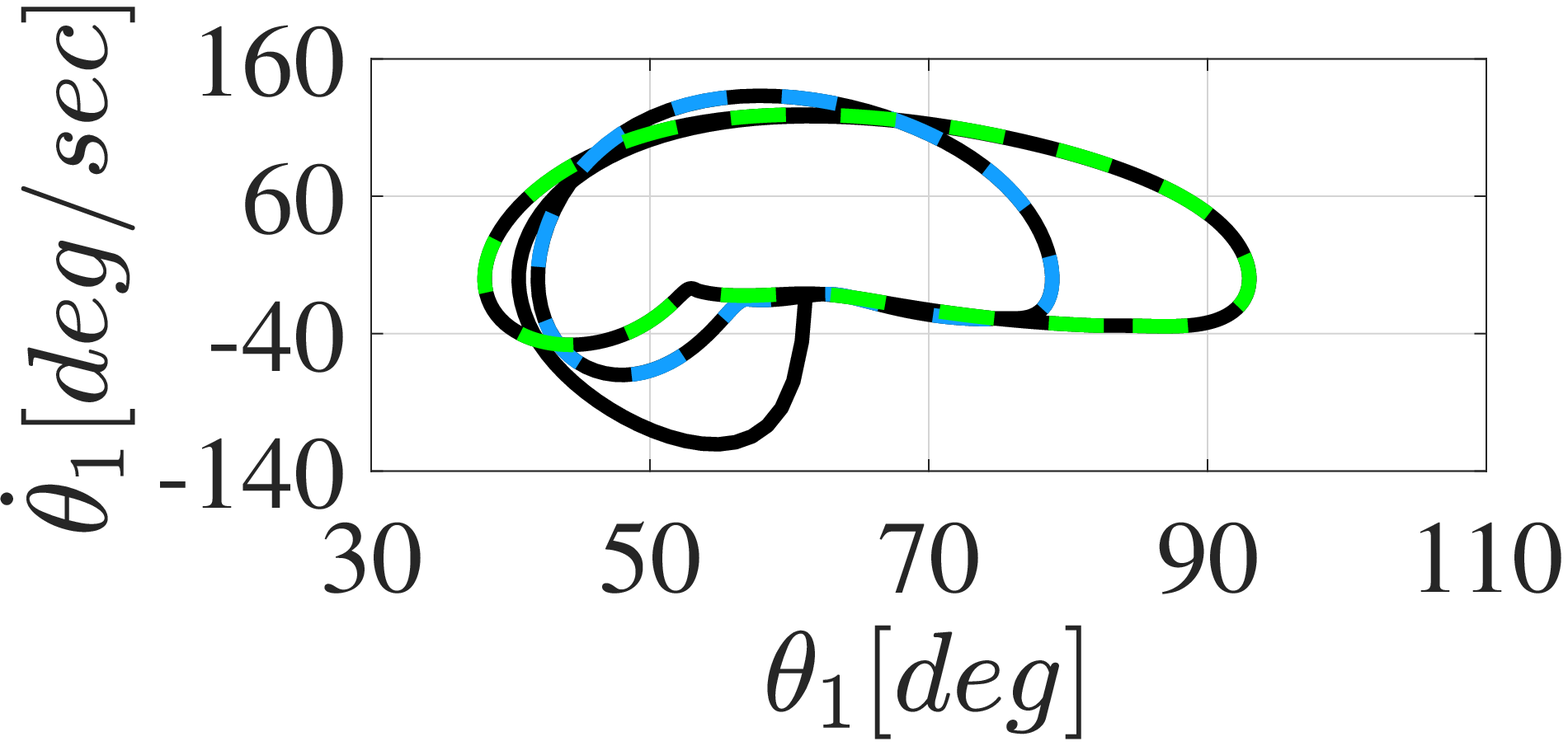}
		\includegraphics[width=0.5\linewidth]{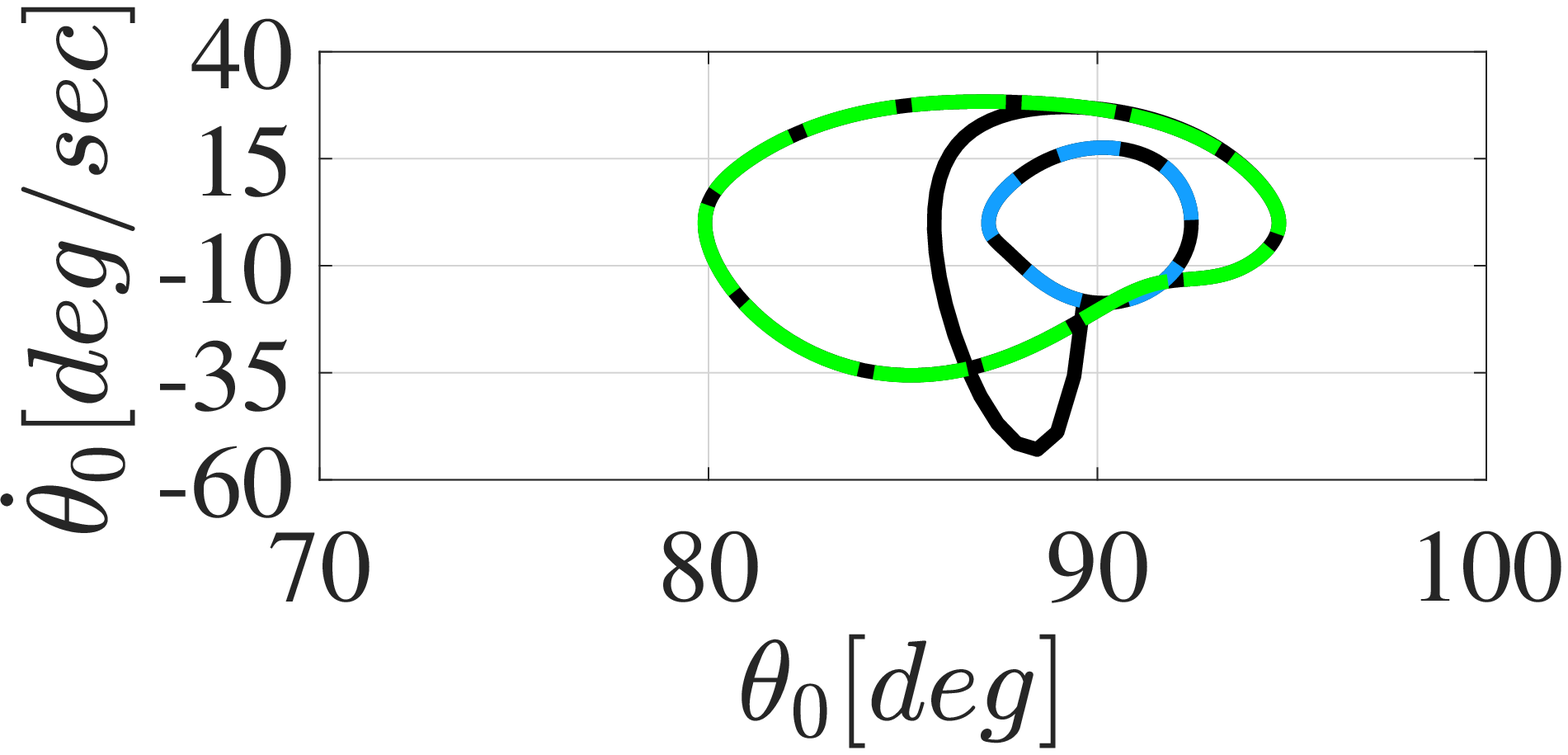}
	}\\
	\subfloat[$t \in \left(35,45\right) (sec)$]{
		\includegraphics[width=0.5\linewidth]{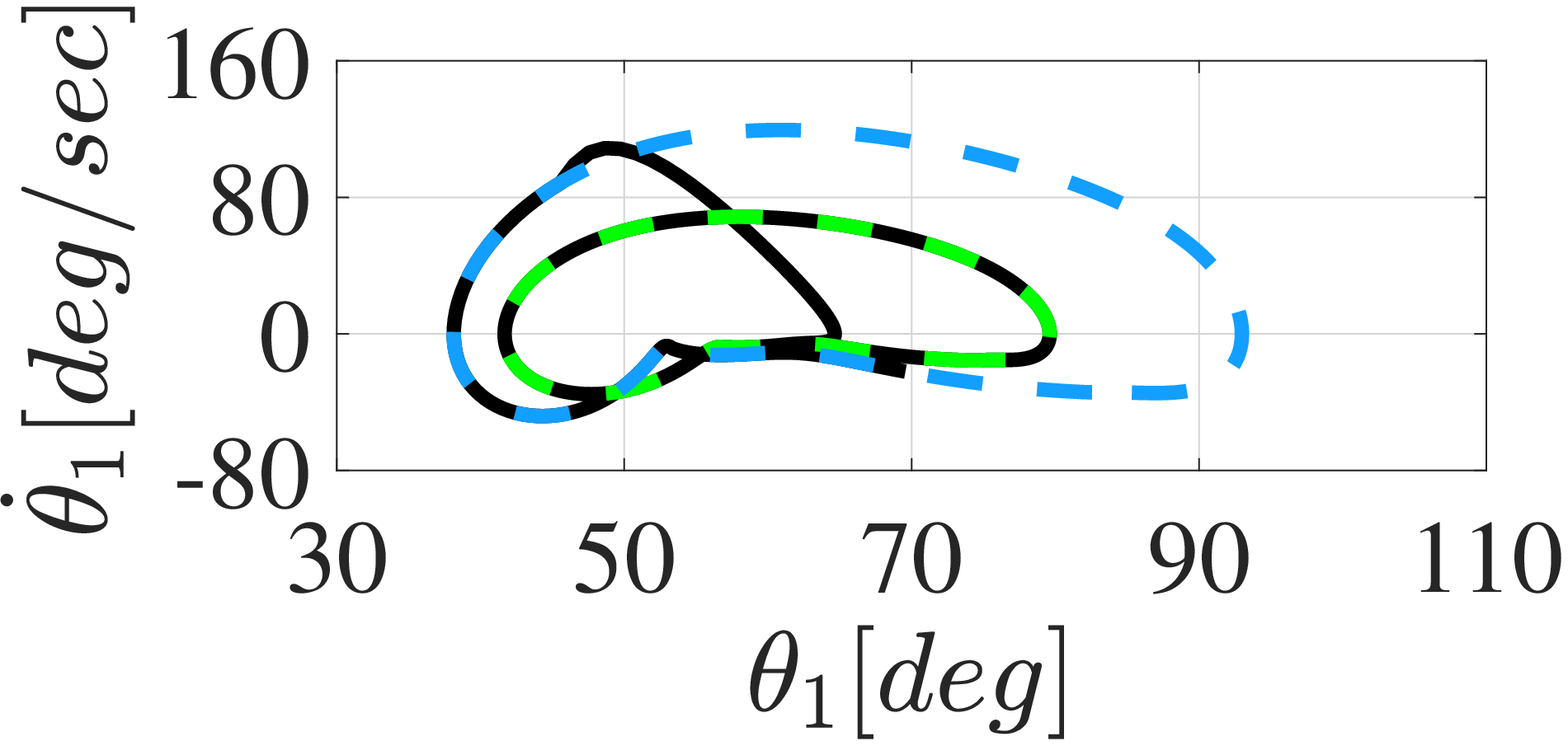}
		\includegraphics[width=0.5\linewidth]{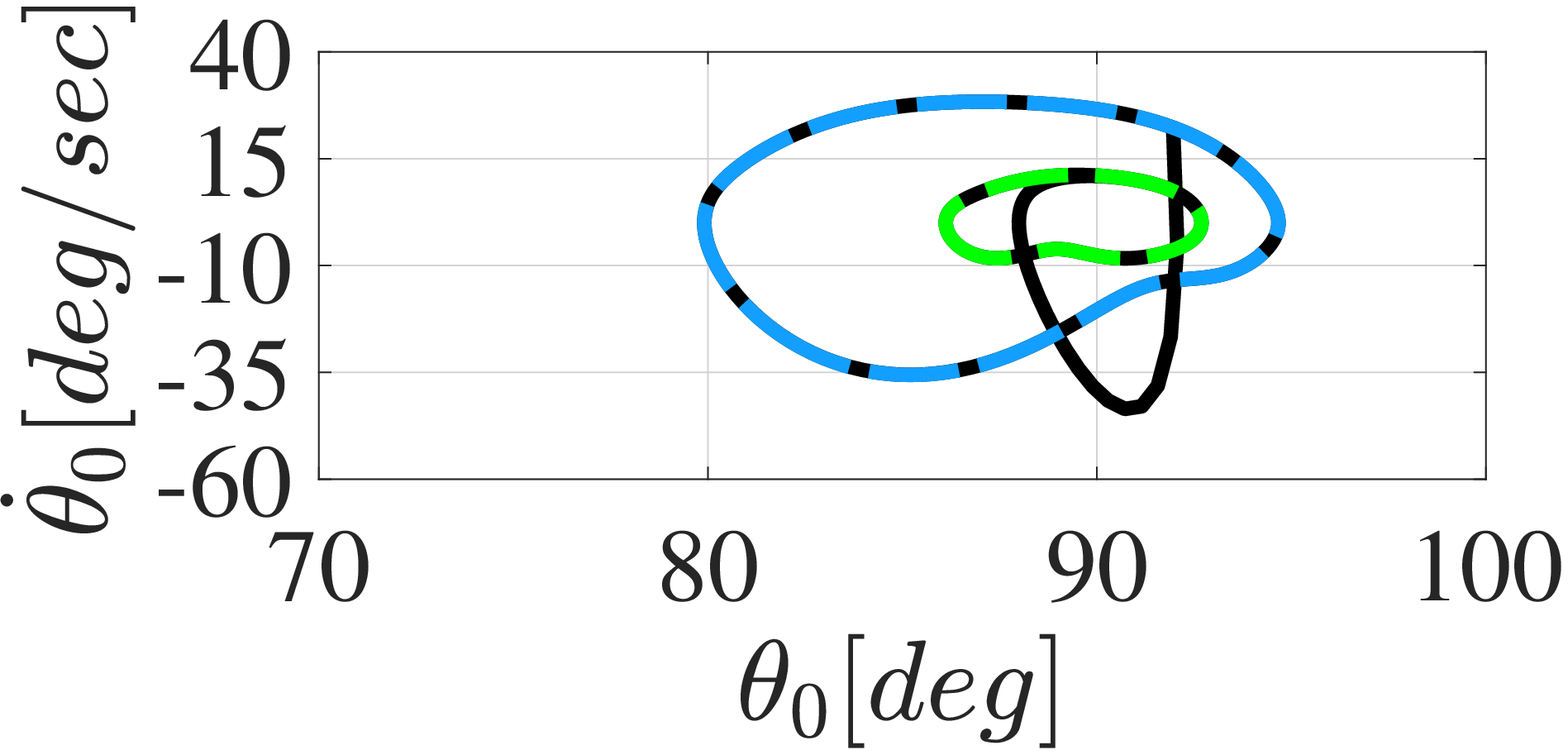}
	}\\
	\subfloat[$t \in \left(45,55\right) (sec)$]{
		\includegraphics[width=0.5\linewidth]{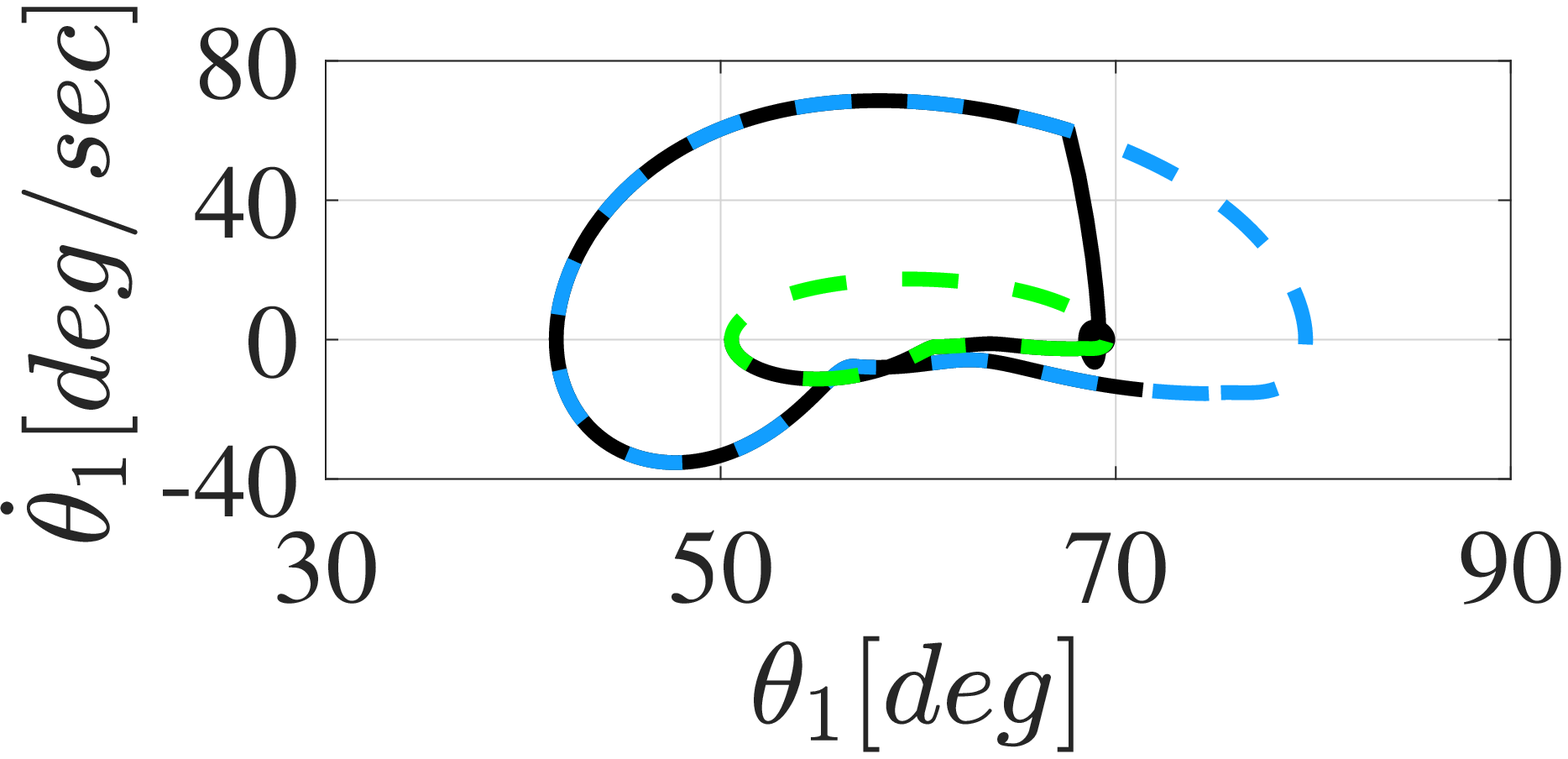}
		\includegraphics[width=0.5\linewidth]{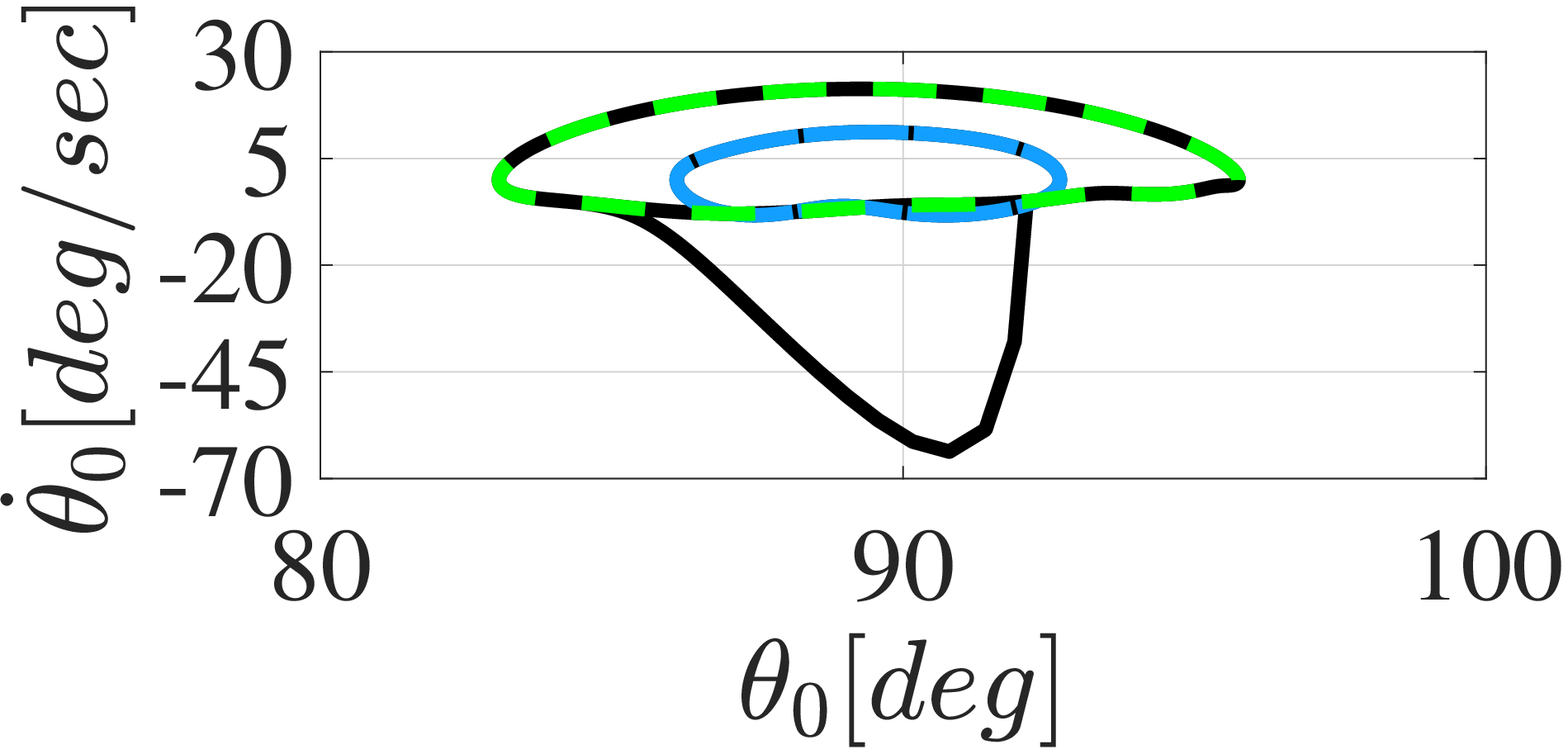}
	}\\
	\subfloat[$t \in \left(55,65\right) (sec)$]{
		\includegraphics[width=0.5\linewidth]{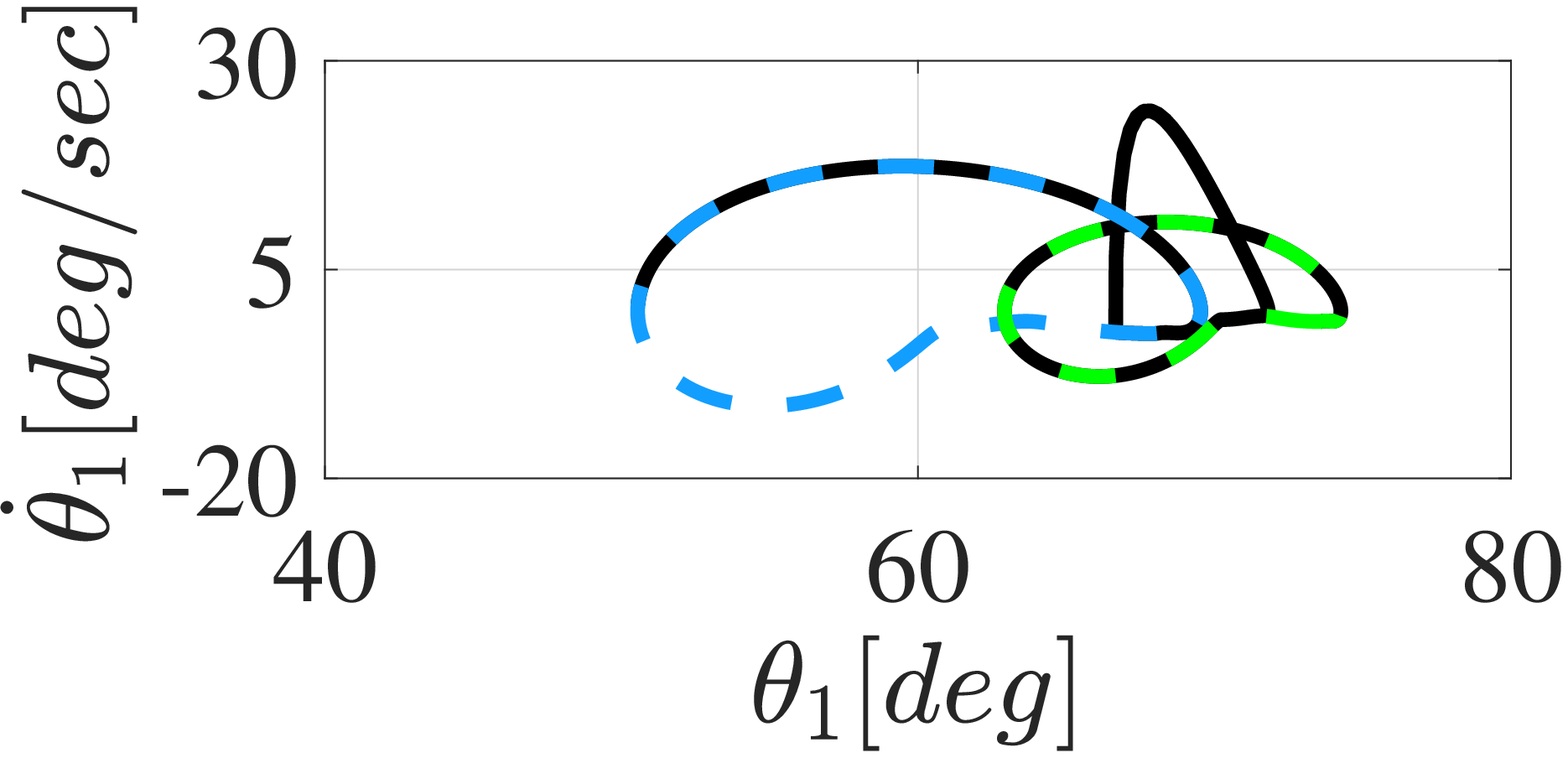}
		\includegraphics[width=0.5\linewidth]{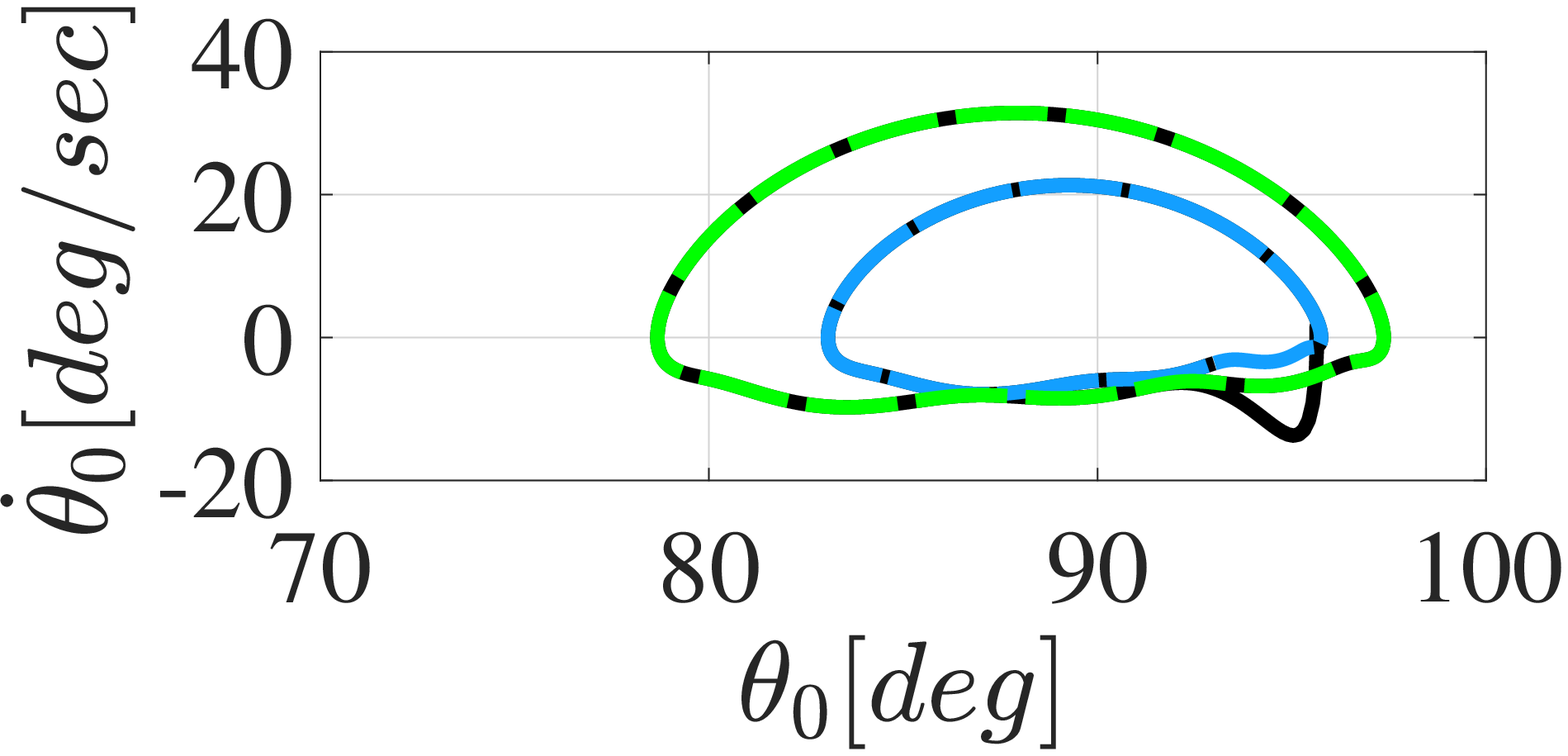}
	}\\
	\subfloat[$t \in \left(65,70\right) (sec)$]{
		\includegraphics[width=0.5\linewidth]{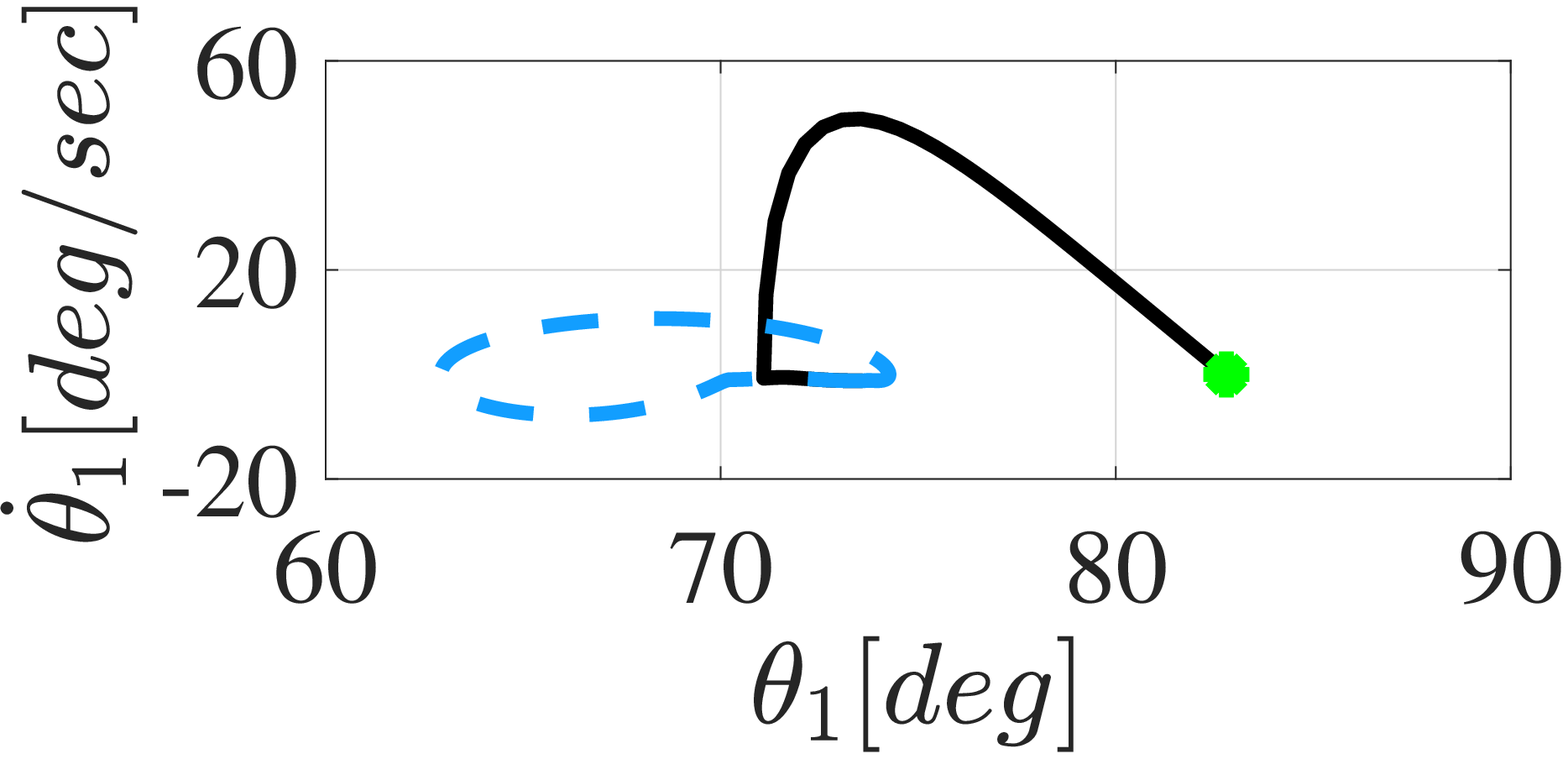}
		\includegraphics[width=0.5\linewidth]{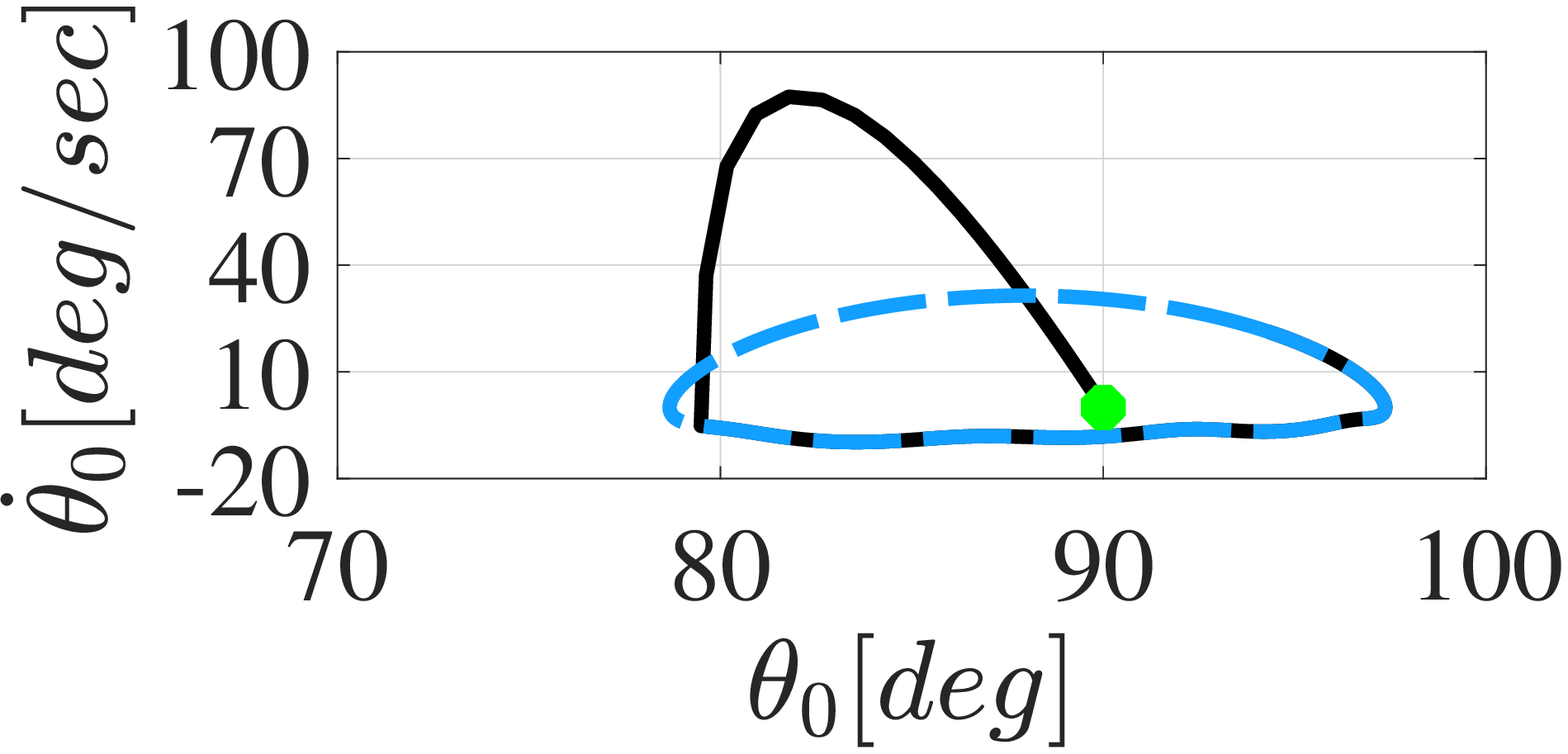}
		\label{fig:sevenLinkWalkingAOS_phasePlot_q1_end}
	}
	\caption{
		The phase space trajectory of $\theta_1$ and $\theta_0$ of the seven link biped robot measured during the walking scenario described in \tablename{~\ref{tab:sevenLinkBiped_highLevelController}}.
		The dashed and dotted curves are the previous and new desired trajectory, respectively.
		The black point is the desired posture for standing.
		The coefficients of the integrated CPG are equal to $B = 40 I_7$, $K = 30 I_7$, $D = 60 I_7$ and $\gamma = 10$.
	}
	\label{fig:sevenLinkWalkingAOS_phasePlot_q1}
\end{figure}

The trajectories of $\theta_1$ and $\theta_0$ in the phase space are shown in \figurename{\ref{fig:sevenLinkWalkingAOS_phasePlot_q1}}.
The blue and green curves denote the previous and new desired trajectories for the motion scenario described in \tablename{~\ref{tab:sevenLinkBiped_highLevelController}}.
As can be seen, the trajectories converge smoothly from one desired trajectory to another one, and finally, converge to the desired standing configuration (the green points in \figurename{\ref{fig:sevenLinkWalkingAOS_phasePlot_q1_end}}).

\figurename{\ref{fig:sevenLinkWAlkingAOS_phaseState} shows the time evolution of $\varphi -t$.
For each desired trajectory, $\varphi - t$ converges to a constant value which means that $\varphi$ converges to $t+\Delta$ where $\Delta$ is a constant value.
For each desired trajectory, $\Delta$ is different.
Moreover, for the time intervals $\left[0,12\right] (sec)$ and $\left[37,48\right](sec)$, $\Delta$ are different ($\Delta$ is equal to $-0.53 (sec)$ and $-0.78 (sec)$, respectively), though according to \tablename{~\ref{tab:sevenLinkBiped_highLevelController}}, the desired trajectory in these two time intervals is the same.
In fact, $\Delta$ depends on the desired trajectory, the system convergent coefficients $(B, K, D)$, and the value of the shape states vector when the desired trajectory changes.
Note that $\Delta$ is not necessarily equal to zero, and thus $\varphi$ does not necessarily converge to $t$, since in this simulation, the integrated CPG is in AOS mode, and hence provides the limit cycle tracking of the desired trajectory.

\begin{figure}
	\centering
	\includegraphics[width=0.9\linewidth]{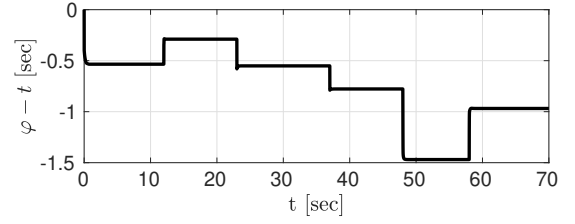}
	\caption{The difference between $\varphi$ and $t$ measured during the walking scenario for the seven link biped robot. The coefficients of the integrated CPG are equal to $B = 40I_7$, $K = 30 I_7$, $D = 60 I_7$, and $\gamma = 10$.}
	\label{fig:sevenLinkWAlkingAOS_phaseState}
\end{figure}

The snapshots of the seven-link biped in the walking simulation are depicted in \figurename{\ref{fig:sevenLinkWalkingAOS_snapshot}}.
The robot converges from one desired motion to another one while the postural stability of the robot is preserved.
For investigating the postural stability of the robot, the time evolution of $zmp_x$ during the walking simulation is depicted in \figurename{\ref{fig:sevelLinkBipedAOS_zmp}}.
The $zmp_x$ location is as \cite{farzaneh_online_2014}
\begin{equation}
zmp_x = \dfrac{\sum_{i = 1}^{7} m_i \left(\left(\ddot{\overline{y}}_i + g\right)\overline{x}_i - \ddot{\overline{x}}_i \, \overline{y}_i\right)}{\sum_{i = 1}^{7} m_i \left(\ddot{\overline{y}}_i + g\right)},
\end{equation}
where $m_i$ is the mass of link $i$, and $\overline{x}_i$ and $\overline{y}_i$ are the horizontal and vertical position of the COM of link $i$.
In \figurename{\ref{fig:sevelLinkBipedAOS_zmp}}, the support polygon in the walking direction is also depicted.
Note that, $zmp_x$ is within the support polygon when the robot follows the desired trajectory and also during the transitions.
Hence, as expected from \figurename{\ref{fig:sevenLinkWalkingAOS_snapshot}}, the postural stability of the robot is preserved.
As mentioned before, the desired trajectories provided from the motion library are designed based on ZMP criteria, and thus for the desired trajectories $zmp_x$ is inside the support polygon and the postural stability of the robot is ensured.
Providing that $zmp_x$ is inside the support polygon in the transient motion from one desired trajectory to another one, the postural stability is also preserved in the transient motion.
Since the desired trajectories satisfy the ZMP criterion, the ZMP criterion is also ensured in a neighborhood of the desired trajectories according to the continuity of the biped dynamics.
Therefore, for the cases where the difference between the previous and new desired trajectories is small, the region between these two trajectories is a subset of the union of the neighborhood of the desired trajectories where the ZMP criterion is satisfied.
%Accordingly, the ZMP criteria is ensured in the region between the previous and new desired trajectories.
On the other hand, in this simulation, $\gamma$ equals $10$, and the desired trajectory is AOS according to Theorem \ref{thm:BMDVO}.
Thus, the transient trajectory from the previous desired trajectory to the new one remains in the region between the two desired trajectories in the state space.
%Thus, in this simulation, the transient motion between two desired trajectories satisfies the ZMP criteria and thus preserves the robot postural stability if the difference between the previous and new desired trajectories is small enough.
Therefore, for preserving the postural stability of the robot, in this simulation, we change the desired step length and time gradually.
For instance, for reducing the desired step length from $80 (cm)$ to $20 (cm)$, we first change the step length to $50 (cm)$, and then to $20 (cm)$ (see \tablename{~\ref{tab:sevenLinkBiped_highLevelController}).

\begin{figure}[!t]
	\centering
	\captionsetup[subfigure]{labelformat=empty,width=\linewidth}
	\subfloat[from standing to walk with $L = 50 (cm) ,T = 2 (sec)$]{
		\hspace{-0.2cm}
		\captionsetup[subfigure]{width=0.1\linewidth}
		\subfloat[$0.0$]{
			\includegraphics[width=0.1\linewidth]{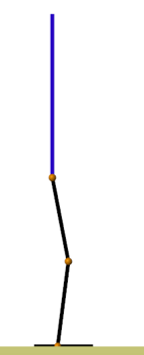}
		}\hspace{-0.4cm}
		\subfloat[$0.1$]{
			\includegraphics[width=0.1\linewidth]{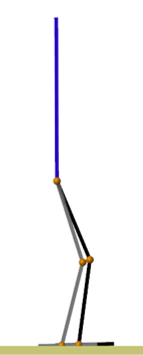}
		}\hspace{-0.4cm}
		\subfloat[$0.2$]{
			\includegraphics[width=0.1\linewidth]{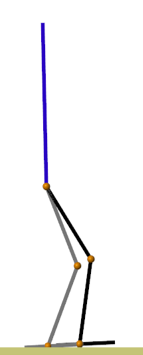}
		}\hspace{-0.4cm}
		\subfloat[$0.3$]{
			\includegraphics[width=0.1\linewidth]{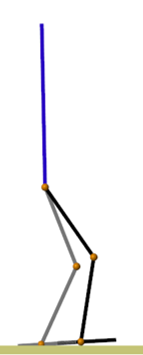}
		}\hspace{-0.4cm}
		\subfloat[$0.4$]{
			\includegraphics[width=0.1\linewidth]{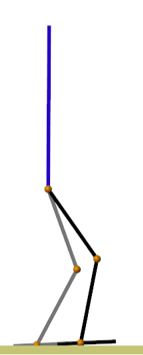}
		}\hspace{-0.4cm}
		\subfloat[$0.5$]{
			\includegraphics[width=0.1\linewidth]{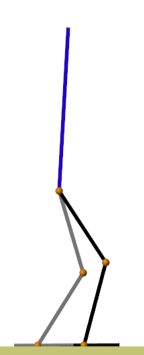}
		}\hspace{-0.4cm}
		\subfloat[$0.6$]{
			\includegraphics[width=0.1\linewidth]{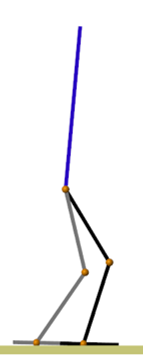}
		}\hspace{-0.4cm}
		\subfloat[$0.7$]{
			\includegraphics[width=0.1\linewidth]{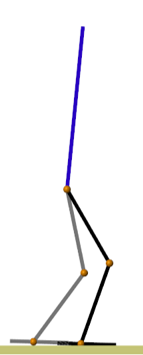}
		}\hspace{-0.4cm}
		\subfloat[$0.8$]{
			\includegraphics[width=0.1\linewidth]{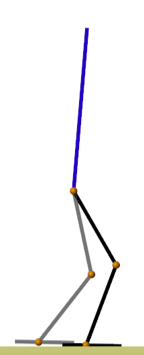}
		}\hspace{-0.4cm}
		\subfloat[$0.9$]{
			\includegraphics[width=0.1\linewidth]{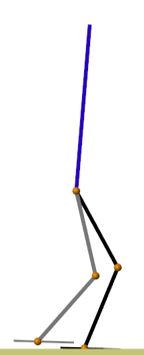}
		}
	}\\
	\subfloat[from walking with the step time $T = 2 (sec)$ to $T = 1 (sec)$]{
		\hspace{-0.2cm}
		\captionsetup[subfigure]{width=0.1\linewidth}
		\subfloat[$12.0$]{
			\includegraphics[width=0.1\linewidth]{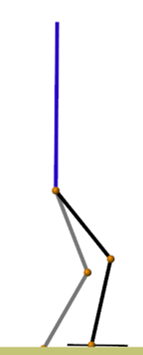}
		}\hspace{-0.4cm}
		\subfloat[$12.1$]{
			\includegraphics[width=0.1\linewidth]{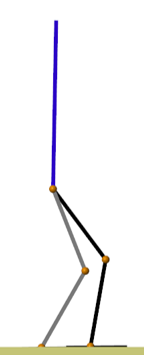}
		}\hspace{-0.4cm}
		\subfloat[$12.2$]{
			\includegraphics[width=0.1\linewidth]{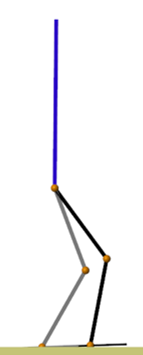}
		}\hspace{-0.4cm}
		\subfloat[$12.3$]{
			\includegraphics[width=0.1\linewidth]{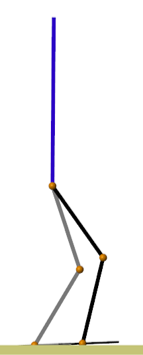}
		}\hspace{-0.4cm}
		\subfloat[$12.4$]{
			\includegraphics[width=0.1\linewidth]{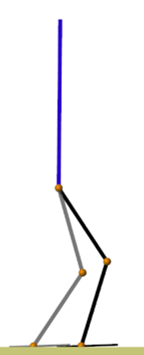}
		}\hspace{-0.4cm}
		\subfloat[$12.5$]{
			\includegraphics[width=0.1\linewidth]{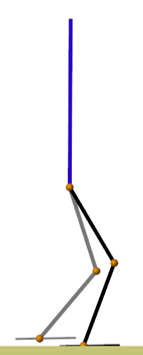}
		}\hspace{-0.4cm}
		\subfloat[$12.6$]{
			\includegraphics[width=0.1\linewidth]{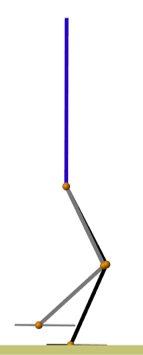}
		}\hspace{-0.4cm}
		\subfloat[$12.7$]{
			\includegraphics[width=0.1\linewidth]{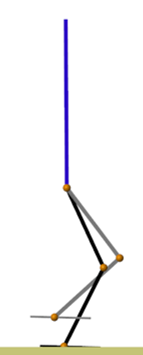}
		}\hspace{-0.4cm}
		\subfloat[$12.8$]{
			\includegraphics[width=0.1\linewidth]{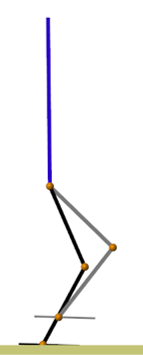}
		}\hspace{-0.4cm}
		\subfloat[$12.9$]{
			\includegraphics[width=0.1\linewidth]{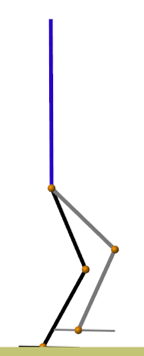}
		}
	}\\
	\subfloat[from $L = 50 (cm)$ and $T = 1 (sec)$ to $L = 80 (cm)$ and $T = 1.7 (sec)$]{
		\hspace{-0.2cm}
		\captionsetup[subfigure]{width=0.1\linewidth}
		\subfloat[$23.0$]{
			\includegraphics[width=0.1\linewidth]{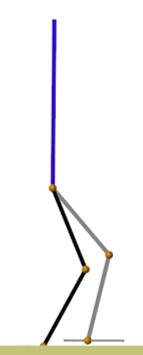}
		}\hspace{-0.4cm}
		\subfloat[$23.1$]{
			\includegraphics[width=0.1\linewidth]{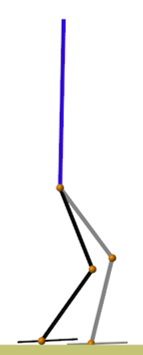}
		}\hspace{-0.4cm}
		\subfloat[$23.2$]{
			\includegraphics[width=0.1\linewidth]{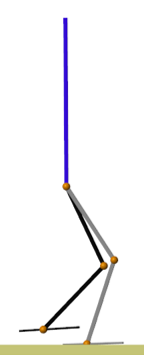}
		}\hspace{-0.4cm}
		\subfloat[$23.3$]{
			\includegraphics[width=0.1\linewidth]{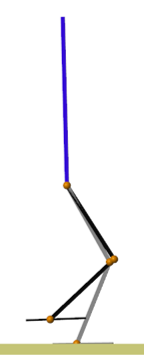}
		}\hspace{-0.4cm}
		\subfloat[$23.4$]{
			\includegraphics[width=0.1\linewidth]{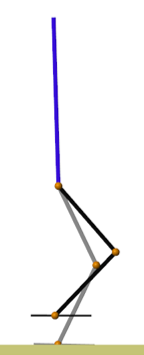}
		}\hspace{-0.4cm}
		\subfloat[$23.5$]{
			\includegraphics[width=0.1\linewidth]{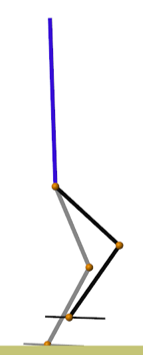}
		}\hspace{-0.4cm}
		\subfloat[$23.6$]{
			\includegraphics[width=0.1\linewidth]{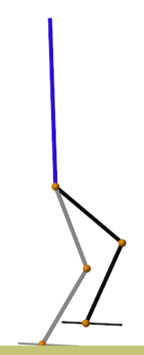}
		}\hspace{-0.4cm}
		\subfloat[$23.7$]{
			\includegraphics[width=0.1\linewidth]{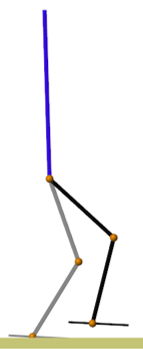}
		}\hspace{-0.4cm}
		\subfloat[$23.8$]{
			\includegraphics[width=0.1\linewidth]{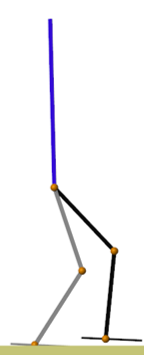}
		}\hspace{-0.4cm}
		\subfloat[$23.9$]{
			\includegraphics[width=0.1\linewidth]{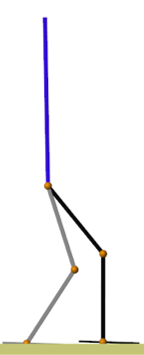}
		}
	}\\
	\subfloat[from walking with $L = 10 (cm)$ and $T = 4 (sec)$ to standing]{
		\hspace{-0.2cm}
		\captionsetup[subfigure]{width=0.1\linewidth}
		\subfloat[$68.0$]{
			\includegraphics[width=0.1\linewidth]{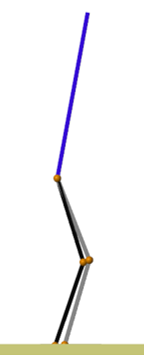}
		}\hspace{-0.4cm}
		\subfloat[$68.1$]{
			\includegraphics[width=0.1\linewidth]{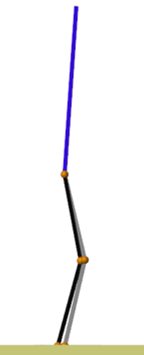}
		}\hspace{-0.4cm}
		\subfloat[$68.2$]{
			\includegraphics[width=0.1\linewidth]{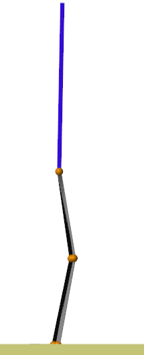}
		}\hspace{-0.4cm}
		\subfloat[$68.3$]{
			\includegraphics[width=0.1\linewidth]{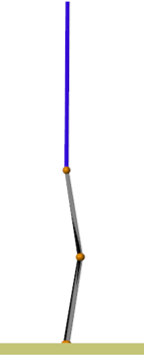}
		}\hspace{-0.4cm}
		\subfloat[$68.4$]{
			\includegraphics[width=0.1\linewidth]{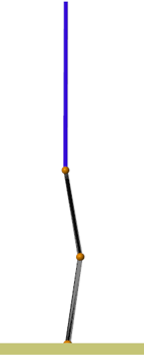}
		}\hspace{-0.4cm}
		\subfloat[$68.5$]{
			\includegraphics[width=0.1\linewidth]{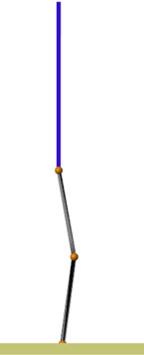}
		}
	}
	\caption{
		The snapshots of the simulation of the seven link biped in MATLAB.
		The time instance of the snapshots is also reported in second.
		The figures show the transient motion from standing to walking, during the change in the step period/length of walking, and finally the transition to standing based on the  scenario described in \tablename{~\ref{tab:sevenLinkBiped_highLevelController}}.
		The parameters of the CPG are selected as $B = 40 I_7$, $K = 30 I_7$, $D = 60 I_7$ and $\gamma = 10$.
	}
	\label{fig:sevenLinkWalkingAOS_snapshot}
\end{figure}

For comparing the trajectory tracking and the  limit cycle tracking of the desired trajectory, the seven-link biped walking is simulated for the integrated CPG in the AS and AOS modes (\ie $\gamma = 0$ and $\gamma = 10$). The results for the transition motion from standing to walking with the step length of $50 (cm)$ and step period of $2 (sec)$ are shown in \figurename{\ref{fig:sevenLinkBiped_compareASandAOS}}.
The CPG parameters are selected such that the control effort for converging to the desired walking trajectory is the same for both modes.
The results reveal that for $\gamma = 10$, the robot converges to the desired walking trajectory faster than the case where $\gamma = 0$.
Thus, the limit cycle tracking of the desired trajectory provides a higher convergence rate with the same control effort in comparison with the trajectory tracking of the desired trajectory.

\begin{figure}[!t]
	\centering
	\includegraphics[width=0.95\linewidth]{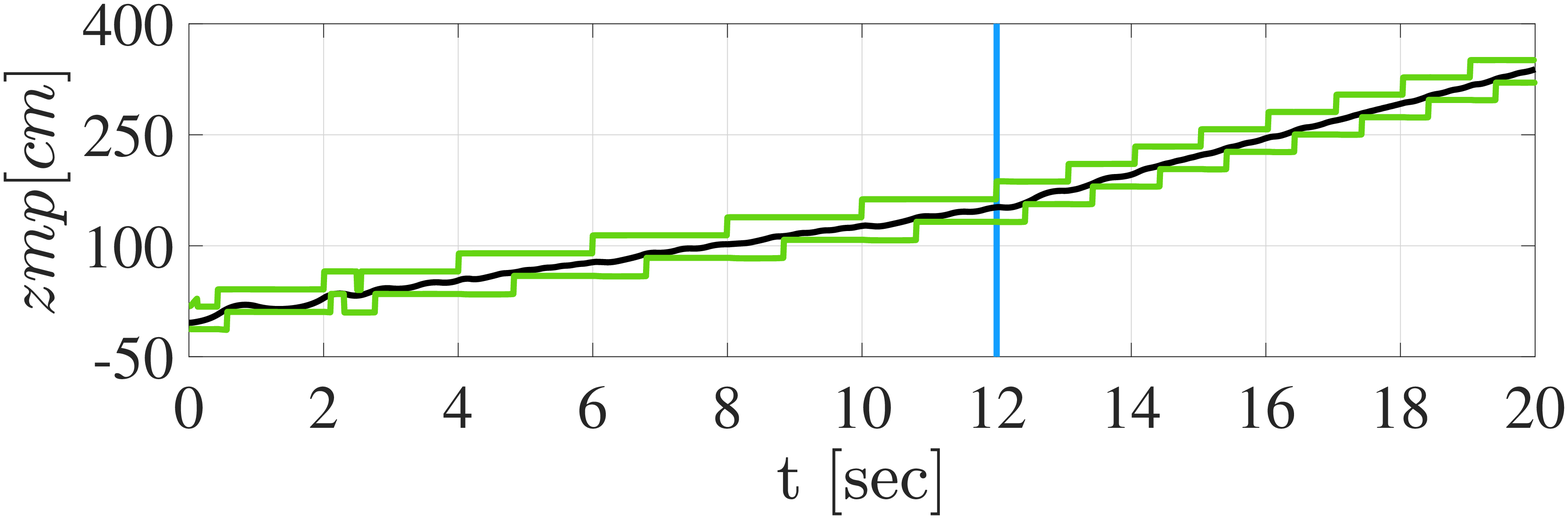}\\
	\includegraphics[width=0.95\linewidth]{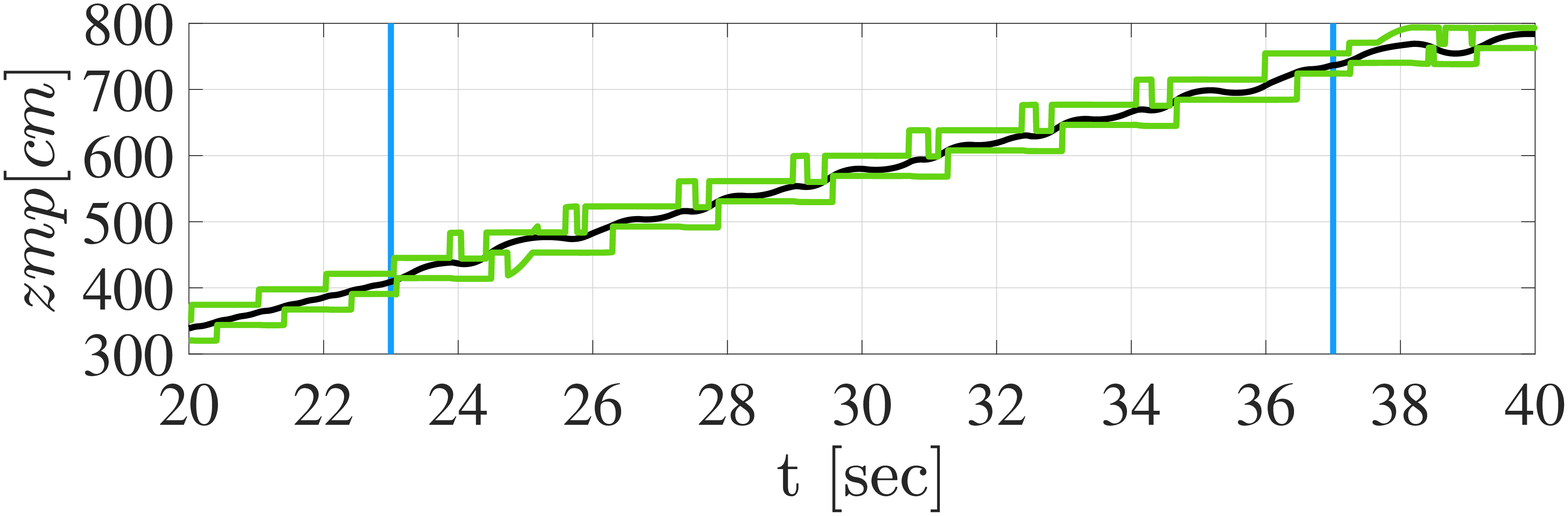}\\
	\includegraphics[width=0.95\linewidth]{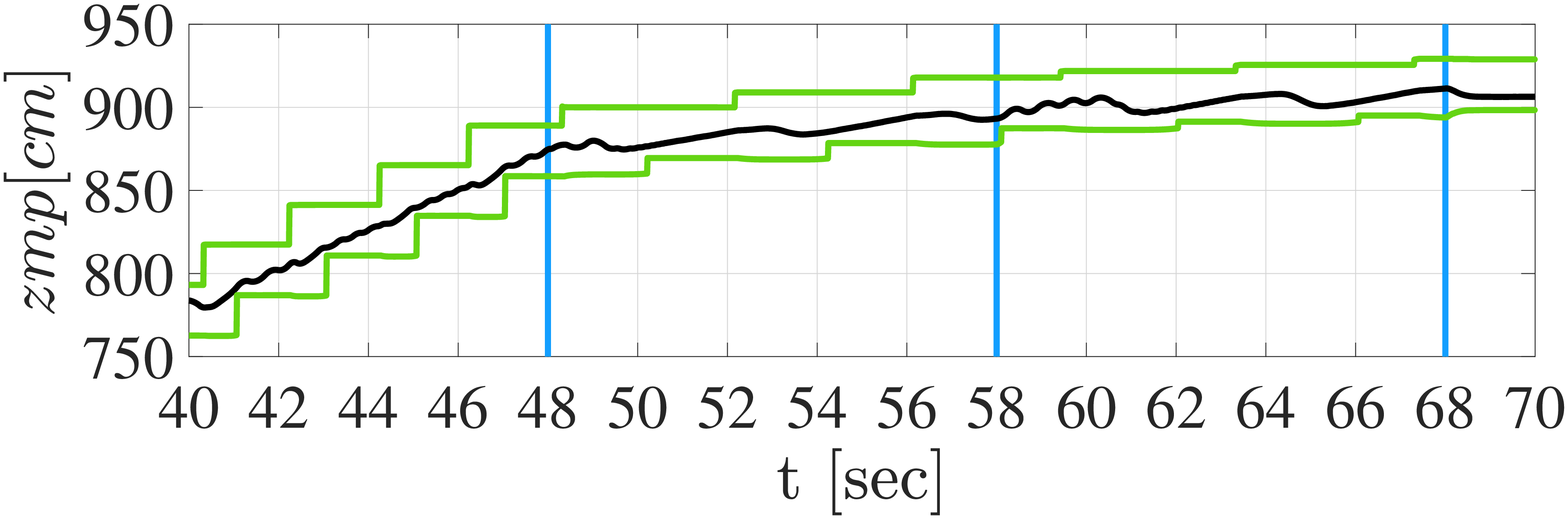}
	\caption{
		The time evolution of ZMP in the walking direction.
		The green curves denote the lower and upper bounds of the support polygon, and the blue lines show the time instance that the desired motion changes.
		The coefficients of the CPG are selected as $B = 40 I_7$,
		$K = 30 I_7$,
		$D = 60 I_7$ and $\gamma = 10$.
	}
	\label{fig:sevelLinkBipedAOS_zmp}
\end{figure}

\begin{figure}[!b]
	\centering
	\subfloat{
		\includegraphics[width=0.43\linewidth]{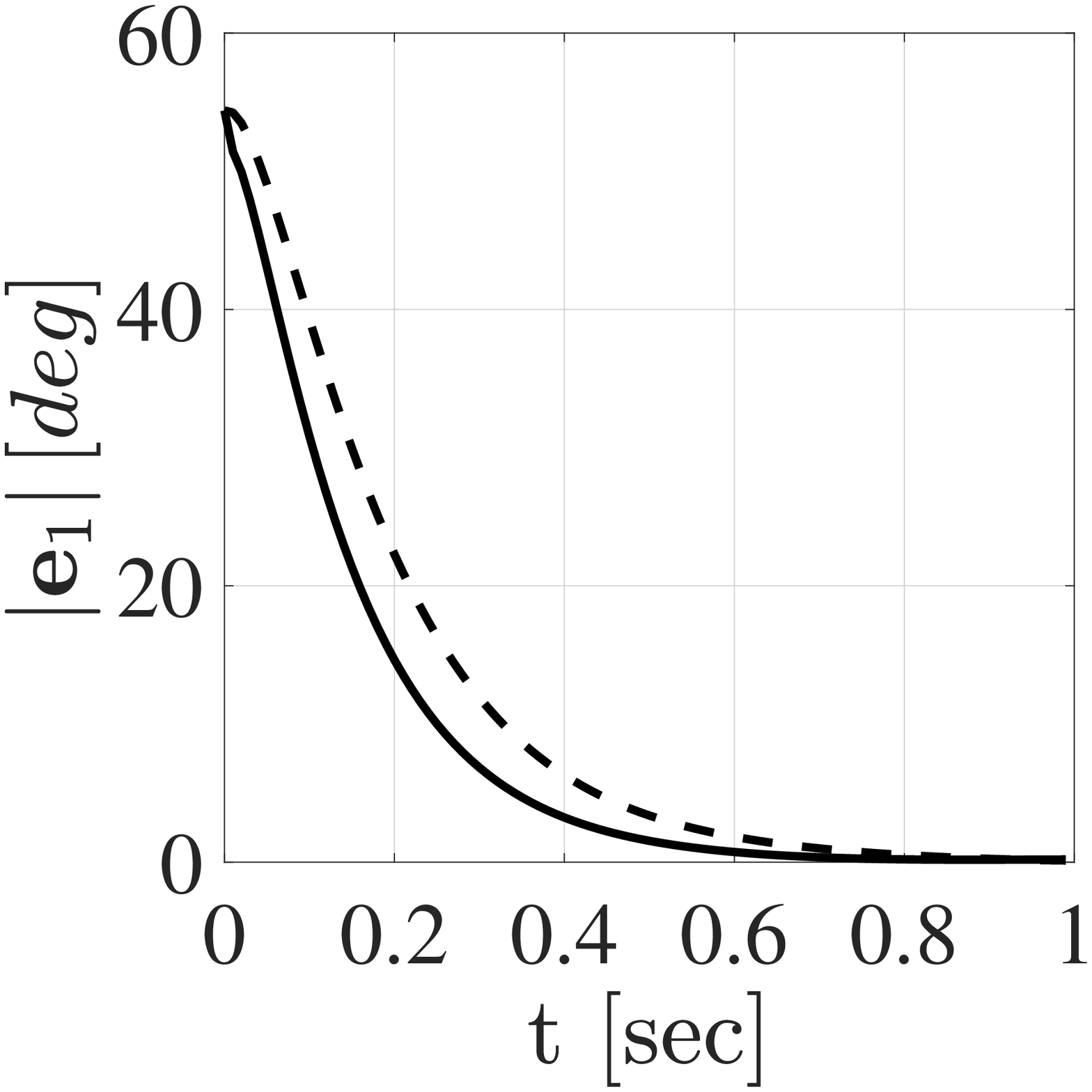}
		\includegraphics[width=0.43\linewidth]{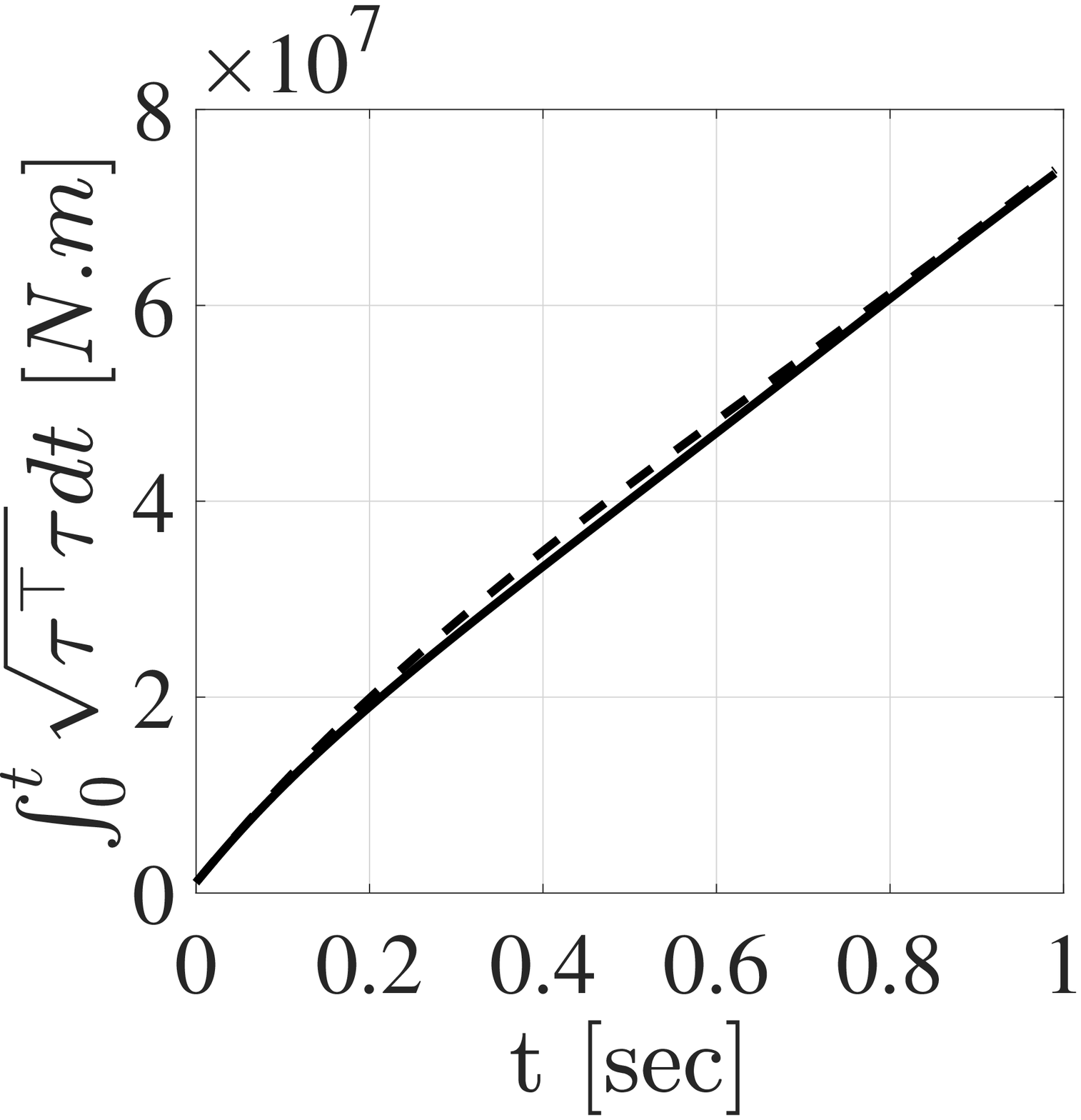}
	}
	\caption{Comparing the convergence rate and control effort of the seven-link biped controlling by the integrated CPG in AS and AOS modes during the transition motion from standing to walking with step length of $50 (cm)$ and step time of $2 (sec)$.
		The left and right figures depict $\left\|\be_1\right\| = \left\| \bs_1-\bg_p\right\|$ and the of $\left\|\tau\right\|$ where $\tau$ is the vector of joint actuator torques of the robot.}
	\label{fig:sevenLinkBiped_compareASandAOS}
\end{figure}
\section{Conclusion}
\label{sec:conclusion}

This paper presents a programmable oscillator that tracks multidimensional periodic trajectories.
The programmable oscillator ensures the global stability and convergence of the desired trajectory and can also provide both asymptotic stability and asymptotic orbital stability of the desired trajectory according to an oscillator parameter.
The programmable oscillator is also modified by a parametrization technique for bounding the output signal and its first time derivative.
We used Lyapunov analysis to prove the global convergence to the desired trajectory for both original and modified oscillators.
Both oscillators feature a parameter that controls the convergence to achieve either asymptotic stability (trajectory tracking) or asymptotic orbital stability (limit cycle tracking) of the desired trajectory.
Finally, we presented an integrated CPG architecture based on the bounded output programmable oscillator.
The proposed architecture is specifically designed for robotic applications where motion modulation is required.
The proposed CPG generates smooth, synchronized, and bounded reference trajectory tracking the desired one from any initial condition irrespective of the parameters of the CPG.
The desired trajectory is a multidimensional function whose components are either periodic functions or constants.
This property is important for those applications where both periodic motions and constant posturing are required.
Moreover, the proposed CPG can provide both trajectory tracking and limit cycle tracking of the desired trajectory.
Accordingly, one can use the proposed CPG in the automatic or imitating modes.
In the automatic mode, the oscillator guarantees
the limit cycle tracking of the desired trajectory provided by
a library. 
In the imitating mode, the CPG guarantees trajectory tracking of the desired trajectory measuring online from a human or another robot.
The proposed CPG is a reliable and programmable architecture for online motion modulation.
We validated the soundness of the proposed CPG with robotic rehabilitation experiments on the Kuka iiwa robot arm, and also walking simulations on a seven-link biped.
% or


\appendices
\section{Proof of Theorem \ref{thm:stabilityOfProgrammableOscillator}}
\label{app:MDVOstability}

Theorem \ref{thm:stabilityOfProgrammableOscillator} has 3 items that we prove them in the following.

%%%%%%%%%%%%%%%%%%%%%%%%%%%%%%%%%%%%%%%%%%%%%

\subsection{$\bx(t)$ is bounded and converges to $\bq_d (\varphi)$ from ...}

Starting with
\begin{equation}
\label{eq:Lyapunov_1}
V_1 \left( \bx , \varphi \right) = \dfrac{1}{2} \be_{\bs}^\top K \be_{\bs} + \dfrac{1}{2} \be_{\dbs}^\top \be_{\dbs},
\end{equation} 
where $B$ is the PD matrix used in \eqref{eq:MDVO_dynamic} and
\[\left[ \be_{\bs}^\top,\be_{\dbs}^\top \right] = \bx^\top - \bq_d^\top(\varphi) = \left[\bs^\top-\bbf^\top (\varphi) ,\dbs^\top - \bbf^{\prime \top} (\varphi) \right],\]
the time derivative of $V_1$ along the system trajectory \eqref{eq:MDVO_dynamic} is
\begin{equation}
\dot{V}_1 = -\gamma \left( \be_{\bs}^\top K \pbf(\varphi) + \be_{\dbs}^\top \ppbf (\varphi) \right)^2 - \be_{\dbs}^\top B \be_{\dbs}.
\end{equation}
Since $B$ is a PD matrix and $\gamma$ is a nonnegative constant, $\dot{V}_1 \leq 0$.
Moreover, \eqref{eq:Lyapunov_1} shows that $V_1 \geq 0$ where $V_1 = 0$ iff $\left[\be_{\bs},\be_{\dbs}\right] = \left[\mathbf{0}_n,\mathbf{0}_n\right]$.
Hence $\be_{\bs}$ and $\be_{\dbs}$ are bounded.
Assuming that $\bbf(t)$ and its first time derivative are bounded, one also concludes the boundedness of $\bx(t)$ for any initial condition.
To get convergence of $\bq_d(\varphi)$ (or $\left[\be_{\bs},\be_{\dbs}\right] = \left[\mathbf{0}_n,\mathbf{0}_n\right]$), it is sufficient to examine the largest invariant subset of the set $\Omega = \left\{\left[\bx^\top,\varphi\right] : \dot{V}_1 = 0 \right\}$.
Considering \eqref{eq:MDVO_dynamic}, it is possible to verify that $\left\{ \left[\bq_d^\top (\varphi),\varphi \right] \right\}$ is the only invariant subset of $\Omega$.
Therefore, according to the LaSalle Lemma \cite{khalil2002nonlinear}, we can conclude that $\bx$ converges to $\bq_d(\varphi)$.
The global convergence of $\bx$ to $\bq_d(\varphi)$ is the consequence of the radially unbounded property of $V_1$ w.r.t. $\|\bx\|$.

Note that the phase state $\varphi$ is, as expected, an unbounded state.
While, from \eqref{eq:MDVO_phaseDynamic} and the assumption A1.1, one can conclude that $\varphi$ is continuously differentiable, and hence $\varphi$ does not have finite escape time \cite{khalil2002nonlinear}.

%%%%%%%%%%%%%%%%%%%%%%%%%%%%%%%%%%%%%%%%%%%%%%%%%

\subsection{if $\gamma = 0$, then $\left[\bq_d^\top (t + \varphi (0)), t + \varphi (0)\right]^\top$ is globally AS}
\label{sec:appedix_theorem1_partB}

If $\gamma = 0$, then we have $\dot{\varphi} = 1$, and thus $\varphi = t + \varphi(0)$.
Hence, the system \eqref{eq:MDVO_dynamic} is simplified to a $2n$ dimensional system whose state space vector is $\bx$.
Accordingly, the function $V_1$ is the Lyapunov function of the system, and thus the global AS of $\bq_d(t+\varphi(0)) = {\small\left[\bbf^\top (t+\varphi(0)),\bbf^{\prime \top} (t+\varphi(0))\right]^\top}$ is concluded based on the Lyapunov stability theorem \cite{khalil2002nonlinear}.

%%%%%%%%%%%%%%%%%%%%%%%%%%%%%%%%%%%%%%%%%%%%%%%%%%

\subsection{if $\gamma \rightarrow \infty$, then $\bx = \bq_d (t)$ is the globally stable limit cycle}

The limit cycle tracking and the trajectory tracking concepts are the same for a constant desired trajectory (see Section \ref{sec:background_stability}).
The trajectory tracking of any desired trajectory including a constant one is proved in part \ref{sec:appedix_theorem1_partB}.
In the following, we discuss the limit cycle tracking of a non-constant desired trajectory.

To investigate the stability of the set $\bq_d(t)$, we prove that for $\gamma \rightarrow \infty$,
\begin{itemize}
	\item $\forall \left[\bx^\top,\varphi\right] \in \mathcal{P}_{\bbf}, V_1(\bx, \varphi) = 0$,
	 \item $\forall \left[\bx^\top,\varphi\right] \notin \mathcal{P}_{\bbf}, V_1(\bx, \varphi) > 0$, and
	 \item $\forall \left[\bx^\top,\varphi\right] \notin \mathcal{P}_{\bbf}, \dot{V}_1(\bx, \varphi) < 0$,
\end{itemize}
where $\mathcal{P}_{\bbf}$ is the set of the states belonging to the trajectory of $\bq_d (t)$, \ie
\begin{equation}
\mathcal{P}_{\bbf} = \left\{ \left[ \bs^\top, \dbs^\top, \varphi \right] \, : \, \exists \varphi, \, \bs=\bbf(\varphi),\dbs=\pbf(\varphi) \right\}.
\end{equation}
Thus, one can conclude that $\mathcal{P}_{\bbf}$ is the stable limit cycle of the system \eqref{eq:MDVO_dynamic} according to the extension of the Lyapunov stability theorem for invariant sets \cite{zubov1964methods}.
The global stability of the limit cycle $\mathcal{P}_{\bbf}$ is the result of the radially unbounded property of $V_1$.
For this purpose, we prove that for $\gamma \rightarrow \infty$, $\left[\be_{\bs},\be_{\dbs}\right] = \left[\mathbf{0}_n,\mathbf{0}_n\right]$ iff $\left[\bx^\top,\varphi\right] \in \mathcal{P}_{\bbf}$.
Therefore, the Lyapunov function presented in \eqref{eq:Lyapunov_1} satisfies the above three conditions, and we can conclude that $\bq_d(t)$ is the globally stable limit cycle of the system.
Since $\be_{\bs} = \bs - \bbf(\varphi)$ and $\be_{\dbs} = \dbs - \pbf(\varphi)$, one can figure out that if $\be_{\bs}=\be_{\dbs} = \mathbf{0}_n$, then $\left[\bx^\top,\varphi\right] \in \mathcal{P}_{\bbf}$.
To prove that for $\left[\bx^\top,\varphi\right] \in \mathcal{P}_{\bbf}$, we have $\be_{\bs}=\be_{\dbs} = \mathbf{0}_n$, we show that for $\gamma \rightarrow \infty$ and $\left[\bx^\top,\varphi\right] \in \mathcal{N}_\delta$ where $\mathcal{N}_\delta$ is the $\delta$-neighborhood of $\mathcal{P}_{\bbf}$, $\bq_d(\varphi)$ is the point from the curve of $\mathcal{P}_{\bbf}$ in the shape space having the minimum distance from the current shape states.
In this regard, we define the distance between $\bx = {\small \left[\bs^\top,\dbs^\top\right]^\top}$ and the curve of $\mathcal{P}_{\bbf}$ in the shape space as
\begin{equation}
\dist \left( \bx,\mathcal{P}_{\bbf} \right) = \min_{\tau \in [0,T]} \dis(\bx,\bq_d(\tau)).
\end{equation}
where $T \in \mathbb{R}^+$ is the period of $\bq_d(t)$, and
\begin{equation}
\label{eq:EDVO_ptpDistance}
\begin{aligned}
\dis (\bx,\bq_d(\tau)) = &\dfrac{1}{2}\left(\bs - \bbf(\tau)\right)^\top K \left(\bs - \bbf(\tau) \right)\\
& + \dfrac{1}{2}\left(\dbs - \pbf(\tau) \right)^\top \left(\dbs - \pbf(\tau) \right).
\end{aligned}
\end{equation}

In the following, we prove that $\dis(\bx,\bq_d(\varphi)) = \dist(\bx,\mathcal{P}_{\bbf})$ if $\gamma \rightarrow \infty$.
Since $\dis(\bx,\bq_d(\varphi))$ is continuous, its extreme is at $\dfrac{\partial \dis  (\bx,\bq_d(\varphi))}{\partial \varphi} = 0$.
Thus, we consider the first and second derivatives of $\dis (\bx, \bq_d(\varphi))$ w.r.t. the phase state, \ie
{\small
	\begin{equation}
	\label{eq:EDVO_dpp}
	\begin{aligned}
	&\dis_\varphi = \dfrac{\partial \dis  (\bx,\bq_d(\varphi))}{\partial \varphi}\\
	& \quad = - \left(\bs-\bbf\left(\varphi\right)\right)^\top K \pbf(\varphi) - \left( \dbs - \pbf(\varphi) \right)^\top \ppbf(\varphi), \\
	&\dis_{\varphi \varphi} = \dfrac{\partial^2 \dis (\bx,\bq_d(\varphi))}{\partial \varphi^2} = \bbf^{\prime \top}(\varphi) K \bbf^\prime (\varphi) + \bbf^{\prime \prime \top}(\varphi)  \bbf^{\prime \prime}(\varphi)  \\
	& \qquad \quad - \left(\bs-\bbf(\varphi) \right)^\top B \ppbf(\varphi) - \left(\dbs - \pbf(\varphi) \right)^\top \pppbf(\varphi).
	\end{aligned}
	\end{equation}
}
The lower bound of $\dis_{\varphi \varphi}$ is as
{\small
	\begin{equation}
	\label{eq:EDVO_dppLowerBound}
	\begin{aligned}
	\dis_{\varphi\varphi} \geq{} &\bbf^{\prime \top}(\varphi)  K \bbf^\prime(\varphi) + \bbf^{\prime \prime \top}(\varphi) \bbf^{\prime \prime}(\varphi) \\
	&- \left\|K^{1/2} \left(\bs-\bbf(\varphi)\right)\right\| \left\|K^{1/2}\ppbf(\varphi)\right\| \\
	&- \left\|\dbs - \pbf(\varphi)\right\| \left\|\pppbf(\varphi)\right\|.
	\end{aligned}
	\end{equation}
}
On the other hand, considering \eqref{eq:EDVO_ptpDistance}, we have
\begin{equation}
\begin{aligned}
&\left\|K^{1/2} (\bs-\bbf(\varphi))\right\| \leq \sqrt{2 \dis (\bx,\bq_d(\varphi))}, \\
&\left\|\dbs-\pbf(\varphi)\right\| \leq \sqrt{2 \dis (\bx,\bq_d(\varphi))}.
\end{aligned}
\end{equation}
Considering the above relations, \eqref{eq:EDVO_dppLowerBound} is simplified as
\begin{equation}
\label{eq:EDVO_dppLowerBound_simple}
\begin{aligned}
\dis_{\varphi\varphi} \geq \bbf^{\prime \top}(\varphi) B \pbf(\varphi) + \bbf^{\prime \prime \top}(\varphi) \ppbf(\varphi) \\
- \left( \ppbf_m + \pppbf_m \right) \sqrt{2 \dis (\bx,\bq_d(\varphi))},
\end{aligned}
\end{equation}
where $\ppbf_m = \max_\varphi \left(\left\| K^{1/2} \ppbf\right\|\right)$ and $\pppbf_m = \max_\varphi \left(\left\| \pppbf\right\|\right)$.
Note that $K$ is PD, and for a non-constant desired trajectory, $\nexists \varphi$ that $\pbf(\varphi) = \ppbf(\varphi) = \mathbf{0}_n$.
Thus, one can find $\delta \in \mathbb{R}^+$ such that $\forall \left[\bx^\top,\varphi\right] \in \mathcal{N}_\delta$ where $\mathcal{N}_\delta$ is the $\delta$-neighborhood of $\mathcal{P}_{\bbf}$, we have $\dis_{\varphi\varphi} > 0$.
As a result, $\forall \left[\bx^\top,\varphi\right] \in \mathcal{N}_\delta$, $\dis(\bx,\bq_d(\varphi)) = \dist(\bx,\mathcal{P}_{\bbf})$ iff $\dis_{\varphi} = 0$.
To investigate the time evolution of $\dis_\varphi$, we compute its time derivative along the trajectories of \eqref{eq:MDVO_dynamic} as
\begin{equation}
	\label{eq:EDVO_dphiDyn}
	{\footnotesize
	\begin{aligned}
	\dot{\dis} _\varphi = & - \gamma \dis _\varphi \left( \bbf^{\prime \top}(\varphi) K \pbf(\varphi) + \bbf^{\prime \prime \top}(\varphi) \ppbf (\varphi)- e_{\bs}^\top K \ppbf (\varphi) \right.\\
	&\left. - e_{\dbs}^\top \pppbf (\varphi) \right) - \left( \bbf ^{\prime \top}(\varphi) K - \bbf^{\prime \prime \top}(\varphi) B + \bbf^{\prime \prime \prime \top}(\varphi) \right) e_{\dbs}.
	\end{aligned}}
	\end{equation}
Considering $V_2 = \frac{1}{2} \dis_{\varphi}^2$, the time derivative of $V_2$ is as
\begin{equation}
{\footnotesize
\begin{aligned}
\dot{V}_2 \leq - &\left( \bbf^{\prime \top}(\varphi) K \pbf(\varphi) + \bbf^{\prime \prime \top}(\varphi) \ppbf(\varphi) - \left( \ppbf_m + \pppbf_m \right) \right.\\
&\left. \sqrt{2 \dis(\bx,\bq_d(\varphi))} \right) \gamma \dis _\varphi^2 + \eta |\dis_\varphi| \sqrt{2 \dis (\bx,\bq_d(\varphi))},
\end{aligned}}
\end{equation}
where $\eta = \max_\varphi \left(\left|\bbf^{\prime \top}(\varphi) K - \bbf^{\prime \prime \top}(\varphi) B + \bbf^{\prime \prime \prime \top}(\varphi)\right|\right)$.
Since $\dis (\bx,\bq_d(\varphi)) = V_1$, and $\dot{V}_1 \leq 0$, we have $\max_{\left[\bx^\top,\varphi\right] \in \mathcal{N}_\delta} \dis (\bx,\bq_d(\varphi)) = d_0 = \dis (\bx(t_\delta),\bq_d(\varphi(t_\delta)))$.
Thus, we have
\begin{equation}
	\begin{aligned}
	\dot{V}_2 \leq - &\left( \bbf^{\prime \top}(\varphi) K \pbf (\varphi) + \bbf^{\prime \prime \top}(\varphi) \ppbf (\varphi) \right. \\
	& \left. - \left( \ppbf_m + \pppbf_m \right) \sqrt{2 d_0} \right) \gamma \dis _\varphi^2 + \eta |\dis_\varphi| \sqrt{2 d_0}.
	\end{aligned}
\end{equation}
Hence, for every $\epsilon \in (0,1)$ if
{\small\begin{equation}
\label{eq:EDVO_eq2}
| \dis _\varphi | \geq \dfrac { \eta \sqrt{2 d_0} }{\epsilon \gamma \left(\bbf^{\prime \top}(\varphi) K \pbf(\varphi) + \bbf^{\prime \prime \top}(\varphi) \ppbf (\varphi) - \left( \ppbf_m + \pppbf_m \right) \sqrt{2 d_0} \right) },
\end{equation}}
we have
\begin{equation}
\begin{aligned}
	\dot{V}_2 & \leq - (1-\epsilon) \left( \bbf^{\prime \top}(\varphi) K \pbf (\varphi) + \bbf^{\prime \prime \top}(\varphi) \ppbf (\varphi) \right.\\
	& \quad \left. - \left( \ppbf_m + \pppbf_m \right) \sqrt{2 d_0} \right) \gamma \dis _\varphi^2\\
	& \leq - (1-\epsilon) \gamma \dis _\varphi^2.
\end{aligned}
\end{equation}
Therefore, $\dis_{\varphi}^2$ decreases when the trajectory enters $\mathcal{N}_\delta$.
On the other hand, we have
\begin{equation}
- \dis_{\varphi}^2 (t_\delta) = 2 \int_{t_\delta}^{t_\nu} \dot{V}_2 dt \leq - (1-\epsilon) \gamma \int_{t_\delta}^{t_\nu} \dis _\varphi^2 dt,
\end{equation}
where $t_\delta \in \mathbb{R}^+$ is the time instance that the trajectory of the system enters $\mathcal{N}_\delta$, and $t_\nu > t_\delta$ is the time instance when $\dis_{\varphi}^2 (t_\nu) = 0$.
From the above equation, one can concludes that $\lim\limits_{\gamma \rightarrow \infty} t_\nu = t_\delta$.
Hence, if $\gamma \rightarrow \infty$, for all $\left[\bx^\top,\varphi\right] \in \mathcal{N}_\delta$, we have $\dis_\varphi = 0$, and thus $\lim\limits_{\gamma \rightarrow \infty} \dis (\bx,\bq_d(\varphi)) = \dist (\bx,\mathcal{P}_{\bbf})$.

%%%%%%%%%%%%%%%%%%%%%%%%%%%%%%%%%%%%%%%%%%%%%%%%%%%%%%%%%%%%%%%%%%%%%%%%%%%%%%%%%%%%%%%%%%%%%%%%%%%%%%%%%%%%%%%%%%%%%%%%%%%%%%%%%%%%%%%%%%%

\section{Proof of Theorem \ref{thm:psiDefinable}}
\label{app:psiDefinable}

Substituting \eqref{eq:BMDVO_stateTransformation}
into \eqref{eq:BMDVO_constrainedForPsi}, we have
\begin{equation}
\begin{aligned}
\breve{\delta}_{\by}^2 \breve{\delta}_{\dot{\by}} \left| \tanh{(\bg_v)} \right| \leq &\left|
\breve{\delta}_{\by}^2 - \breve{\delta}_{\by}^2 \tanh^2 {(\breve{\bg}_p)}\right| \delta_{\dot{\by}} \\
& = \breve{\delta}_{\by}^2 \left| I_n - \tanh^2 {(\breve{\bg}_p)} \right| \delta_{\dot{\by}}.
\end{aligned}
\end{equation}
The above inequality is simplified as
\begin{equation}
\breve{\delta}_{\by} \left| J_{\bg_p}^{-1} \tanh{(\bg_v)} \right| \leq \mathbf{1}_n.
\end{equation}
According to the above inequality and the fact that ${\left| \breve{\delta}_{\by}^{-1} J_{\bs_1} \right| \leq \mathbf{1}_n}$, one concludes that
\begin{equation}
\left| J_{\bs_1} J_{\bg_p}^{-1} \tanh{(\bg_v)} \right| \leq \mathbf{1}_n.
\end{equation}
Thus, the variable $\bsi$ is definable.

%%%%%%%%%%%%%%%%%%%%%%%%%%%%%%%%%%%%%%%%%%%%%%%%%%%%%%%%%%%%%%%%%%%%%%%%%%%%%%%%%%%%%%%%%%%%%%%%%%%%%%%%%%%%%%%%%%%%%%%%%%%%%%%%%%%%%%%%%%%%%

\section{Proof of Theorem \ref{thm:BMDVO}}
\label{app:BMDVO}

Theorem \ref{thm:BMDVO} has 4 items which are proven in the following.

%%%%%%%%%%%%%%%%%%%%%%%%%%%%%%%%%%%%%%%%%%%%%%%%%%%%%%%%%%%%%%%%%%%

\subsection{$\bx(t)$ is bounded and converges to $\bx^*(\varphi)$ from ...}

As the trajectory $\left[ \bs_1^\top,\bs_2^\top,\varphi\right] = \left[ \bg_p^\top,\bg_v^\top,\varphi\right]$ satisfies the differential equations \eqref{eq:BMDVO_Dynamics}, ${\small\left[\bx^{* ^\top},\varphi\right]^\top} = \left[ \bg_p^\top,\bg_v^\top,\varphi \right]^\top$ is the response of the bounded output programmable oscillator for the initial condition $\left[ \bg_p^\top\left(\varphi(0)\right),\bg_v^\top\left(\varphi(0)\right),\varphi(0) \right]^\top$.
To prove the bounded property of $\bx^*$, we consider
\begin{equation}
	V_3 = \dfrac{1}{2}
\begin{bmatrix}
\be_1^\top & \be_2^\top
\end{bmatrix}
P
\begin{bmatrix}
\be_1 \\ \be_2
\end{bmatrix}
\end{equation}
where $\be_1 = \bs_1 - \bg_p (\varphi)$, $\be_2 = \bs_2 - \bsi$ and
\begin{equation}
\label{eq:app_pMatrix}
P = \begin{bmatrix}
D & B \\
B & K
\end{bmatrix}.
\end{equation}
As $B$, $K$, $D$ and $KB^{-1}D-B$ are diagonal positive definite matrices, $V_3$ is a PSD based on the Schur complement for block matrices \cite{zhang2006schur}.
Note that $V_3 = 0$ iff $\left[\be_1^\top,\be_2^\top\right] = \left[\mathbf{0}_n^\top,\mathbf{0}_n^\top\right]$.

The time derivative of $V_3$ along the trajectories of the bounded output programmable oscillator is as
\begin{equation}
{\small
\begin{aligned}
	\dot{V}_3 ={} &\be_2^\top \left(B-KB^{-1}D\right)J_{\bs_1}^{-1}\breve{\delta}_{\dot{\by}} \big( \tanh{(\bs_2)}-\tanh{(\bsi)} \big)\\
& - \left\| B \be_1 + K \be_2 \right\|^2 -\gamma W^2,
\end{aligned}}
\end{equation}
where
\begin{equation}
\begin{aligned}
W = \left( \be_1^\top D + \be_2^\top B \right) J_{\bg_p}^{-1} \breve{\delta}_{\dot{\by}} \tanh{(\bg_v (\varphi))}\\
 + \left( \be_1^\top B + \be_2^\top K \right) \bsi_\varphi.
\end{aligned}
\end{equation}
As $KB^{-1}D-B$, $J_{\bs_1}$ and $\breve{\delta}_{\dot{\by}}$ are diagonal PD matrices, and $\gamma$ is a non-negative constant, $\dot{V}_3 \leq 0$.
Thus, $\be_1$ and $\be_2$ are bounded.
Moreover, as $V_3$ is radially unbounded w.r.t. $\left\| \left[\be_1^\top,\be_2^\top \right]\right\|$, one can conclude that $\be_1$ and $\be_2$ are bounded from any initial condition.
In conclusion, if $\bbf(t) \in \mathcal{Q}$, then the shape state vector $\bx$ is bounded for any initial condition.

For the convergence of $\bx$ to $\bx^*$, it is sufficient to examine the largest invariant subset of $\Gamma = \left\{\left[\bx^\top, \varphi\right]\right\}: \dot{V}_3 = 0$.
According to the dynamic \eqref{eq:BMDVO_Dynamics}, $\left\{\left[\bx^{*^\top}, \varphi\right]\right\}$ is the only invariant subset of $\Gamma$.
Therefore, according to the LaSalle Lemma \cite{khalil2002nonlinear}, we can conclude the convergence of $\bx$ to $\bx^*$.
The global convergence of $\bx^*$ is the result of the radially unbounded property of $\dot{V}_3$ w.r.t. $\left\|\bx-\bx^*\right\|$.

Note that, the same as the programmable oscillator, the phase state $\varphi$ is an unbounded state.
While, from \eqref{eq:BMDVO_phaseDynamic} and the assumption A3.1, one can conclude that $\varphi$ is continuously differentiable, and hence $\varphi$ does not have finite escape time~\cite{khalil2002nonlinear}.

%%%%%%%%%%%%%%%%%%%%%%%%%%%%%%%%%%%%%%%%%%%%%%%%%%%%%%%%%%%%%%%%%%%%

\subsection{if $\gamma = 0$, $\left[\bx^{* \top}(t+\varphi(0)),t+\varphi(0)\right]^\top$ is globally AS}

For $\gamma = 0$, \eqref{eq:BMDVO_phaseDynamic} is simplified to $\dot{ \varphi } = 1$, and thus $\varphi(t) = t + \varphi(0)$.
Hence, the system \eqref{eq:BMDVO_Dynamics} is simplified to the $2n$ dimensional system whose state space vector is $\bx$.
Accordingly, the function $V_3$ is the Lyapunov function of the system, and thus the global AS of ${\small\left[\bg_p^\top (t+\varphi(0)),\bg_v^\top (t+\varphi(0))\right]^\top}$ is concluded based on the Lyapunov stability theorem \cite{khalil2002nonlinear}.

%%%%%%%%%%%%%%%%%%%%%%%%%%%%%%%%%%%%%%%%%%%%%%%%%%%%%%%%%%%%%%%%%%%%

\subsection{if $\gamma \rightarrow \infty$, $\bx^*(t)$ is the globally stable limit cycle}

$\bx^*(t)$ is the globally stable limit cycle iff $\mathcal{P}_{\bg}$ is globally AS where $\mathcal{P}_{\bg}$ is the set of the states belonging to $\bx^*(t)$, \ie \cite{khalil2002nonlinear}
\begin{equation}
\mathcal{P}_{\bg} = \left\{ \left[ \bs_1^\top, \bs_2^\top, \varphi \right] \, : \, \exists \varphi, \, \bs_1=\bg_p(\varphi),\bs_2=\bg_v(\varphi) \right\}.
\end{equation}
To investigate the AS of the set $\mathcal{P}_{\bg}$, we prove that for $\gamma \rightarrow \infty$,
\begin{itemize}
	\item $\forall \left[\bx^\top,\varphi\right] \in \mathcal{P}_{\bg}, V_3(\bx, \varphi) = 0$,
	\item $\forall \left[\bx^\top,\varphi\right] \notin \mathcal{P}_{\bg}, V_3(\bx, \varphi) > 0$, and
	\item $\forall \left[\bx^\top,\varphi\right] \notin \mathcal{P}_{\bg}, \dot{V}_3(\bx, \varphi) < 0$.
\end{itemize}
Thus, $\mathcal{P}_{\bg}$ is the stable limit cycle of the system \eqref{eq:BMDVO_Dynamics} according to the extension of the Lyapunov stability theorem for invariant sets \cite{zubov1964methods}.
The global stability of $\mathcal{P}_{\bg}$ is the result of the radially unbounded property of $V_3$.
For this purpose, we prove that for $\gamma \rightarrow \infty$, $\left[\be_1,\be_2\right] = \left[\mathbf{0}_n,\mathbf{0}_n\right]$ iff $\left[\bx^\top,\varphi\right] \in \mathcal{P}_{\bg}$.
Since $\be_1 = \bs_1 - \bg_p(\varphi)$ and $\be_2 = \bs_2 - \bg_v(\varphi)$, one can figure out that if $\be_1=\be_2 = \mathbf{0}_n$, then $\left[\bx^\top,\varphi\right] \in \mathcal{P}_{\bg}$.
To prove that for $\left[\bx^\top,\varphi\right] \in \mathcal{P}_{\bg}$, we have $\be_1=\be_2 = \mathbf{0}_n$, we show that for $\gamma \rightarrow \infty$ and $\left[\bx^\top,\varphi\right] \in \mathcal{N}_\rho$ where $\mathcal{N}_\rho$ is the $\rho$-neighborhood of $\mathcal{P}_{\bg}$, $\bx^*(\varphi)$ is the point from the curve of $\mathcal{P}_{\bg}$ in the shape space having the minimum distance from the current shape states.
In this regard, we define the distance between $\bx = {\small \left[\bs_1^\top,\bs_2^\top\right]^\top}$ and the curve of $\mathcal{P}_{\bg}$ in the shape space as
\begin{equation}
\dist(\bx,\mathcal{P}_{\bg}) = \min_{\tau \in \left[0,T\right]} \dis(\bx,\bx^*(\tau)),
\end{equation}
where $T \in \mathbb{R}^+$ is the period of $\bbf(t)$ and $\dis(\bx,\bx^*(\tau))$ is as
\begin{equation}
\label{eq:app_BMDVO_distancDef}
\begin{aligned}
	\dis(\bx,\bx^*(\tau)) ={}& \dfrac{1}{2}
	\left[
	\left(\bs_1-\bg_p(\tau)\right)^\top, \left(\bs_2-\bg_v(\tau)\right)^\top
	\right]\\
	&P
    \left[
	\left(\bs_1-\bg_p(\tau)\right)^\top, \left(\bs_2-\bg_v(\tau)\right)^\top
	\right]^\top.
\end{aligned}
\end{equation}
In the following, we prove that $\dis(\bx,\bx^*(\varphi)) = \dist(\bx,\mathcal{P}_{\bg})$ if $\gamma \rightarrow \infty$.
Since $\dis(\bx,\bx^*(\varphi))$ is continuous, its extreme is at $\dfrac{\partial \dis  (\bx,\bx^*(\varphi))}{\partial \varphi} = 0$.
Thus, we consider the first and second derivatives of $\dis (\bx, \bx^*(\varphi))$ w.r.t. the phase state, \ie
\begin{equation}
\begin{aligned}
& \dis_\varphi = \dfrac{\partial \dis}{\partial \varphi} (\bx,\bx^*) =
- \begin{bmatrix}
\be_1^\top & \be_2^\top
\end{bmatrix}
P
\begin{bmatrix}
\bg^\prime_p (\varphi) \\ \bsi_\varphi
\end{bmatrix}, \\
&\dis_{\varphi \varphi} = \dfrac{\partial^2 \dis}{\partial \varphi^2} (\bx,\bx^*) = 
- \begin{bmatrix}
\be_1^\top & \be_2^\top
\end{bmatrix}
P
\begin{bmatrix}
\bg^{\prime \prime}_p (\varphi) \\ \bsi^{\prime \prime}
\end{bmatrix}\\
& \qquad \qquad \qquad \qquad \quad + \begin{bmatrix}
\bg_p^{\prime \top} (\varphi) & \bsi^\top_\varphi
\end{bmatrix}
P
\begin{bmatrix}
\bg^\prime_p(\varphi) \\ \bsi_\varphi
\end{bmatrix}.
\end{aligned}
\end{equation}
The lower bound of $\dis_{\varphi \varphi}$ is as
\begin{equation}
\label{eq:app_lowerBoundDff}
\dis_{\varphi\varphi} \geq \underline{\lambda} \begin{Vmatrix}
\bg_p^{\prime \top}(\varphi) & \bsi^\top_\varphi
\end{Vmatrix}^2
- \kappa \sqrt{2 \dis (\bx,\bx^*)},
\end{equation}
where $\underline{\lambda}$ is the minimum eigenvalue of $P$, and $\kappa = \max_\varphi \left\| P^{1/2} 
\begin{bmatrix}
\bg_p^{\prime \prime} (\varphi) & \bsi_{\varphi \varphi}
\end{bmatrix} \right\|$.
Note that $D-BK^{-1}B $ is PD, thus $\underline{\lambda} > 0$.
Since $\underline{\lambda}$ is positive and $\not\exists \varphi$ that $\bg_p^\prime(\varphi) = \bsi_\varphi(\varphi) = \mathbf{0}_n$, one can find $\rho \in \mathbb{R}^+$ such that $\forall \left[\bx^\top,\varphi\right] \in \mathcal{N}_\rho$ where $\mathcal{N}_\rho$ is the $\rho$-neighborhood of $\mathcal{P}_{\bg}$, we have $\dis_{\varphi\varphi} > 0$.
As a result, $\forall \left[\bx^\top,\varphi\right] \in \mathcal{N}_\rho$, $\dis(\bx,\bx^*(\varphi)) = \dist(\bx,\mathcal{P}_{\bg})$ iff $\dis_{\varphi} = 0$.
To investigate time evolution of $\dis_\varphi$, we compute its time derivative along the trajectories of \eqref{eq:BMDVO_Dynamics} as
\begin{equation}
{\small
	\label{eq:app_BMDVO_dppDiff}
	\begin{aligned}
	\dot{\dis} _\varphi = &- \left( \begin{bmatrix}
	\bg_p^{\prime \top}(\varphi) & \bsi^{\top}_\varphi
	\end{bmatrix} P \begin{bmatrix}
	\bg_p^{\prime \top}(\varphi) \\ \bsi^{\top}_\varphi
	\end{bmatrix} 
	- \begin{bmatrix}
	\be_1^\top & \be_2^\top
	\end{bmatrix} P \begin{bmatrix}
	\bg_p^{\prime \prime}(\varphi) \\ \bsi_{\varphi \varphi}
	\end{bmatrix} \right) \gamma \dis _\varphi \\
	&- \begin{bmatrix}
	\be_1^\top & \be_2^\top
	\end{bmatrix} \left( P
	\begin{bmatrix}
	\bg_p^{\prime \prime}(\varphi) \\ \bsi_{\varphi \varphi} + \dot{\bsi}_\varphi
	\end{bmatrix}
	- \begin{bmatrix}
	B^2 & BK \\ KB & K^2
	\end{bmatrix} \begin{bmatrix}
	\bg^\prime_p(\varphi) \\ \bsi_\varphi
	\end{bmatrix}\right) \\
	&- \left( \tanh{(\bs_2)} - \tanh{(\bsi)} \right) J_{\bs_1}^{-1} \breve{\delta}_{\dot{\by}} \left( B-B^{-1}DK \right) \bsi_\varphi
	\end{aligned}}
\end{equation}
Considering $V_4 = \frac{1}{2} \dis_{\varphi}^2$, the time derivative of $V_4$ is as
\begin{equation}
\label{eq:app_BMDVO_V11Diff}
\begin{aligned}
\dot{V}_4 = \dis_\varphi \dot{\dis}_\varphi \leq {}& - \left( \underline{\lambda} 
\begin{Vmatrix}
\bg_p^{\prime \top}(\varphi) & \bsi^{\top}_\varphi
\end{Vmatrix}^2
- \kappa \sqrt{2 d_0} \right) \gamma \dis _\varphi^2 \\
&+ \rho(d_0) |\dis_\varphi|,
\end{aligned}
\end{equation}
where
\begin{equation}
\rho(d) = \sqrt{2 d_0} \left( \kappa + \upsilon \right)
+ 2  \tanh(\sqrt{2 d_0}\mu)
\end{equation}
and $\upsilon = \max_\varphi \left( \left\| \begin{bmatrix}
B^2 & BK \\ KB & K^2
\end{bmatrix} \begin{bmatrix}
\bg^\prime_p(\varphi) \\ \bsi_\varphi
\end{bmatrix} \right\| \right)$
and ${\mu = \max_\varphi \left( J_{\bs_1}^{-1} \breve{\delta}_{\dot{\by}} \left( B-B^{-1}DK \right) \bsi_\varphi \right)}$.
Based on \eqref{eq:app_BMDVO_V11Diff}, for $\epsilon \in (0,1)$ if 
\begin{equation}
\label{eq:app_BMDVO_AOSregion}
| \dis _\varphi | \geq \dfrac { \rho(d_0) }{\epsilon \gamma \left( \underline{\lambda} \begin{Vmatrix}
	\bg_p^{\prime \top}(\varphi) & \bsi^{\top}_\varphi
	\end{Vmatrix}^2
	- \kappa \sqrt{2 d_0} \right) },
\end{equation}
then we have
\begin{equation}
\dot{V}_4 \leq - (1-\epsilon) \left( \underline{\lambda} \begin{Vmatrix}
\bg_p^{\prime \top}(\varphi) & \bsi^{\top}_\varphi
\end{Vmatrix}^2
- \kappa \sqrt{2 d_0} \right) \gamma \dis _\varphi^2.
\end{equation}
Therefore, $\dis_{\varphi}^2$ decreases when the trajectory enters $\mathcal{N}_\rho$.
On the other hand, we have
\begin{equation}
- \dis_{\varphi}^2 (t_\rho) = 2 \int_{t_\rho}^{t_\nu} \dot{V}_4 dt \leq - (1-\epsilon) \gamma \xi \int_{t_\delta}^{t_\nu} \dis _\varphi^2 dt,
\end{equation}
where $t_\rho \in \mathbb{R}^+$ is the time instance that the trajectory of the system enters $\mathcal{N}_\rho$, $t_\nu > t_\rho$ is the time instance when $\dis_{\varphi}^2 (t_\nu) = 0$, and $\xi = \min \left| \underline{\lambda} \begin{Vmatrix}
\bg_p^{\prime \top}(\varphi) & \bsi^{\top}_\varphi
\end{Vmatrix}^2
- \kappa \sqrt{2 d_0} \right|$.
From the above equation, one can conclude that $\lim\limits_{\gamma \rightarrow \infty} t_\nu = t_\rho$.
Hence, if $\gamma \rightarrow \infty$, for all $\left[\bx^\top,\varphi\right] \in \mathcal{N}_\rho$, we have $\dis_\varphi = 0$, and thus $\lim\limits_{\gamma \rightarrow \infty} \dis (\bx,\bx^*(\varphi)) = \dist (\bx,\mathcal{P}_{\bg})$.

%%%%%%%%%%%%%%%%%%%%%%%%%%%%%%%%%%%%%%%%%%%%%%%%%%%%%%%%%%%%%%%%%%%%

\subsection{the output $\by(t)$ converges to the curve of $\bq_d(t)$ in ...}

As $\bx$ converges to $\bx^*(\varphi)$, $\by(t)$ converges to the curve of $\bq_d(t)$ in the shape $(\by,\dot{ \by })$.
Moreover, according to the definition of $\by$ in \eqref{eq:BMDVO_Dynamics} and the fact that $\bx$ is bounded, we have ${\by_{min}=\by_{avg} - \delta_{\by} < \by < \by_{avg} + \delta_{\by} = \by_{max}}$.
In addition, based on the dynamic of the bounded output programmable oscillator, the time derivative of the output is as
\begin{equation}
\dot{\by} = \delta_{\dot{y}} \tanh{(\bs_1)}. 
\end{equation}
Thus, $ |\dot{\by}| \leq \delta_{\dot{y}}$.
In conclusion, we have $\by(t) \in \mathcal{Q}$.

\ifCLASSOPTIONcaptionsoff
  \newpage
\fi


\bibliographystyle{IEEEtran}
\bibliography{ref}


\begin{IEEEbiography}[{\includegraphics[width=1in,height=1.20in,clip]{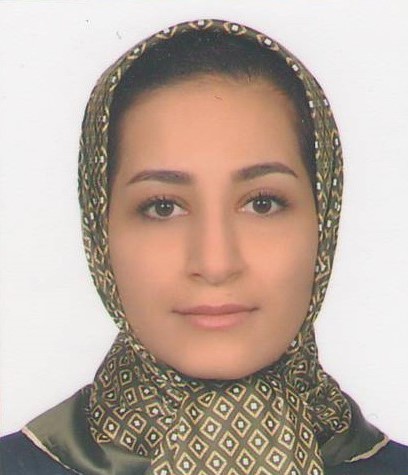}}]{Venus Pasandi}
received her bachelor and master degrees in mechanical engineering from Isfahan University of Technology, Iran and Amirkabir University of Technology, Iran in 2010 and 2012, respectively. 
In 2020, she earned her PhD title in Mechanical Engineering (Robotics and Control) from Isfahan University of Technology, Iran.
From 2018-2019, she was a PhD student fellow in the Dynamic Interaction Control lab at Istituto Italiano di Tecnologia, Italy.
She is currently a post-doctoral researcher in the Artificial and Mechanical Intelligence lab at Istituto Italiano di Tecnologia, Italy.
Her current research interests include nonlinear systems and control, robotics, trajectory planning, and optimization.
\end{IEEEbiography}
\vskip -2.5\baselineskip plus -1fil
\begin{IEEEbiography}[{\includegraphics[width=1in,height=1.25in,clip,keepaspectratio]{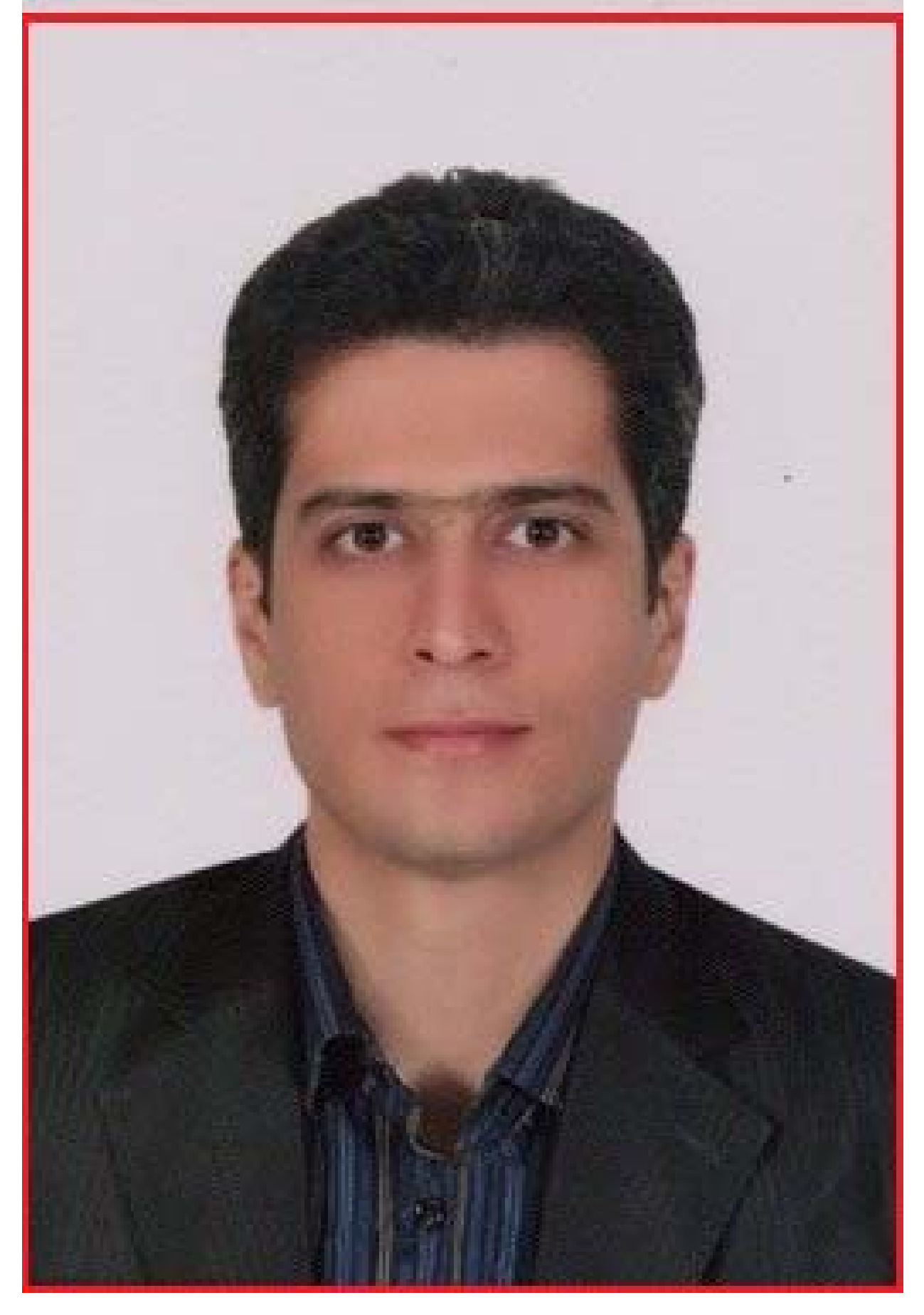}}]{Hamid Sadeghian}
	received the B.Sc. and M.Sc. degrees in Mechanical Engineering from Isfahan University of Technology and Sharif University of Technology, Tehran, Iran, in 2005 and 2008, respectively. He received the Ph.D. degree in mechanical engineering (robotics and control) from Isfahan University of Technology in 2013. From November 2010 to January 2013, he was a visiting scholar with PRISMA Laboratory, Department of Electrical Engineering and Information Technology, University of Naples, Naples, Italy. Starting from 2013 he is an assistant professor of Control and Robotics with Engineering Dept. in University of Isfahan. In Sep. 2015 he visited Institute of Cognitive System at Technical University of Munich and German Aerospace Center (DLR), Munich, Germany, for one year post-doctoral research on locomotion of bipeds. His research interests include physical human-robot interaction/cooperation, medical robotics, and nonlinear control of mechanical systems. He is currently a senior scientist at  Munich Institute of Robotics and Machine Intelligence, Technical University of Munich, Munich, Germany. 
\end{IEEEbiography}
\vskip -2.5\baselineskip plus -1fil
\begin{IEEEbiography}[{\includegraphics[width=0.9in,height=1.15in,clip]{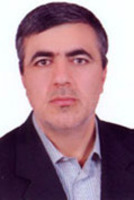}}]{Mehdi Keshmiri}
	received his BSc and MSc degrees in Mechanical Engineering from Sharif University of Technology, Iran, in 1986 and 1989, respectively. He holds a PhD in Mechanical Engineering (space dynamics) from McGill University, Montreal, Canada. He is currently a Full Professor in the Department of Mechanical Engineering of Isfahan University of Technology (IUT), Isfahan, Iran. His research is mainly focused on system dynamics, control systems and dynamics and control of robotic systems. He has presented and published more than 150 papers in international conferences and journals and supervised more than 80 PhD and master students. He has been also involved in science and technology park development in Iran and is known as one of the leaders in this regard.
\end{IEEEbiography}
\vskip -2.5\baselineskip plus -1fil
\begin{IEEEbiography}[{\includegraphics[width=1in,height=1.10in,clip]{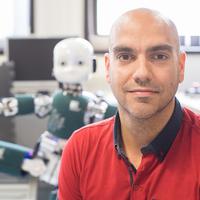}}]{Daniele Pucci}
	received the bachelor and master degrees in Control Engineering with highest honors from ”Sapienza”, University of Rome, in 2007 and 2009, respectively. In 2009, he also received the ”Excellence Path Award” from Sapienza. In 2013, he earned the PhD title with a thesis prepared at INRIA Sophia Antipolis, France, under the supervi- sion of Tarek Hamel and Claude Samson. The PhD program was jointly with ”Sapienza”, University of Rome, so in 2013 Daniele received also the PhD title in Control Engineering from Sapienza with the supervision of Salvatore Monaco. From 2013 to 2017, he has been a postdoc at the Istituto Italiano di Tecnologia (IIT) working within the EU project CoDyCo. From August 2017 to August 2021, he has been the head of the Dynamic Interaction Control lab. The main lab research focus was on the humanoid robot locomotion and physical interaction problem, with specific attention on the control and planning of the associated nonlinear systems. Also, the lab has been pioneering Aerial Humanoid Robotics, whose main aim is to make flying humanoid robots. 
	Daniele has also been the scientific PI of the H2020 European Project AnDy, he is task leader of the H2020 European Project SoftManBot, and he is the coordinator of the joint laboratory between IIT and Honda JP. 
	In 2019, he was awarded as Innovator of the year Under 35 Europe from the MIT Technology Review magazine. 
	Since 2020 and in the context of the split site PhD supervision program, Daniele is a visiting lecturer at University of Manchester. In July 2020, Daniele was selected as a member of the Global Partnership on Artificial Intelligence (GPAI). 
	Since September 1st 2021, Daniele is the PI leading the Artificial and Mechanical Intelligence research line at IIT. The research team combines AI and Mechanics to devise the next generation of Humanoid Robots.
\end{IEEEbiography}
		

%\vfill


\end{document}